\documentclass{article}
\usepackage[T1]{fontenc}
\usepackage[utf8]{inputenc}
\usepackage[margin=1.125in]{geometry}
\usepackage{lineno}
\usepackage{amssymb,amsmath}
\usepackage{lipsum}
\usepackage{color}
\usepackage{subfigure}
\usepackage{epsfig}
\usepackage{epstopdf}
\usepackage{graphicx}
\usepackage{algpseudocode}
\usepackage{algorithm}
\usepackage{hyperref}
\usepackage{amsopn}
\usepackage{mathrsfs}
\usepackage[normalem]{ulem}
\ifpdf
  \DeclareGraphicsExtensions{.eps,.pdf,.png,.jpg}
\else
  \DeclareGraphicsExtensions{.eps}
\fi

\usepackage{enumitem}
\setlist[enumerate]{leftmargin=.5in}
\setlist[itemize]{leftmargin=.5in}

\usepackage{amsthm}

\numberwithin{equation}{section}
\numberwithin{algorithm}{section}

\newcommand{\norm}[1]{\lVert#1\rVert}

\newcommand{\st}{\mbox{s.t.}}

\newcommand{\tr}{{\rm{Tr}}}

\DeclareMathOperator*{\diag}{diag}

\DeclareMathOperator*{\argmin}{argmin}

\newcommand{\vu}{\mathbf{u}}
\newcommand{\vv}{\mathbf{v}}

\newcommand{\vx}{\mathbf{x}}

\newcommand{\cN}{\mathcal{N}}

\DeclareMathOperator*{\rank}{rank}
\DeclareMathOperator{\shrink}{shrink}

\newcommand{\R}{\mathbb{R}}
\graphicspath{{./figs/}}

\providecommand{\keywords}[1]{\textbf{Keywords: } #1}

\title{Human Motion Detection Based on Dual-Graph and Weighted Nuclear Norm Regularizations}

\author{Jing Qin\thanks{Department of Mathematics, University of Kentucky, Lexington, KY
  (\texttt{jing.qin@uky.edu}).}
  \and Biyun Xie\thanks{Department of Electrical and Computer Engineering, University of Kentucky, Lexington, KY, USA}
  }
\date{}

\begin{document}
\maketitle

\begin{abstract}
Motion detection has been widely used in many applications, such as surveillance and robotics. Due to the presence of the static background, a motion video can be decomposed into a low-rank background and a sparse foreground. Many regularization techniques that preserve low-rankness of matrices can therefore be imposed on the background. In the meanwhile, geometry-based regularizations, such as graph regularizations, can be imposed on the foreground. Recently, weighted regularization techniques including the weighted nuclear norm regularization have been proposed in the image processing community to promote adaptive sparsity while achieving efficient performance. In this paper, we propose a robust dual graph regularized moving object detection model based on a novel weighted nuclear norm regularization and spatiotemporal graph Laplacians. Numerical experiments on realistic human motion data sets have demonstrated the effectiveness and robustness of this approach in separating moving objects from background, and the enormous potential in robotic applications.

\end{abstract}


%%%%%%%%%%%%%%%%%%%%%%%%%%%%%%%%%%%%%%%%%%%%%%%%%%%%%%%%%%%%%%%%%%%%%%%%%%%%%%%%
\section{Introduction}

The development of advanced robotic technologies has released traditional robots isolated by fences or other protective barriers to environments with human beings \cite{zanchettin2019towards}. Such kinds of robots that are safe and intelligent enough to work alongside or directly interact with humans are called collaborative robots, including lightweight industrial robots, social robots, and service robots \cite{stein2020collaborative}. Human motion detection plays a significant role in the motion planning and control of collaborative robots to improve the safety and efficiency of human-robot interaction. On the one hand, the detected human motion will be used as the input information of various real-time motion planning algorithms to prevent the potential collision between a robot and a human subject and guarantee the safety of human-robot interaction \cite{sajedi2022uncertainty}. On the other hand, the detected human motion can be further used for human motion analysis and prediction to enable robots to comprehend human intention and enhance the efficiency of human-robot interaction \cite{unhelkar2018human}. In this paper, we aim to develop an effective human motion detection algorithm with excellent accuracy and efficiency.

Detection of moving objects in a video with static background is usually done by separating foreground from background, and the moving objects are typically considered as the foreground. Background modeling is crucial in designing a moving object detection algorithm. Many subspace learning methods such as principal component analysis (PCA) have been developed to model background \cite{bouwmans2009subspace} by reducing the dimensionality and learning the intrinsic low-dimensional subspaces. In practice, a background matrix can be generated by concatenating the vectorized versions of background images of a video, which naturally possesses the low-rank structure. Thus sparsity of singular values is expected for a background matrix. In one of the most popular methods - robust PCA (RPCA) \cite{candes2011robust}, nuclear-norm regularization is used to enforce the matrix low-rankness as a convex relaxation of the matrix rank. Numerous variants of RPCA have been proposed \cite{javed2015background,javed2016spatiotemporal} and a comprehensive review can be found \cite{bouwmans2017decomposition}. %To enhance the low-rankness, some other regularizations based on matrix norms, including the matrix max-norm \cite{javed2015background,javed2016spatiotemporal}, have been developed to model the background.
Recently, adaptive regularization techniques have been developed to promote sparsity and achieve fast convergence of the regularized algorithms. For example, the weighted nuclear norm (WNN) regularization has shown effectiveness in various image and data processing applications \cite{gu2014weighted}, which can be considered as a natural extension of reweighted L1 \cite{candes2008enhancing} and a more general ERror Function based regularization (ERF) \cite{guo2021novel}.

In this paper, we use the ERF-weighted nuclear norm regularization (ERF-WNN) imposed on the matrix singular values to enforce the adaptive low-rankness. In addition, a video usually has complicated geometry and varying smoothness in either the spatial domain or the temporal domain. To preserve those geometrical structures in the background, we create a spatial graph and a temporal graph, which are then embedded in the graph regularizations of the background matrix. Generation of the spatial graph is implemented by computing the patchwise similarity to exploit the nonlocal similarity. To reduce the computational cost, we only consider the $k$-nearest neighboring pixels in terms of similarity when calculating the pairwise similarity. On the other hand, the $\ell_1$-regularization is imposed on the foreground due to its sparsity. Thus far, we propose a spatiotemporal dual-graph regularized moving object detection model, which is solved by the alternating direction method of multipliers (ADMM). After introducing a few auxiliary variables and splitting regularizers, we obtain a sequence of subproblems. Among them, one quadratic subproblem is solved by gradient descent, and the other subproblems all have closed-form solutions which can be implemented efficiently. Furthermore, we test our algorithm on the two real RGB videos containing  whole-body motion and an arm motion under a static background, respectively. Performance is compared with other related methods in terms of background recovery and foreground detection accuracy.

There are three major contributions for this work.
\begin{enumerate}
\item We propose a novel motion detection model which involves spatiotemporal graph regularizations and a weighted nuclear norm regularizer with weights being defined by an error function so that both spatiotemporal geometry and low-rankness of the background can be preserved during the separation of foreground from background. Although dual-graph regularization has been explored in some works, e.g., \cite{javed2015background,javed2016spatiotemporal}, hybrid type of graph and weighted regularizations has rarely been studied in literature. In addition, we adopt the L1-type of data fidelity to make the model robust to the noise or outliers. This brings theoretical contributions to mathematical modeling and low-rank matrix approximation.
\item We develop an efficient algorithm based on ADMM and reweighting scheme to solve the proposed model. We adopt the computationally cheap gradient descent to solve the quadratic subproblem rather than applying a Sylvester equation solver due to the gigantic size of the coefficient matrix. It has empirically shown that a good-quality result can typically be reached within a few dozens of iterations.
\item We test real data sets that are generated to simulate the human motion settings for robotic applications, including linear motion, rotation, and multi-object motion. Numerical experiments have shown the great potential and effectiveness of this algorithm over the other related works in real-time applications.
\end{enumerate}

The rest of this paper is organized as follows. In Section \ref{sec:LR}, we provide a brief introduction of moving object detection and low-rank based models. In Section \ref{sec:method}, we propose a novel spatiotemporal dual graph regularized moving object detection method based on the ERF-WNN regularization. Numerical experiments on two realistic videos with moving objects and the results are reported in Section \ref{sec:exp}. Finally, conclusions of this research and future work are presented in Section \ref{sec:con}.

\section{Low-Rank Models}\label{sec:LR}
Throughout the paper, we use boldface lowercase letters to denote vectors and uppercase letters to denote matrices. For $p\geq1$, the $\ell_p$-norm of a vector $\vx\in\R^n$ is given by $\norm{\vx}_p=(\sum_{i=1}^n|x_i|^p)^{1/p}$. The entry-wise $\ell_1$-norm of a matrix $X\in\R^{n\times m}$ is defined as $\norm{X}_1=\sum_{i,j}|x_{ij}|$ where $x_{ij}$ is the $(i,j)$-th entry of $X$. The Frobenious norm of $X$, denoted by $\norm{X}_F$, is defined as $\sqrt{\sum_{i,j}|x_{ij}|^2}$. The nuclear norm of $X$, denoted by $\norm{X}_*$, is defined as the sum of all singular values of $X$. We use the symbol $\diag(\vx)$ to denote a diagonal matrix whose diagonal entries form the vector $\vx$, and $I_n$ as the $n$-by-$n$ identity matrix.

Consider a video with a static background consisting of $m$ frames of gray-scale images with size $n_1\times n_2$. By reshaping each image as a vector, we convert a video to a matrix $D$ of size $n\times m$ where $n=n_1n_2$ is the number of total spatial pixels. Assume that $D$ can be decomposed into the background component $L$ and the foreground component $S$, where $L,S\in\R^{n\times m}$. Here we let $S$ correspond to the moving object. That is, we have $D=L+S$ in the noise-free case. In order to retrieve $L$ and $S$ from $D$ simultaneously, we apply regularization techniques on both variables. Since the background is static, the matrix $L$ typically has low-rank structures. In the meanwhile, the object occupies a small portion of each frame and thereby $S$ is sparse. Thus we consider the problem
\[
\min_{L,S}\mathrm{rank}(L)+\lambda\norm{S}_1\quad \st\quad D=L+S.
\]
Here $\lambda>0$ is a regularization parameter and $\rank(L)$ equals the number of nonzero singular values of $L$. Since this problem is NP-hard, matrix rank is replaced by the nuclear norm which leads to the RPCA model \cite{candes2011robust}
\[
\min_{L,S}\norm{L}_*+\lambda\norm{S}_1\quad\st\quad D=L+S.
\]
Here $\norm{L}_*$ is the nuclear norm of $L$, i.e., sum of all the singular values of $L$.
In some RPCA variants \cite{shen2014online,javed2015background}, the matrix max-norm based regularizer has been used to replace the nuclear norm
\[
\min_{L,S}\norm{L}_{\max}+\lambda\norm{S}_1\quad \st\quad D=L+S.
\]
Here the max-norm of $L$ is given by
\[
\norm{L}_{\max}=\min_{L=UV'}\norm{U}_{2\to\infty}\norm{V}_{2\to\infty},
\]
where $V'$ is the transpose of $V$ and the operator norm $\norm{U}_{2\to\infty}=\max_{\norm{\vx}_2=1}\norm{U\vx}_\infty$. See \cite{srebro2005rank} for the connections between the matrix nuclear norm and the max-norm. They both are convex and can be used to describe the low-rankness of the background matrix.

Recently, weighted nuclear norm minimization (WNNM) has been proposed and shown outstanding performance in a lot of image processing applications \cite{gu2014weighted}. Specifically, weighted nuclear norm (WNN) is defined as
\begin{equation}\label{eqn:matnorm}
\norm{L}_{W,*}:=\sum_{i}w_i\sigma_i(L),
\end{equation}
where $\sigma_i(L)$ is the $i$-th singular value of $L$ in the decreasing order and $w_i\in[0,1]$ is the weight associated with the $i$-th singular value.
The selection of weights is related to adaptive sparsity regularizers such as iteratively reweighted L1 (IRL1) \cite{candes2008enhancing}. Specifically, IRL1 considers a reciprocal of the previous iterates and yields the form in nuclear norm setting
$
w_i=\frac1{\sigma_i(L^{(i)})+\varepsilon}
$
where $L^{(i)}$ is the recovered low-rank component from the $i$-th iteration and $\varepsilon>0$ is a parameter to ensure the stability.
More recently, ERF generalizes IRL1 with improved sparsity and convergence speed \cite{guo2021novel}, where the weight is based on the error function.
Both can be naturally extended to the singular values in the WNN framework to promote the low-rankness. In this paper, we adopt a novel ERF-WNN as the regularizer that will be detailed in the next section.

\section{Proposed Method}\label{sec:method}
The problem of motion detection can be cast as the foreground and the background separation. In addition to the low-rankness assumption of the background matrix, we can use spatial and temporal graph regularizations to preserve the sophisticated geometry. To split multiple regularization terms in the proposed motion detection model, we apply ADMM to derive an efficient algorithm.

\subsection{Spatial and Temporal Graph Laplacians}\label{sec:lap}
In what follows, we will describe the generation of spatial and temporal graph Laplacians and their corresponding graph regularizers on the background.

For a reshaped video $D\in\mathbb{R}^{n\times m}$, rows and columns of $D$ correspond to the spatial and the temporal samples, respectively. Consider a weighted temporal graph $G_t=(V_t,E_t,A_t)$ where $V_t=\{\vv_i^t\}_{i=1}^m$ is a set of temporal samples, $E_t$ is an edge set and $A_t\in\R^{m\times m}$ is the adjacency matrix which defines the weights. First, we generate an adjacency matrix $A_t$ whose $(i,j)$-th entry is given by
\begin{equation}\label{eqn:sims}
(A_t)_{i,j}=\exp\left(-\frac{\norm{\vv_i^t-\vv_j^t}_2^2}{h_t^2}\right),\quad
\end{equation}
for $i,j\in\{1,\ldots,m\}$. Here $h_t>0$ is a temporal filtering parameter. Let $W_t$ be the degree matrix of $G_t$ where $(W_t)_{i,i}=\sum_{j=1}^m(A_t)_{i,j}$. Next we define a symmetrically normalized temporal graph Laplacian $\Phi_t\in\R^{m\times m}$ given by
\[
\Phi_t=I_m-W_t^{-1/2}A_tW_t^{-1/2}.
\]
Note that $W_t^{-1/2}$ is a diagonal matrix whose $i$-th diagonal entry is $(W_t)_{i,i}^{-1/2}$.

Likewise, we consider a weighted spatial graph $G_s=(V_s,E_s,A_s)$ where $V_s=\{\vv_i^s\}_{i=1}^n$ is a set of spatial samples, $E_s$ is the edge set and $A_s\in\R^{n\times n}$ is a spatial adjacency matrix. Slightly different from the construction of $A_t$, we consider the patchwise similarity in the spatial domain for $A_s$. Specifically, the $(i,j)$-th entry of $A_s$ is given by
\begin{equation}\label{eqn:simt}
(A_s)_{i,j}=\exp\left(-\frac{\norm{\mathcal{N}(\vv_i^s)-\mathcal{N}(\vv_j^s)}_F^2}{h_s^2}\right)
\end{equation}
for $i,j\in\{1,\ldots,n\}$, where $\mathcal{N}(\vv_i^s)\in\R^{p^2\times m}$ is a reshaped version of the video patch centered at the $i$-th pixel and $h_s>0$ is the spatial filtering parameter. To reduce the computational cost, we consider the $k$-nearest neighbors in terms of location for calculating $A_s$. Specifically, we use the four-nearest neighboring spatial pixels to compute the patch-based similarity for generating the spatial adjacency matrix $A_s$. Likewise we use the four-nearest neighboring temporal pixels to compute $A_t$.  Moreover, it is worth noting that Gaussian smoothing could be embedded to the calculation of patchwise similarity in the presence of noise. Now we define the symmetrically normalized graph Laplacian in the spatial domain as
\[
\Phi_s=I_n-W_s^{-1/2}A_sW_s^{-1/2}.
\]
Similar to $W_t$, $W_s$ is the degree matrix corresponding to $G_s$ which can be obtained using $A_s$. Here we save all graph Laplacians as sparse matrices to circumvent the out-of-memory issue.

Furthermore, we can use Nystr\"{o}m method \cite{fowlkes2004spectral} to retrieve a low-rank approximation of symmetrically normalized graph Laplacian by taking random samples of patchwise similarity in the spatial domain and treating spectral information as a feature space.

\subsection{Robust Dual-Graph Regularized Method}
Let $D\in\R^{n\times m}$ be the reshaped video with $n$ spatial pixels and $m$ temporal frames. Assume that $\Phi_s\in\R^{n\times n}$ and $\Phi_t\in\R^{m\times m}$ are the respective spatial and temporal graph Laplacians, which are obtained from Section~\ref{sec:lap}. We propose a robust foreground-background separation model of the form
\[\begin{aligned}
&\min_{L,S\in\R^{n\times m}}\norm{D-L-S}_1+\lambda_1\norm{L}_{W,*}+\lambda_2
\norm{S}_1\\
&+\frac{\gamma_1}2\tr(L^T\Phi_s L)+\frac{\gamma_2}2\tr(L\Phi_t L^T).
\end{aligned}\]
Here we adopt the $L_1$-norm in the first data fidelity term to enforce the robustness of the method and suppress the outliers for recovering the low-rank component, and $\norm{\cdot}_{W,*}$ is the WNN defined in \eqref{eqn:matnorm} with weights generated by ERF. The last two graph regularization terms are used to enforce the spatiotemporal smoothness for the background. By introducing two auxiliary variables $U$ and $V$, we rewrite the above problem
\begin{equation}
\min_{L,S,U,V}\norm{V}_1+\lambda_1\norm{U}_{W,*}+\lambda_2
\norm{S}_1+\frac{\gamma_1}2\tr(L^T\Phi_s L)
+\frac{\gamma_2}2\tr(L\Phi_t L^T),\quad \st \quad U=L,\,D-L-S=V.
\end{equation}
Define the augmented Lagrangian
\[\begin{aligned}
\mathcal{L}&(V,U,S,\widetilde{U},\widetilde{V})
=\norm{V}_1+\lambda_1\norm{U}_{W,*}+\lambda_2
\norm{S}_1\\
&+\frac{\gamma_1}2\tr(L^T\Phi_s L)+\frac{\gamma_2}{2}\tr(L\Phi_t L^T)
+\frac{\rho_1}2\norm{U-L+\widetilde{U}}_F^2\\
&+\frac{\rho_2}2\norm{D-L-S-V+\widetilde{V}}_F^2.
\end{aligned}
\]
Based on the ADMM framework, we obtain the algorithm
\[
\left\{\begin{aligned}
L&\leftarrow \argmin_L\frac{\gamma_1}2\tr(L^T\Phi_s L)+\frac{\gamma_2}2\tr(L\Phi_t L^T)\\&
+\frac{\rho_1}2\norm{U-L+\widetilde{U}}_F^2+\frac{\rho_2}2\norm{D-L-S-V+\widetilde{V}}_F^2\\
S&\leftarrow \argmin_S\lambda_2\norm{S}_1+\frac{\rho_2}2\norm{D-L-S-V+\widetilde{V}}_F^2\\
U&\leftarrow \argmin_U\lambda_1\norm{U}_{W,*}+\frac{\rho_1}2\norm{U-L+\widetilde{U}}_F^2\\
&=\argmin_U\frac{\lambda_1}{\rho_1}\norm{U}_{W,*}+\frac12\norm{U-L+\widetilde{U}}_F^2\\
V&\leftarrow\argmin_V\norm{V}_1+\frac{\rho_2}2\norm{D-L-S-V+\widetilde{V}}_F^2\\
\widetilde{U}&\leftarrow \widetilde{U}+(U-L)\\
\widetilde{V}&\leftarrow \widetilde{V}+(D-L-S+V)
\end{aligned}\right.
\]
The first $L$-subproblem can be solved by gradient descent.
Note that although the critical equation is a Sylvester equation, the giant matrix $\Phi_s$ will make the standard Sylvester solver very slow. Moreover, conjugate gradient descent can also be applied to solve this subproblem while taking extra steps for computing the adaptive stepsize. Based on our experiments, it is sufficient to apply only a few iterations ($\sim 20$) of gradient descent in order to achieve a good global solution.
The gradient of the objective function reads as
\[\begin{aligned}
\nabla f(L)&=\gamma_1\Phi_sL+\gamma_2 L\Phi_t+\rho_1(L-U-\widetilde{U})\\
&+\rho_2(L+S-D+V-\widetilde{V}).
\end{aligned}
\]
By computing the generalized derivatives for functions defined on matrix spaces, we obtain
\[\begin{aligned}
\frac{d}{dX}\tr(X^TAX)&=(A+A^T)X,\\
\frac{d}{dX}\tr(XAX^T)&=X(A+A^T).
\end{aligned}\]
If $A$ is a symmetric matrix, then $\frac{d}{dX}\tr(X^TAX)=2AX$ and $\frac{d}{dX}\tr(XAX^T)=2XA$.
At each step, we therefore update $L$ with fixed $S,U,\widetilde{U},V,\widetilde{V}$ via
\begin{equation}\label{eqn:Lupdate}
L\leftarrow L-dt\cdot \nabla f(L),
\end{equation}
where $dt>0$ is a step size. It can be empirically shown that only a few steps of gradient descent are sufficient.
Next, the $S$-subproblem has the closed-form solution
\begin{equation}\label{eqn:Supdate}
S\leftarrow \shrink(D-L-V+\widetilde{V}, \lambda_2/\rho_2).
\end{equation}
Here the shrinkage operator is defined as
\[
(\shrink(A,\mu))_{ij}=\mathrm{sign}(a_{ij})\cdot\max\{|a_{ij}|-\mu,0\}
\]
where $a_{ij}$ is the $(i,j)$-th entry of $A$. One can show that the $U$-subproblem has the closed-form solution via a weighted version of the singular value thresholding  (SVT) operator
\begin{equation}\label{eqn:Uupdate}
U\leftarrow A\widetilde{\Sigma}B,\quad \widetilde{\Sigma}=\diag(\shrink(\sigma(\widehat{L}),w_i\lambda_1/\rho_1))
\end{equation}
where $A\Sigma B$ is the singular value decomposition (SVD) form of the matrix $\widehat{L}:=(L-\widetilde{U})$, $\sigma(\widehat{L})$ is the vector containing all the singular values of $\widehat{L}$ with $\sigma_i(\widehat{L})$ as its $i$-th component. Here the weights are constructed iteratively based on the singular values of the matrix $L$ from the previous iteration based on ERF \cite{guo2021novel}:
\begin{equation}\label{eqn:Wupdate}
w_i=\exp(-\sigma_i^2(\widehat{L})/{\sigma^2}).
\end{equation}

Finally, the $V$-subproblem is similar to the $S$-subproblem with the closed-form solution and thereby $V$ is updated via
\begin{equation}\label{eqn:Vupdate}
V\leftarrow \shrink(D-L-S+\widehat{V},1/\rho_2).
\end{equation}
Since the number of frames is usually relatively small compared to the total number of spatial pixels in a video, this step can be quickly done.

As one crucial preprocessing step, we remove motionless frames in the data set if two consecutive frames have small overall changes, i.e., the $\ell_1$-norm of the difference vector of the two adjacent columns of $D$ is below a threshold. The stopping criteria are based on the relative changes in $L$ and $S$, i.e.,
$
\frac{\norm{L^{i+1}-L^i}_F}{\norm{L^i}_F}<tol
$
and
$
\frac{\norm{S^{i+1}-S^i}_F}{\norm{S^i}_F}<tol
$
where $L^i$ and $S^i$ are the obtained background and foreground matrices at the $i$-th iteration and $tol$ is tolerance. Notice that the parameters - $\lambda_2,\,\rho_2$ - in the shrinkage operator can be adaptively updated.
The entire algorithm is summarized in Algorithm~\ref{alg}, which can be extended to handle RGB data sets channelwise. In this work, we focus on gray scale videos by converting all RGB data to gray scale ones.

%\vspace{-4pt}
\begin{algorithm}
\caption{Robust Dual-Graph Regularized Motion Detection}\label{alg}
\begin{algorithmic}
\State\textbf{Inputs}: reshaped video $D\in\R^{n\times m}$, spatial and temporal graph filtering parameters $h_s,h_t>0$, parameters $\lambda_1,\lambda_2,\gamma_1,\gamma_2,\rho_1,\rho_2>0$, step size $dt>0$, maximum number of outer loops $T_{out}$, maximum number of inner loops $T_{in}$, tolerance $tol$
\State\textbf{Outputs}: background $L$ and foreground $S$
\State Generate graph Laplacians $\Phi_t$ and $\Phi_s$
\State Initialize $L$ and $S$
\For{$i=1,2,\ldots,T_{out}$}
        \For{$j=1,2,\ldots,T_{in}$}
        \State Update $L$ via \eqref{eqn:Lupdate}
        \EndFor
        \State Update $S$ via \eqref{eqn:Supdate}
        \State Update $U$ via \eqref{eqn:Uupdate} and singular values $\sigma_i(\widehat{L})$
        \State Update $W$ via \eqref{eqn:Wupdate}
        \State Update $V$ via \eqref{eqn:Vupdate}
        \State $\widetilde{U}\leftarrow \widetilde{U}+(U-L)$
        \State $\widetilde{V}\leftarrow \widetilde{V}+(D-L-S+V)$
        \State {Exit the loop if the stopping criteria are met.}
\EndFor
\end{algorithmic}
\end{algorithm}

\section{Numerical Experiments}\label{sec:exp}
In this section, we will test the proposed Algorithm~\ref{alg} on three real moving object videos. For comparison, we include three closely related algorithms based on the fast robust principal component analysis (RPCA) \cite{candes2011robust}: (1) Largangian optimization method for unconstrained RPCA (LAGO) (2) stable principal component pursuit (SPCP) \cite{driggs2019adapting} and (3) SPGL1 \cite{van2009probing} for solving the problem $\min_{L,S}\max\{\norm{L}_*,\lambda\norm{S}_1\}$ subject to $\norm{D-L-S}_F\leq\varepsilon$. Their source codes are available in fastRPCA \url{https://github.com/stephenbeckr/fastRPCA} \cite{aravkin2014}. There are two groups of metrics for comparing the performance, i.e., comparing the foreground and the background. First, the static background image is extracted from the low-rank component of the given video. We take the mean column of the low-rank matrix $L$ and then reshape it as a matrix. We use the following metrics to evaluate the background recovery quality:
\begin{itemize}
\item relative error (RE): $\mathrm{RE}(\widehat{L},L)=\frac{\norm{L-\widehat{L}}_F}{\norm{L}_F}$;
\item peak signal-to-noise ratio (PSNR): $\mathrm{PSNR}(\widehat{L},L)=20\log({I_{\max}}/\sqrt{\norm{\widehat{L}-L}_F^2/(n_1n_2)})$.
\end{itemize}
Here $\widehat{L}$ is the estimate of the ground truth $L\in\R^{n_1\times n_2}$, and $I_{\max}$ is the maximum image intensity set as 1. In our experiments, all of the videos to be processed are scaled to the range $[0,1]$.

For the foreground assessment, we apply the hard thresholding to extract the foreground masks and then compute the following metrics. Here ground truth foreground masks are manually made. Let $TP$ be the true positive counting the foreground pixels correctly labeled as foreground, $FP$ be the false positive counting the background pixels incorrectly labeled as foreground, and $FN$ be the false negative counting the foreground pixels incorrectly labeled as background. The three metrics are defined as follows.
\begin{itemize}
    \item Precision (Pr): $\mbox{Pr}={TP}/{(TP+FP)}$
    \item Recall (Re):
    $\mbox{Re}={TP}{(TP+FN)}$
    \item F-measure (Fm):
    $\mbox{Fm}=2{\mbox{Re}}/{\mbox{Pr}}$
\end{itemize}
All the three metrics are between 0 and 1. The higher the value is, the more accurate the result is. We also find that various hard thresholding strategies may cause different one or two metrics high while the remaining ones are low.

A Microsoft Azure Kinect Sensor was used to record human motion, including one 1-MP depth sensor, one 7-microphone array, one 12-MP RGB video camera, and one accelerometer and gyroscope (IMU) sensor. Designed to pull together multiple AI sensors in a single device, Azure Kinect sensors have been employed for various applications, such as building telerehabilitation solutions, democratizing home fitness, etc.
In this study, only the RGB video camera was used to record human motion and test the proposed algorithm. All numerical experiments were run in Matlab R2021a on a desktop computer with Intel CPU i9-9960X RAM 64GB and GPU Dual Nvidia Quadro RTX5000 with Windows 10 Pro.

\subsection{Experiment 1: Whole Body Movement Video}
\begin{figure*}[ht]
\centering\setlength{\tabcolsep}{2pt}
\begin{tabular}{ccccc}
\includegraphics[width=0.19\textwidth]{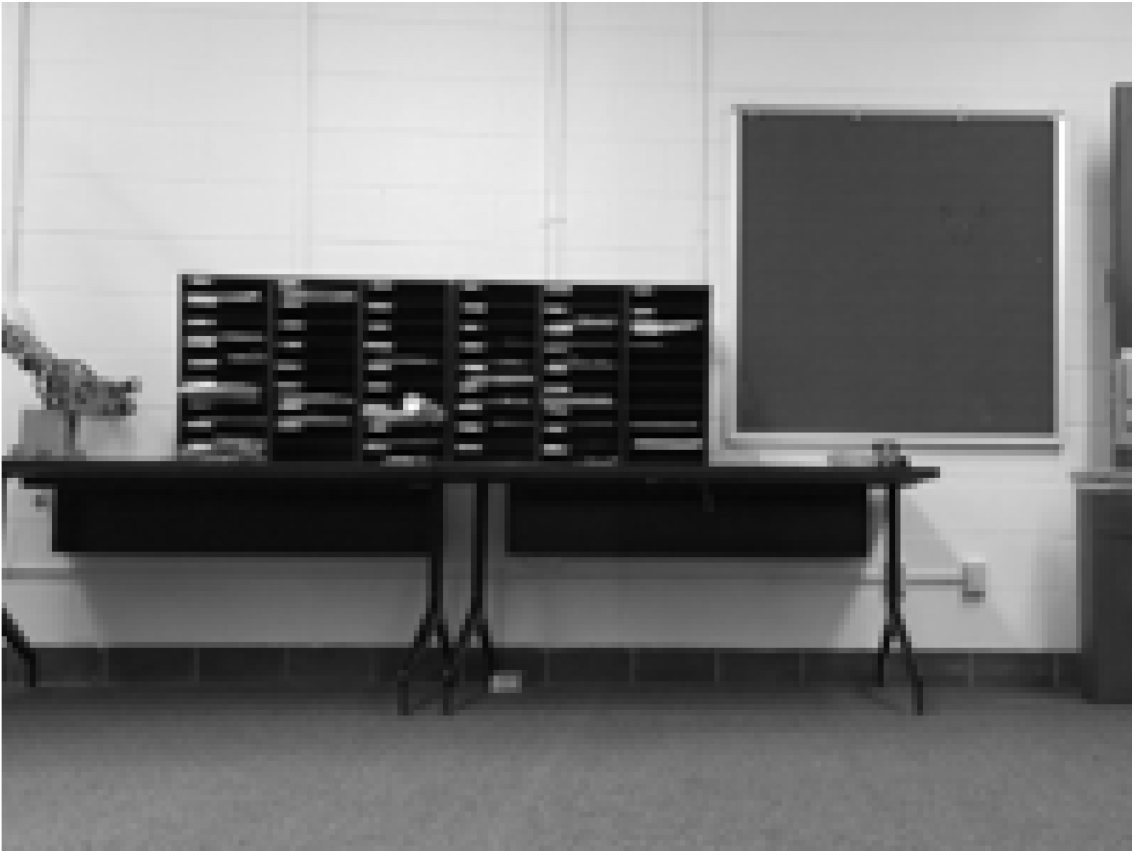}&
\includegraphics[width=0.19\textwidth]{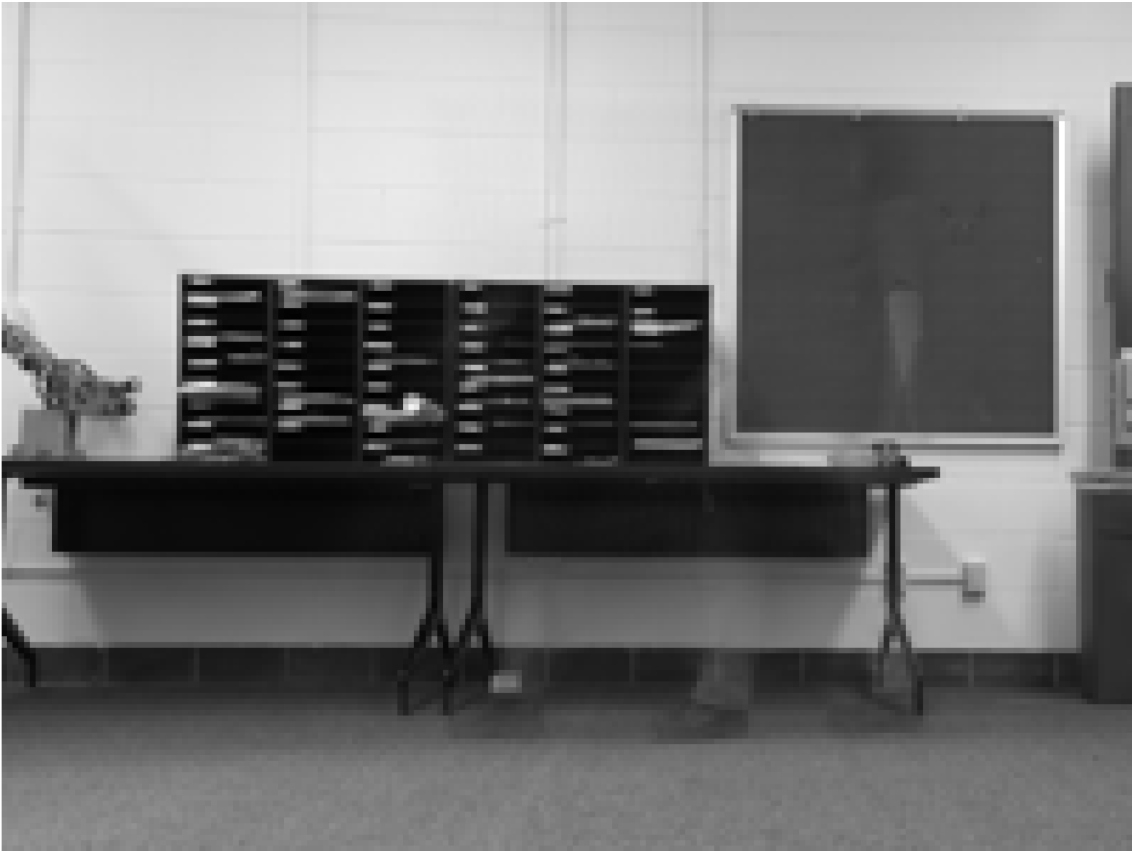}&
\includegraphics[width=0.19\textwidth]{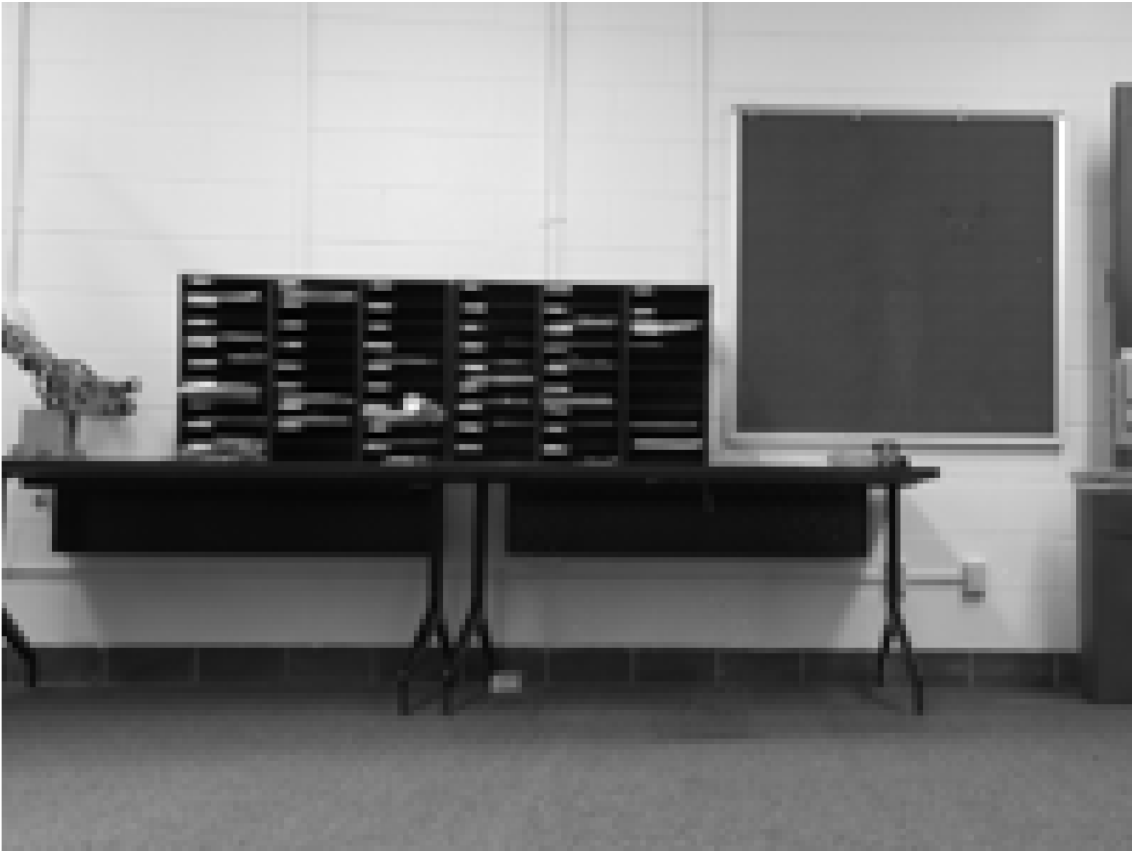}&
\includegraphics[width=0.19\textwidth]{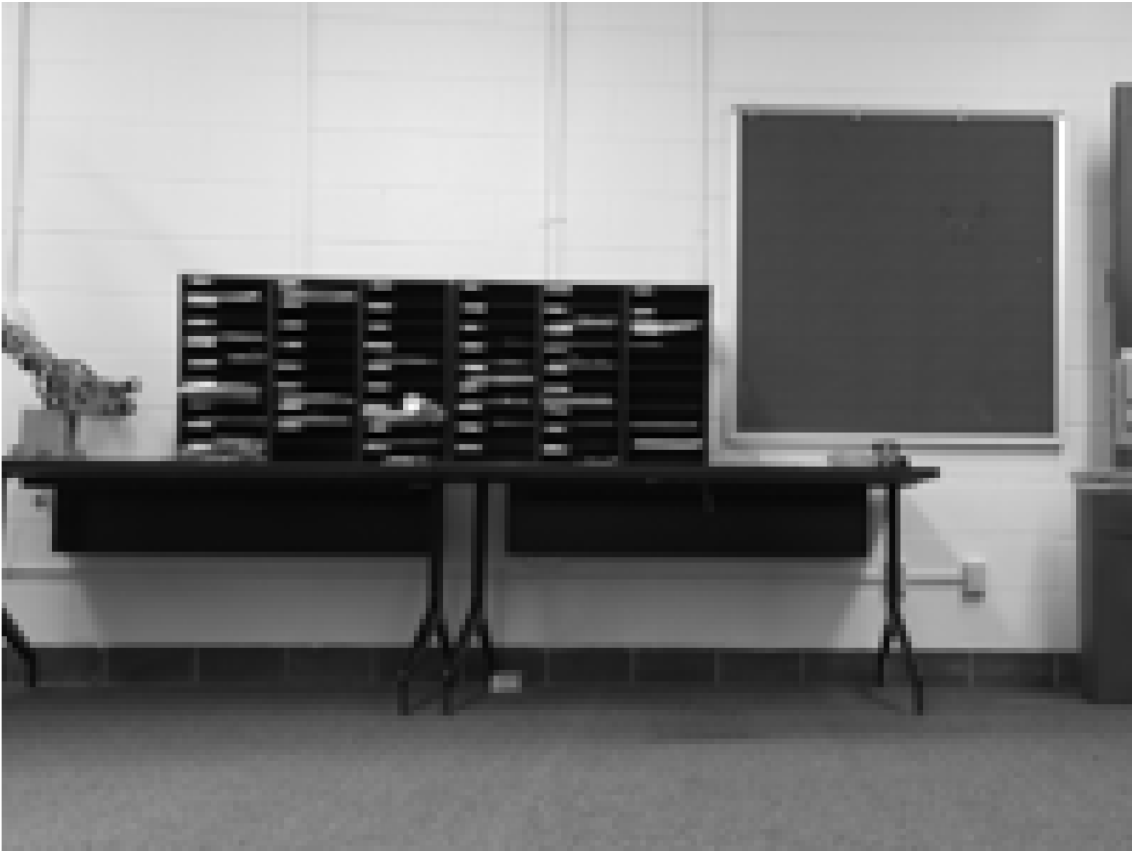}&
\includegraphics[width=0.19\textwidth]{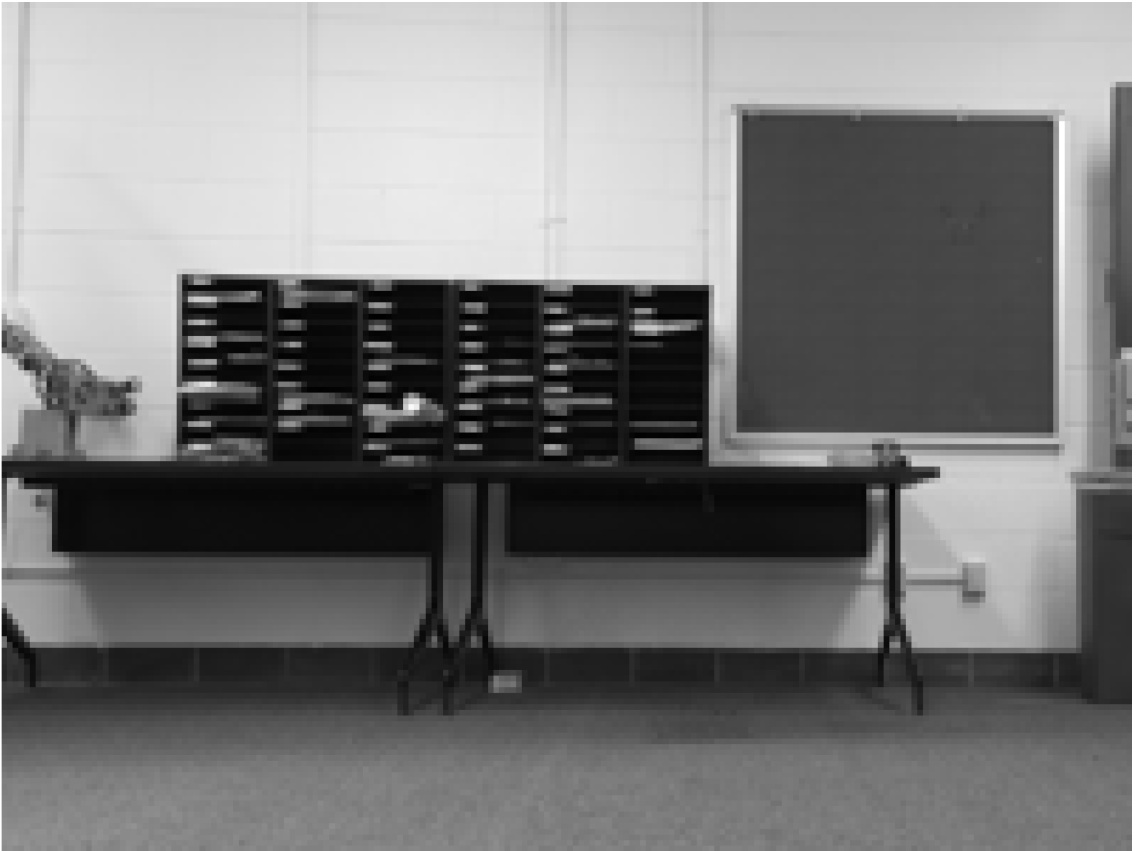}\\
 Ground truth &  LAGO &  SPCP &  SPGL1 &  Alg.1
\end{tabular}
\caption{Recovered backgrounds of the walking video via various methods. }\label{fig:exp1bg}
\end{figure*}

\begin{figure*}[ht]
\centering\setlength{\tabcolsep}{2pt}
\begin{tabular}{ccccc}
\includegraphics[width=0.19\textwidth]{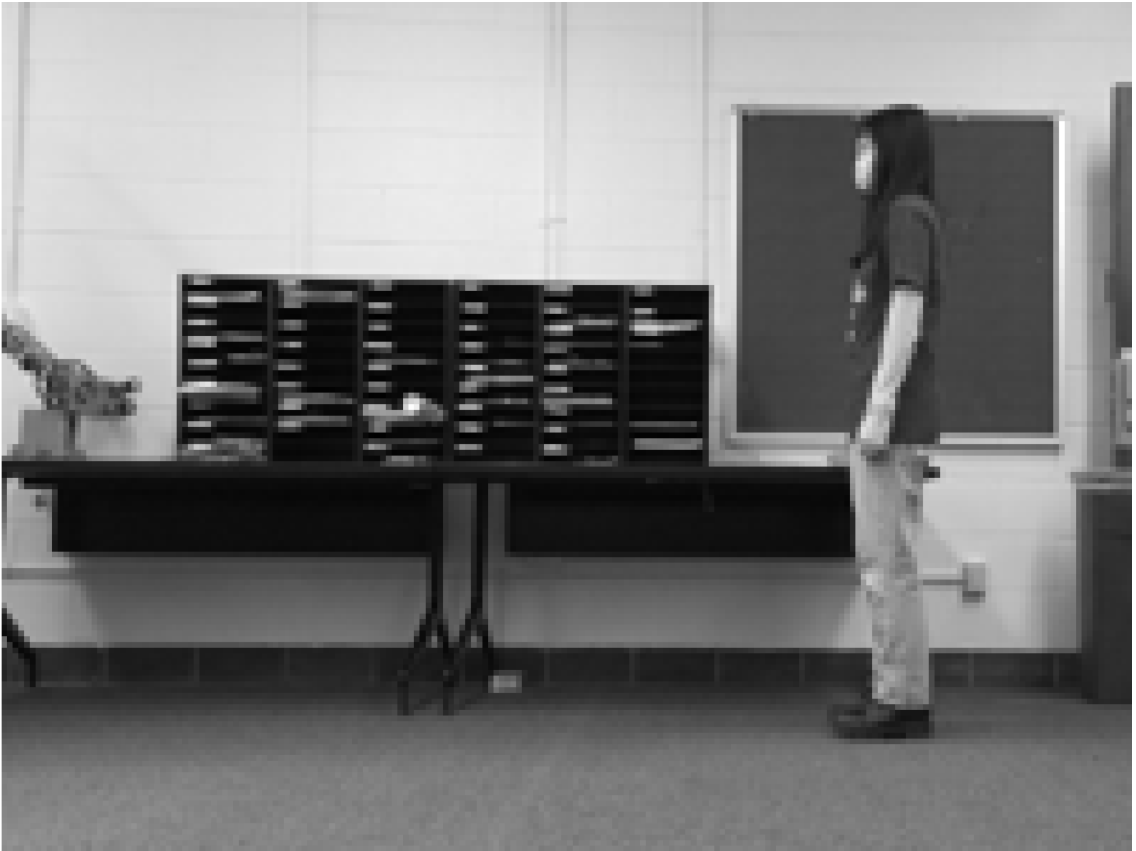}&
\includegraphics[width=0.19\textwidth]{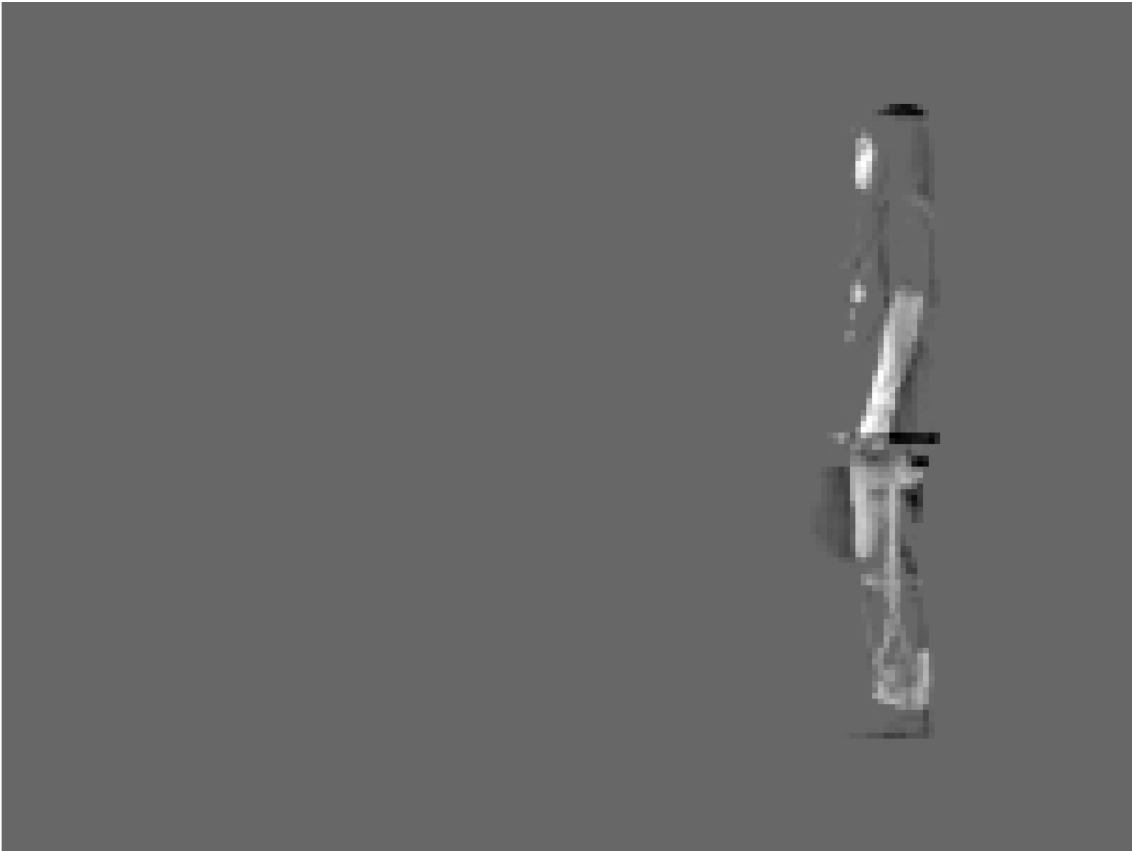}&
\includegraphics[width=0.19\textwidth]{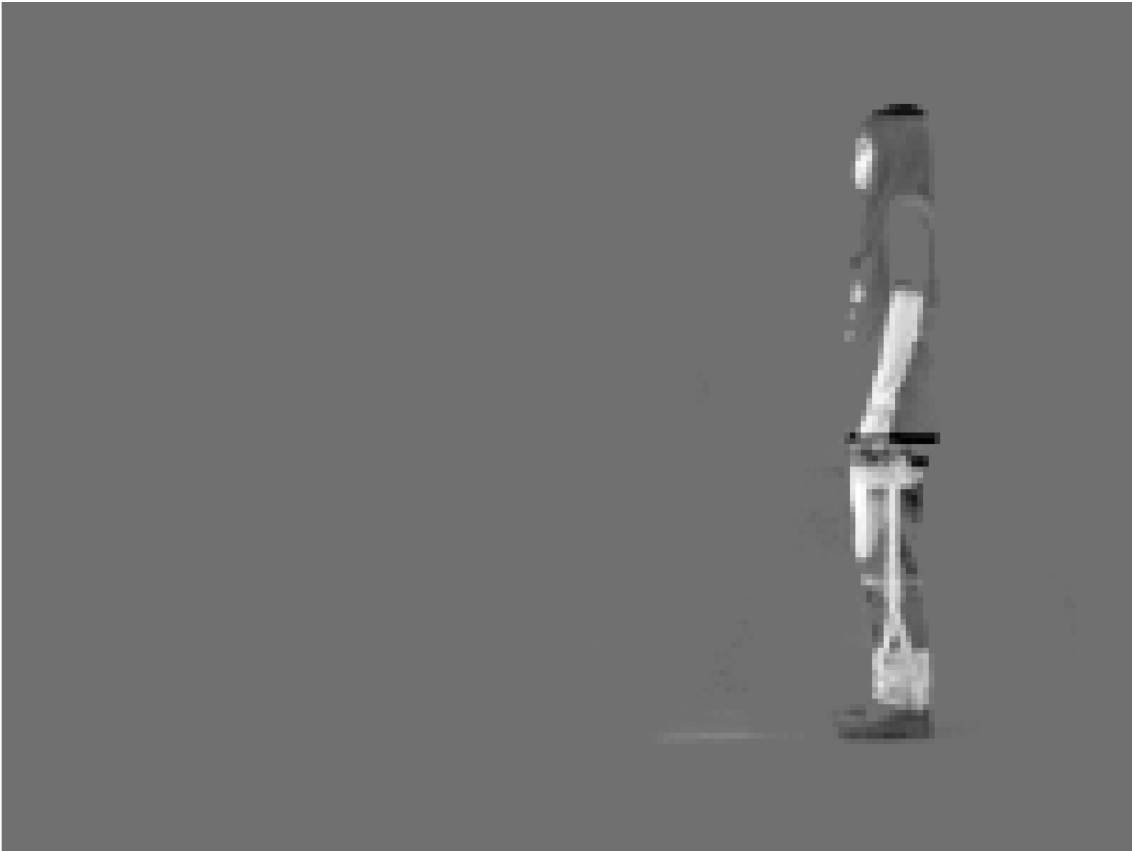}&
\includegraphics[width=0.19\textwidth]{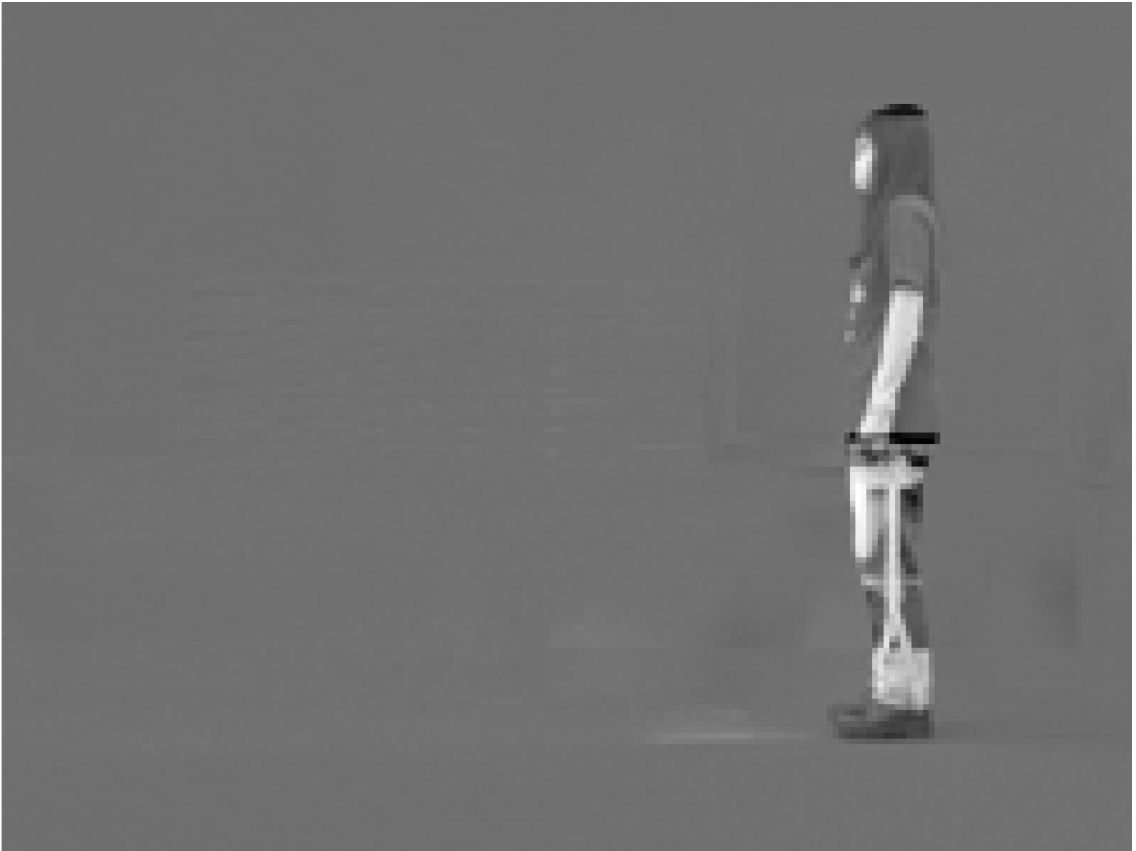}&
\includegraphics[width=0.19\textwidth]{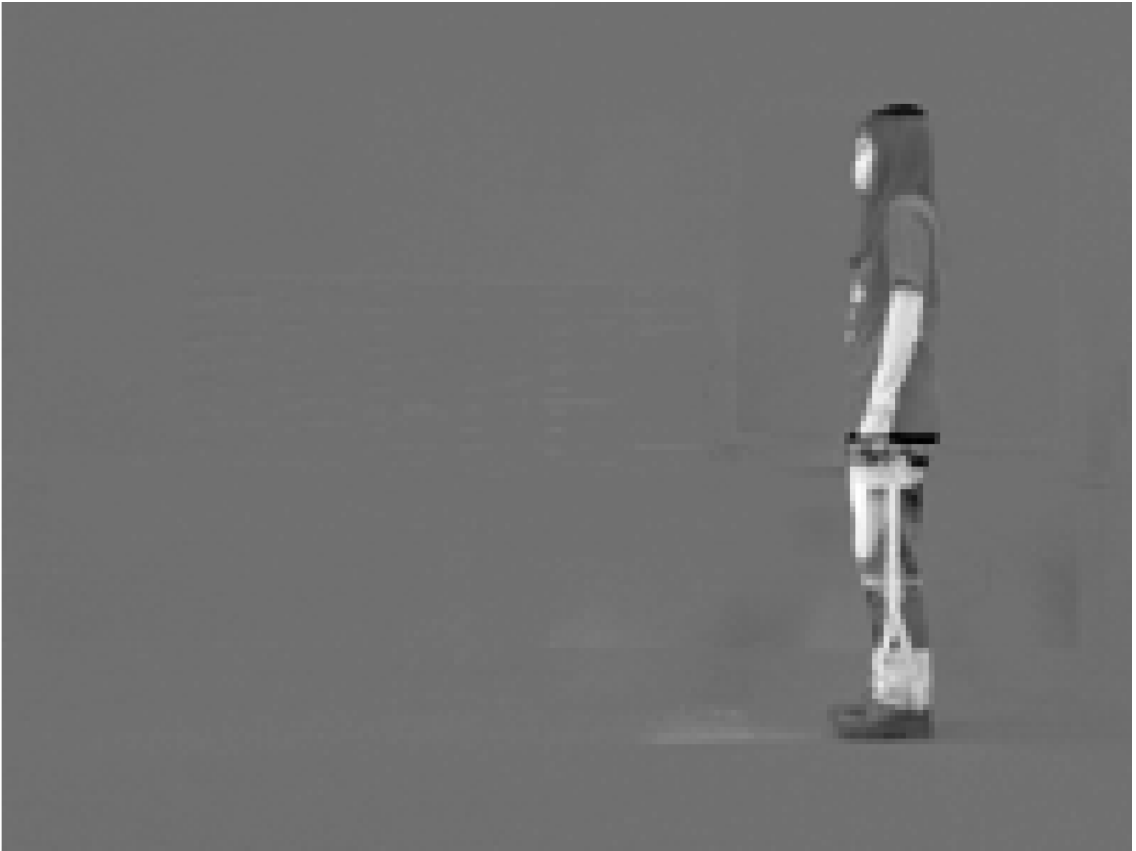}\\
\includegraphics[width=0.19\textwidth]{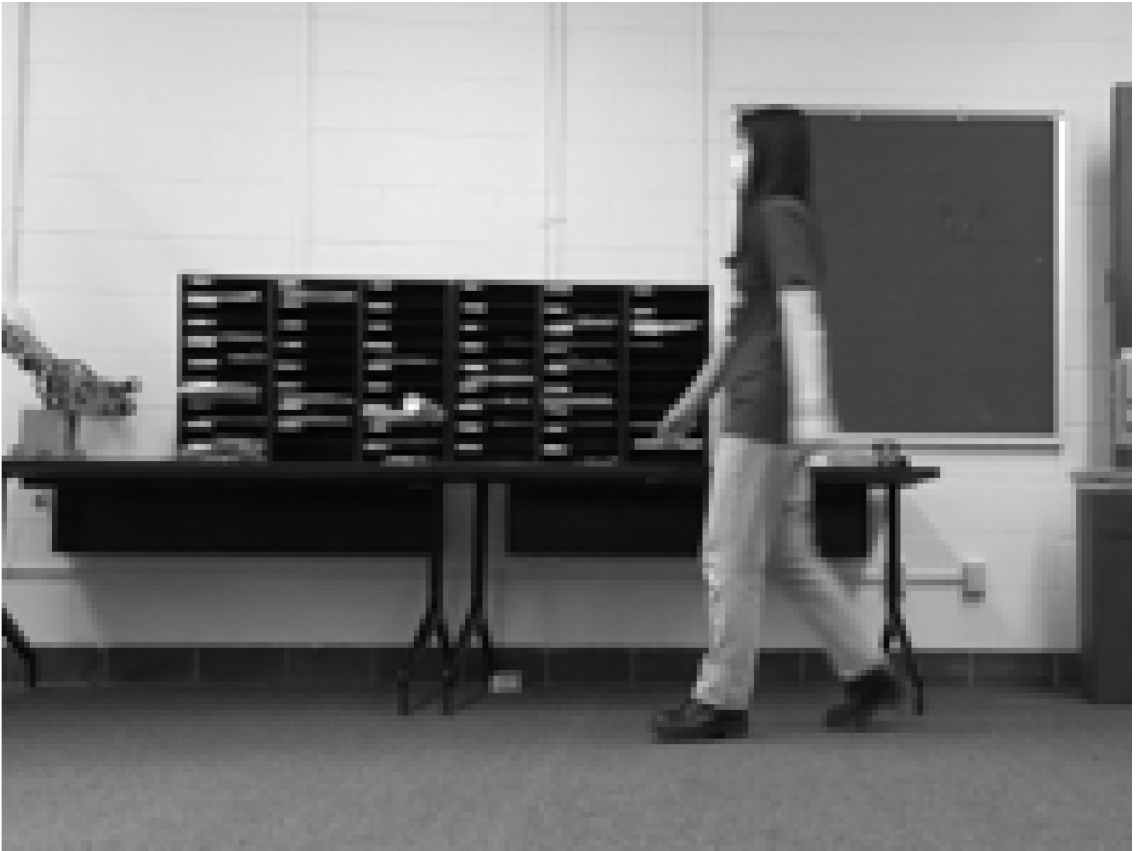}&
\includegraphics[width=0.19\textwidth]{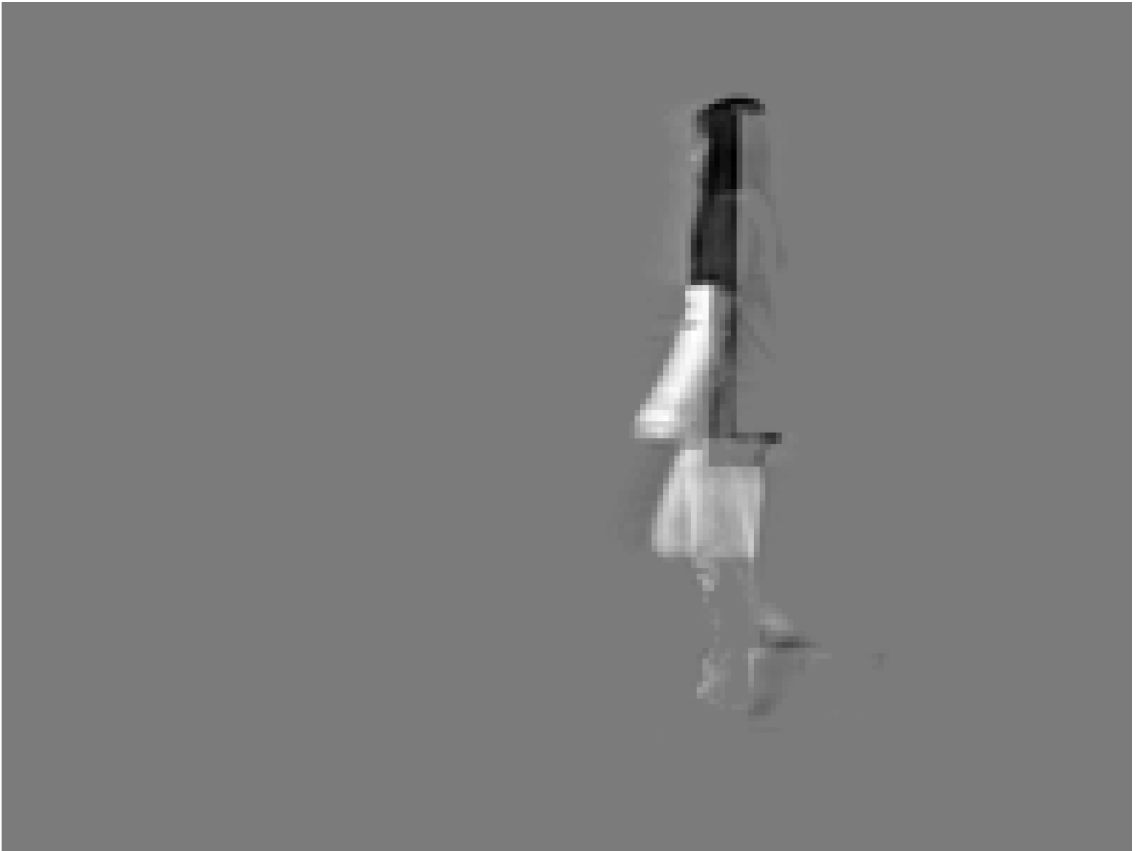}&
\includegraphics[width=0.19\textwidth]{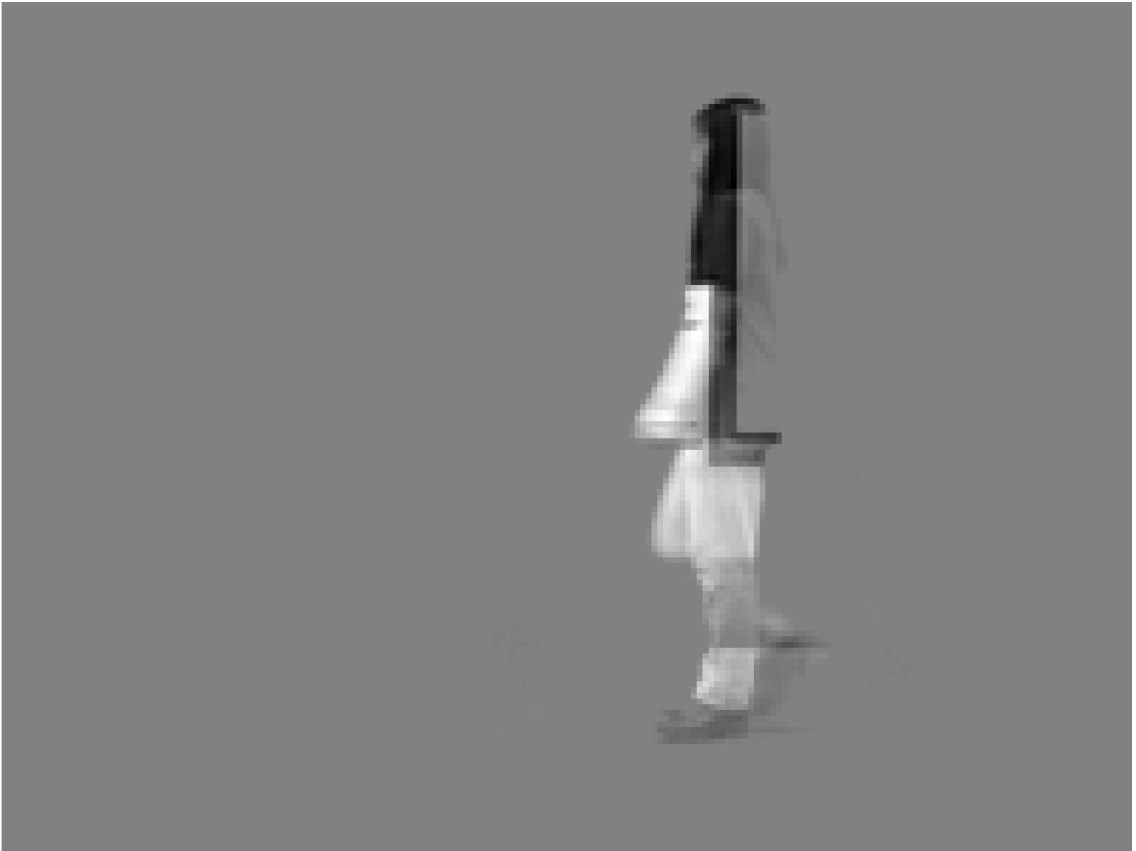}&
\includegraphics[width=0.19\textwidth]{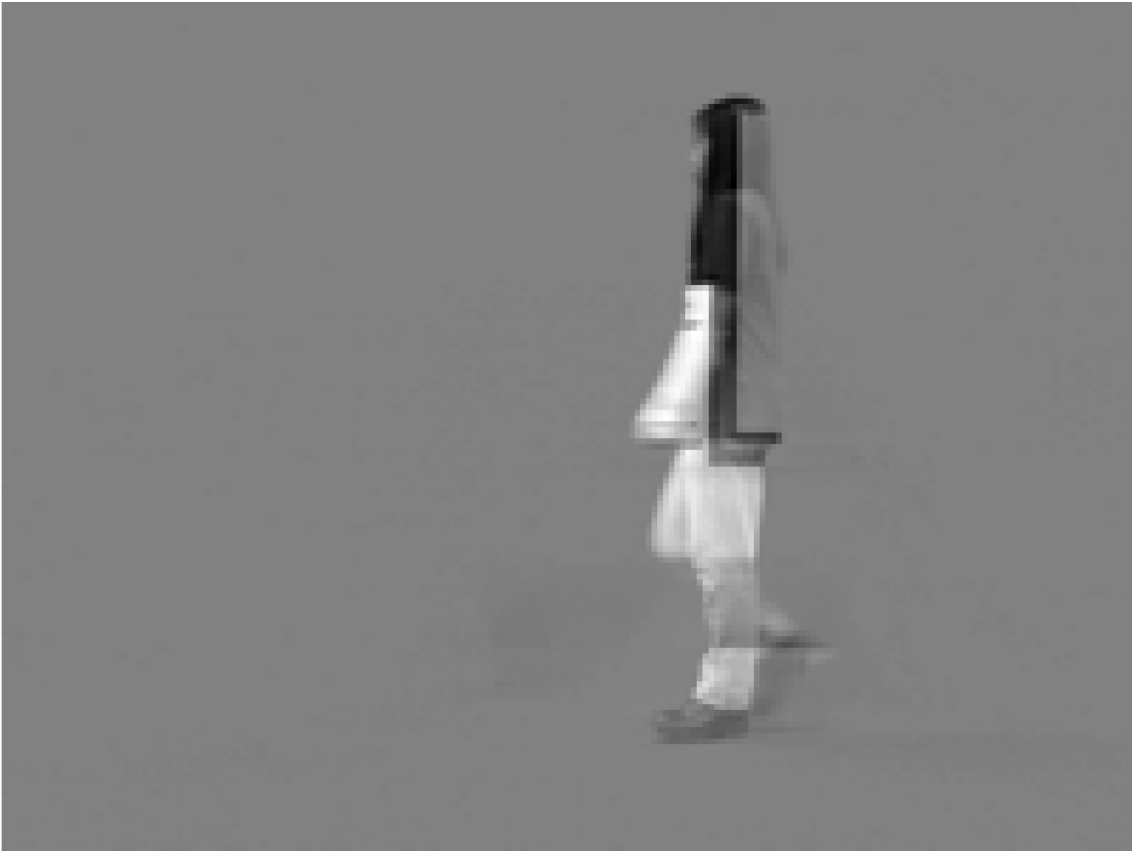}&
\includegraphics[width=0.19\textwidth]{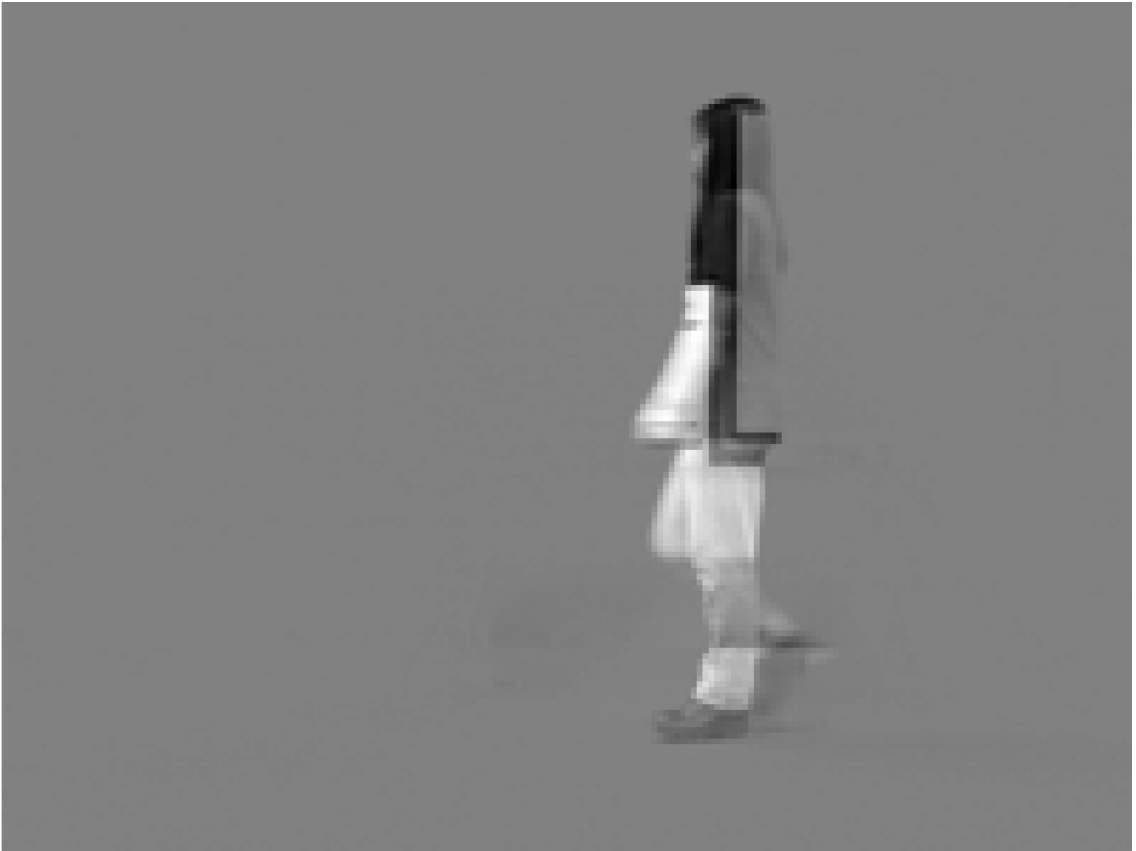}\\
\includegraphics[width=0.19\textwidth]{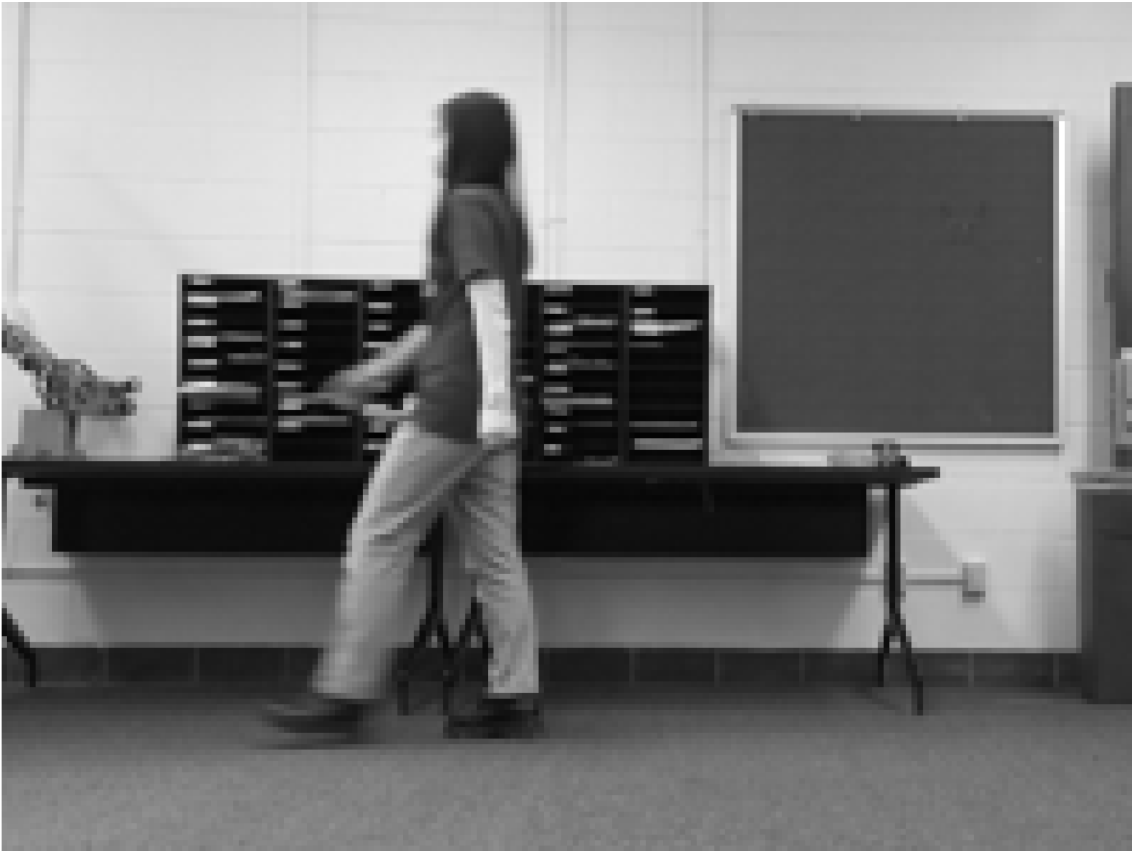}&
\includegraphics[width=0.19\textwidth]{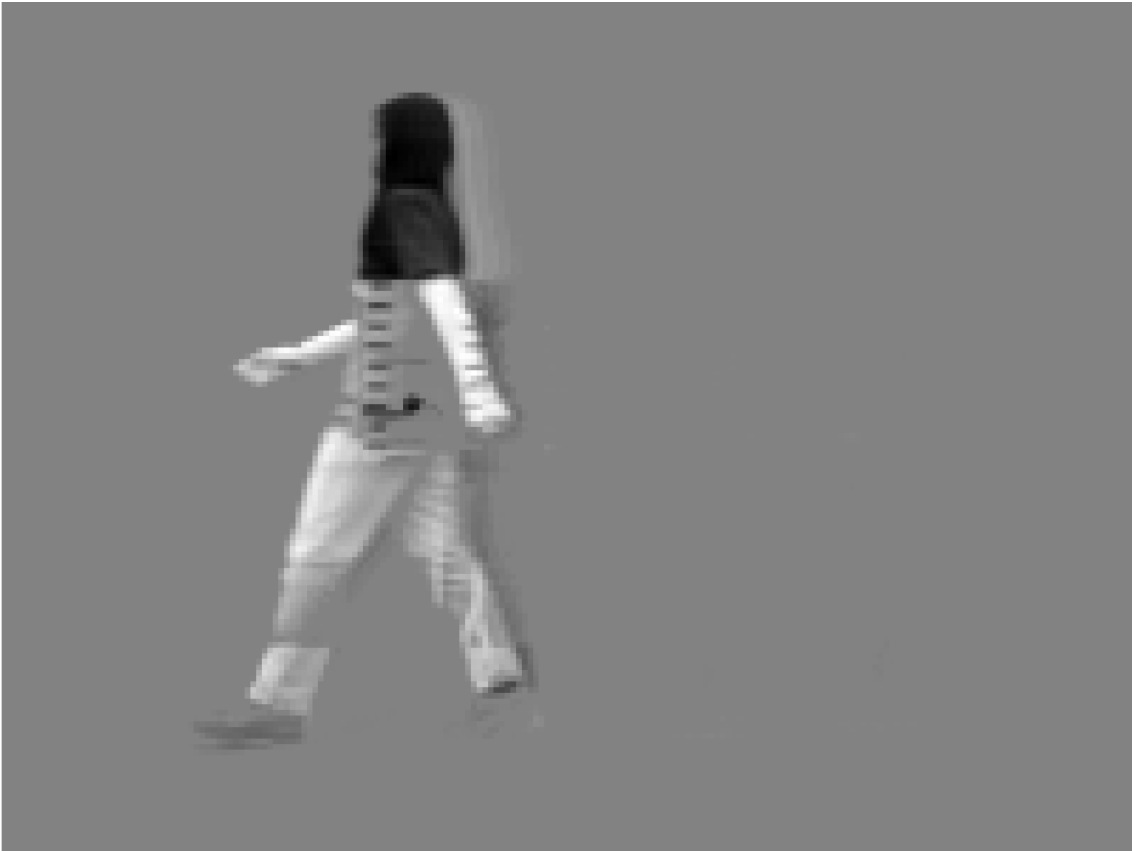}&
\includegraphics[width=0.19\textwidth]{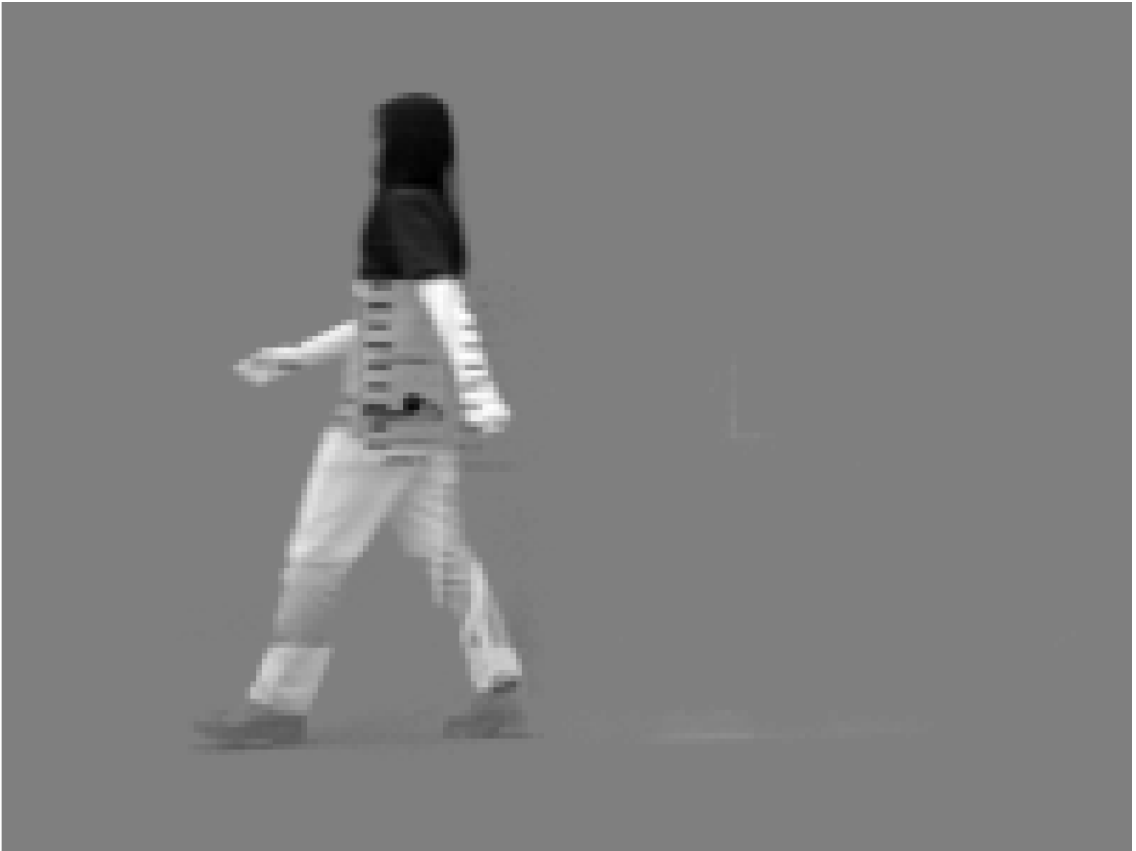}&
\includegraphics[width=0.19\textwidth]{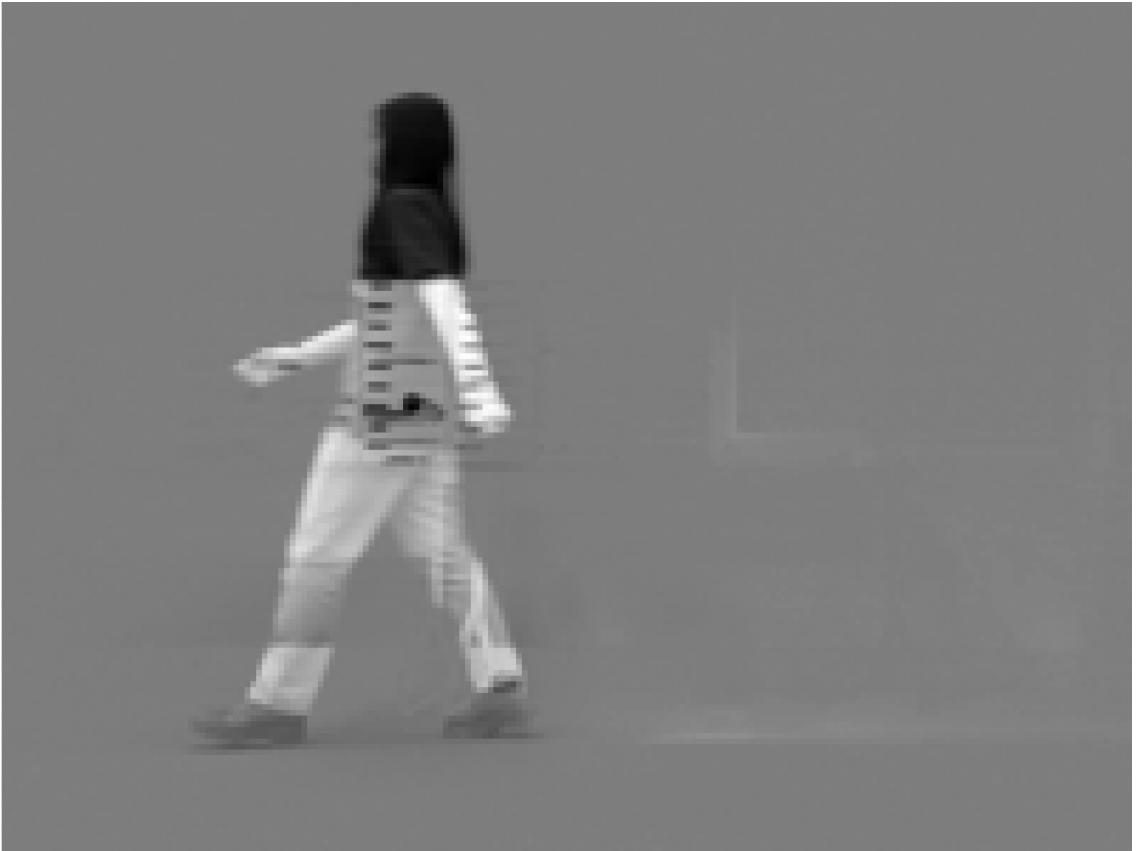}&
\includegraphics[width=0.19\textwidth]{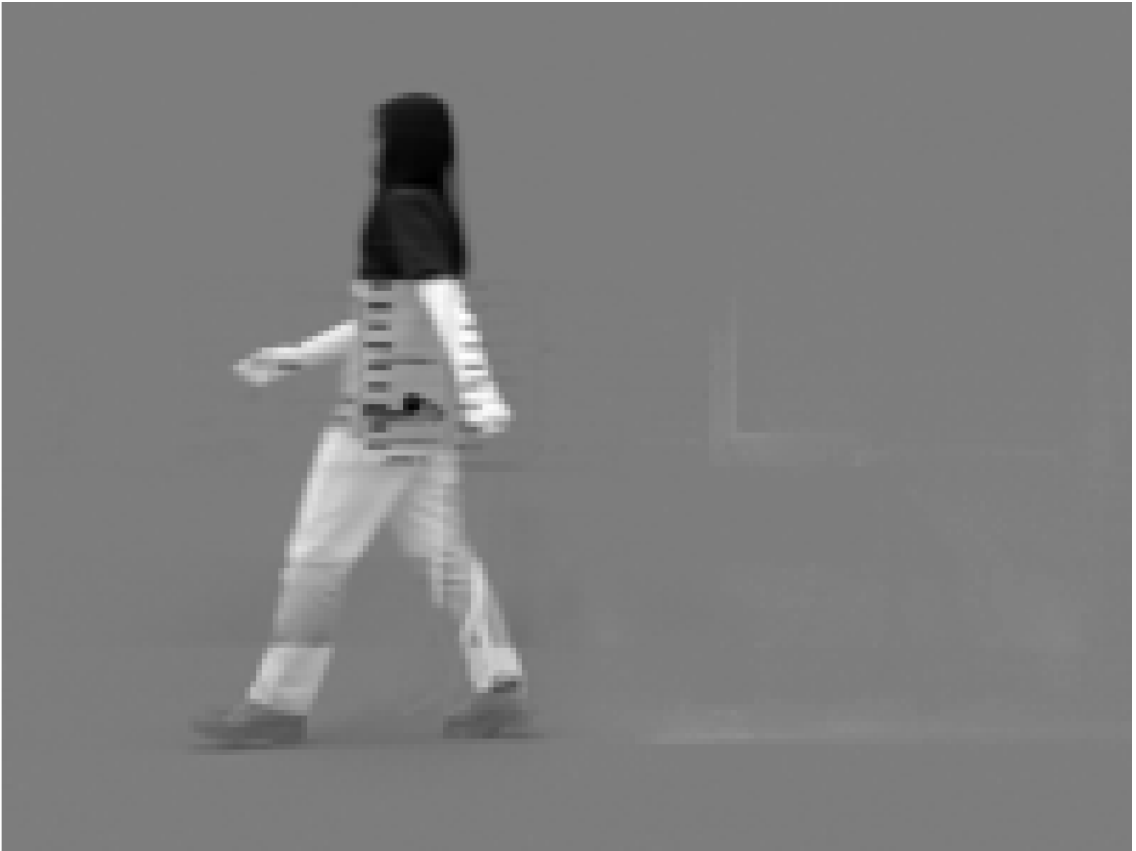}\\
 Original frame &  LAGO &  SPCP &  SPGL1 &  Alg.1
\end{tabular}
\caption{Detected objects for the walking video via various methods. The two rows correspond to the first and the last video frames, respectively.}\label{fig:exp1fg}
\end{figure*}

For the first experiment, we consider a video capturing whole-body movement, which was recorded when one student volunteer was walking naturally at an average speed in a lab room. The video of interest consists of 60 frames where each frame has $150\times 200$ pixels. Due to the limited lighting conditions, there are inevitable shadows of the person and brightness variations in the foreground. In Fig.~\ref{fig:exp1bg}, we compare the recovered background from the various methods. For each method, we take the mean column of the recovered $L$ followed by reshaping it as a matrix, i.e., we use the mean of the obtained backgrounds over 60 frames. There are some white spots in the blackboard mistakenly recognized as foreground in the LAGO result and quite a few still exist in the SPCP result. Both SPGL1 and our results can recover the background well except the light shadow on the ground. In Fig.~\ref{fig:exp1fg}, we show the recovered foregrounds at the first and the last frame. The LAGO result has blurry edges for the human body, and both SPGL1 and our results can detect the shadow motion. The quantitative comparison of the recovered foreground and background for all methods is reported in Table~\ref{tab1}. Our method performs best in terms of all the comparison metrics for this video data.

\begin{table}[ht]
\centering
\caption{Quantitative comparison for the walking video}\label{tab1}
\begin{tabular}{c|cc|ccc}
\hline \hline
    & RE & PSNR & Pr & Re & Fm\\ \hline
LAGO &0.0377&33.67&0.9796&0.4673&0.6328\\
SPCP& 0.0182&39.99&0.9777&0.6354&0.7703 \\
SPGL1 & 0.0148 & 41.81 & 0.9682 & 0.7180&0.8246 \\
Alg.~1 & 0.0145 & 41.95 & 0.9688 & 0.7187&0.8252\\
\hline\hline
\end{tabular}
\end{table}

\begin{figure}
\centering
\includegraphics[width=.45\textwidth]{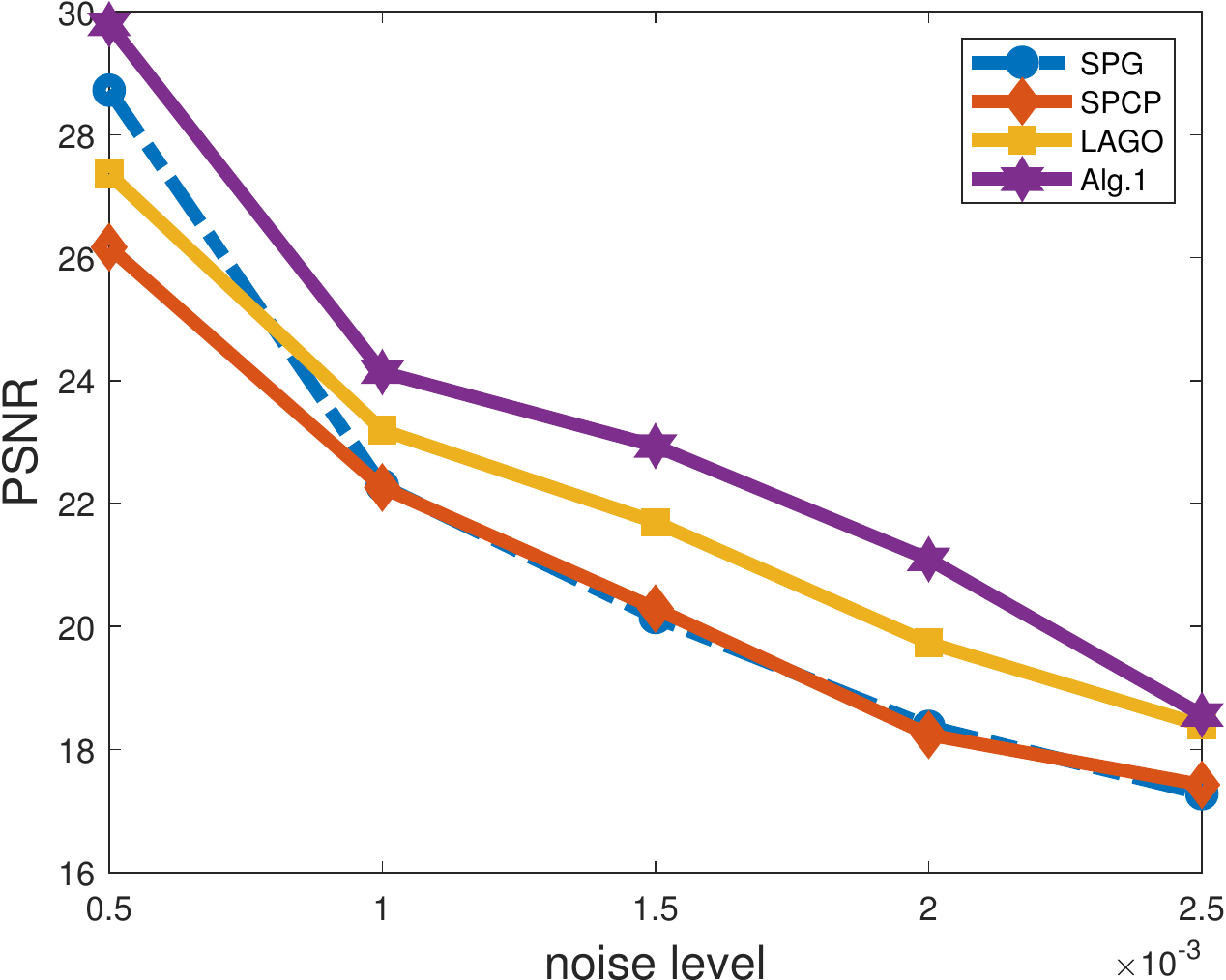}
\caption{PSNR values of recovered backgrounds under various noise levels.}\label{fig:exp1n}
\end{figure}

In addition, we test the performance of all the methods in the case when the given video is degraded by various levels of additive noise. Specifically, we add Gaussian noise with a standard deviation ranging in $0.0005, 0.001, 0.0015, 0.0020, 0.0025$ to the original video that is scaled to $[0,1]$. After choosing the respective optimal parameters, we obtain the PSNR values for various results shown in Fig.~\ref{fig:exp1n}. Note that it becomes more challenging to handle noisy data when the relative errors for the recovered background images are relatively large and also close to each other, so we only show the PSNR plots here. We can see that SPG and SPCP have very similar performance for the high noisy case while our method shows consistent best performance. When the noise level is larger than $0.002$, most of the results except ours have a poor image quality with PSNR less than 20. This indeed shows that the proposed method is more robust to the noise than the other methods.

\subsection{Experiment 2: Arm Movement Video}
\begin{figure*}[ht]
\centering\setlength{\tabcolsep}{2pt}
\begin{tabular}{ccccc}
\includegraphics[width=0.19\textwidth]{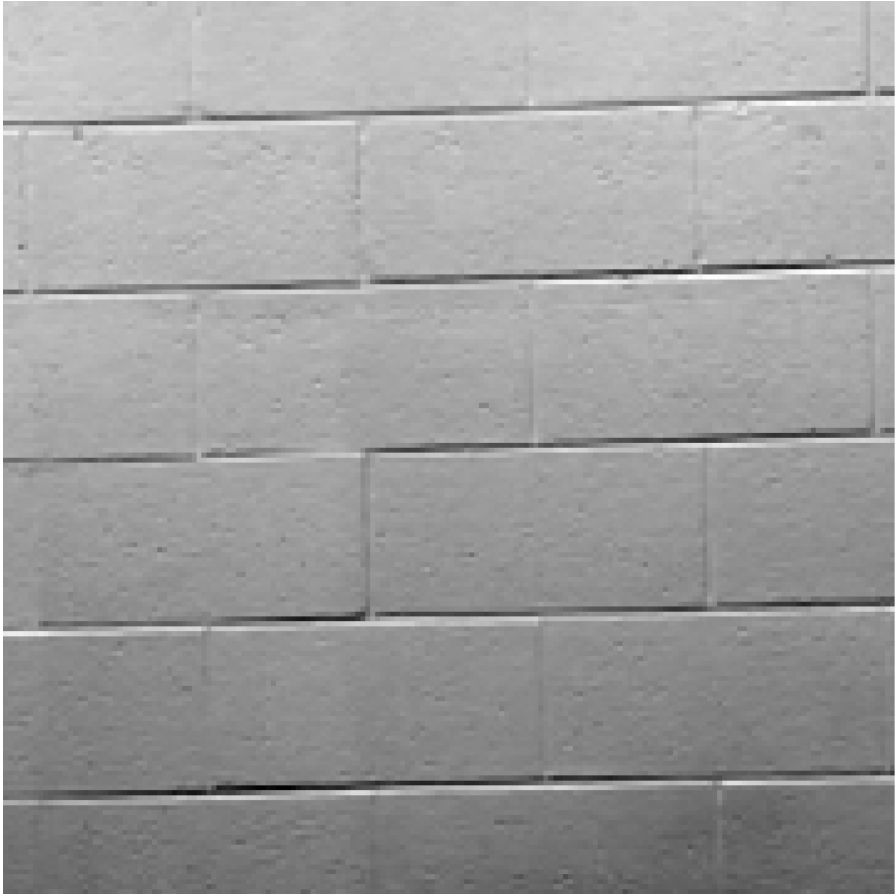}&
\includegraphics[width=0.19\textwidth]{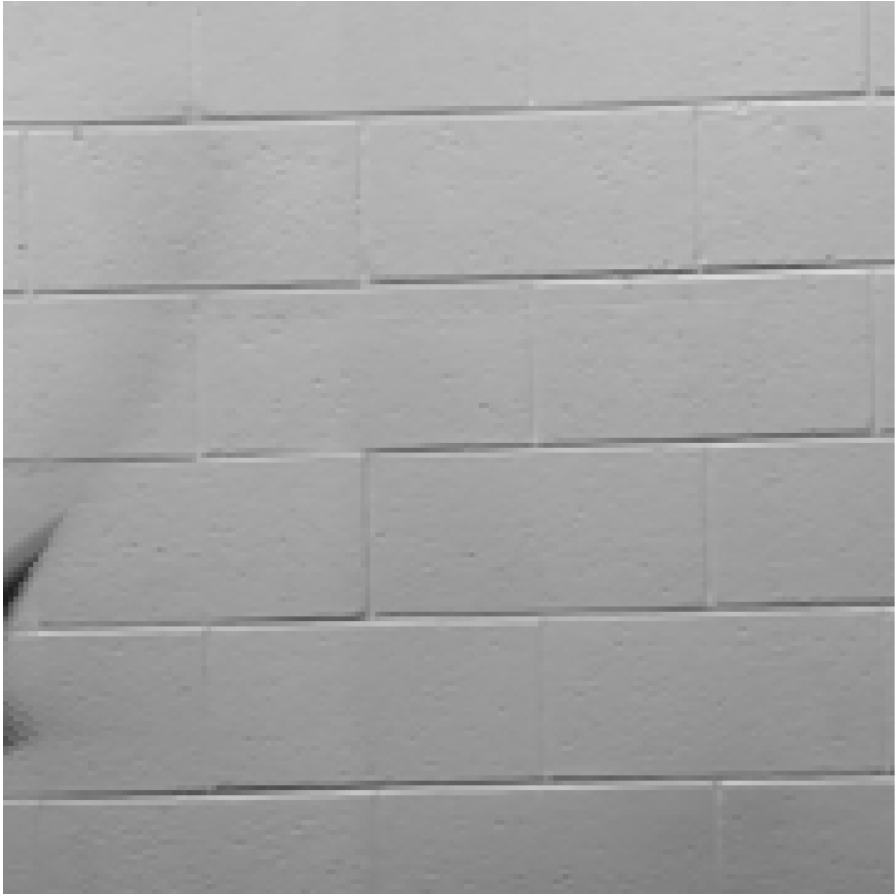}&
\includegraphics[width=0.19\textwidth]{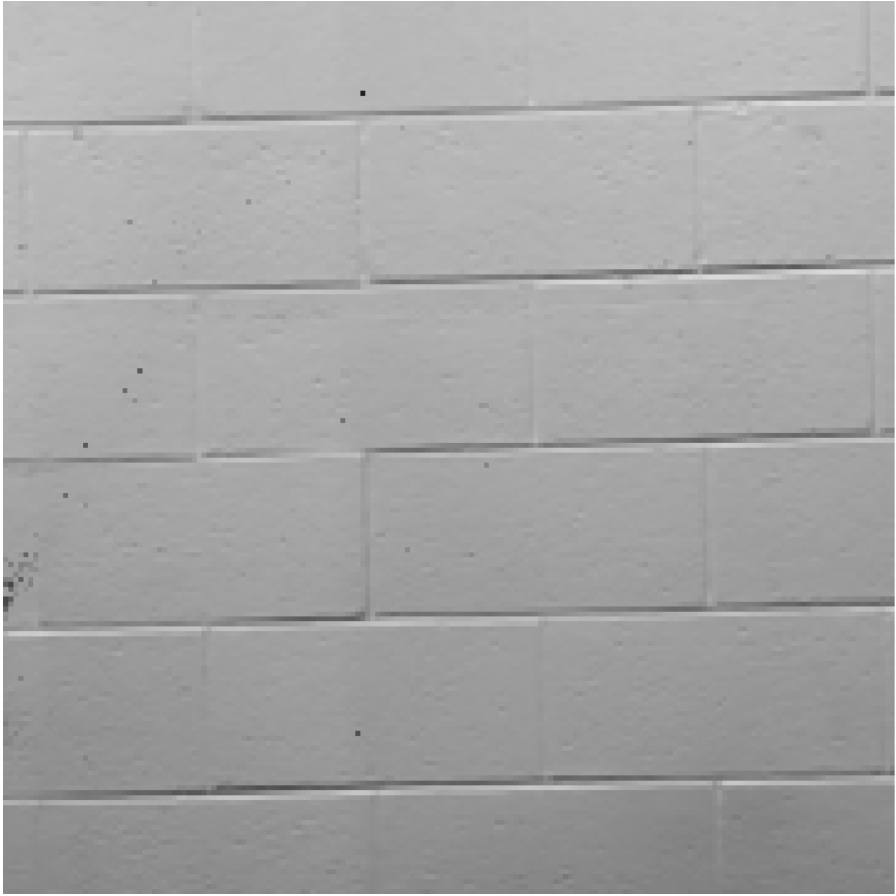}&
\includegraphics[width=0.19\textwidth]{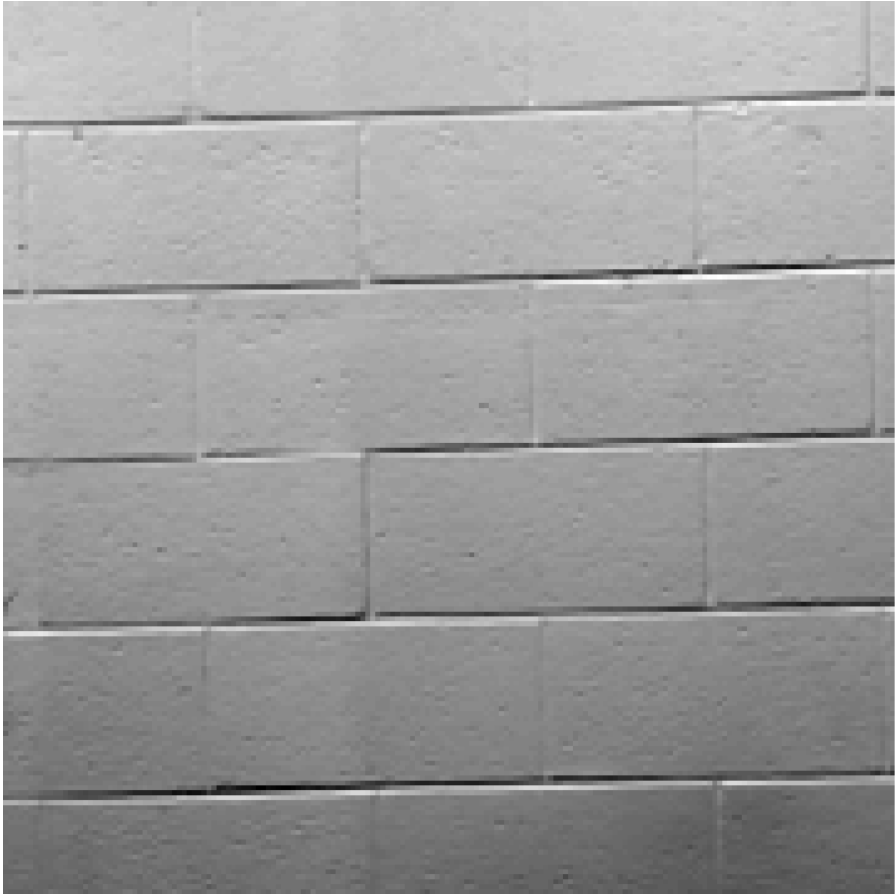}&
\includegraphics[width=0.19\textwidth]{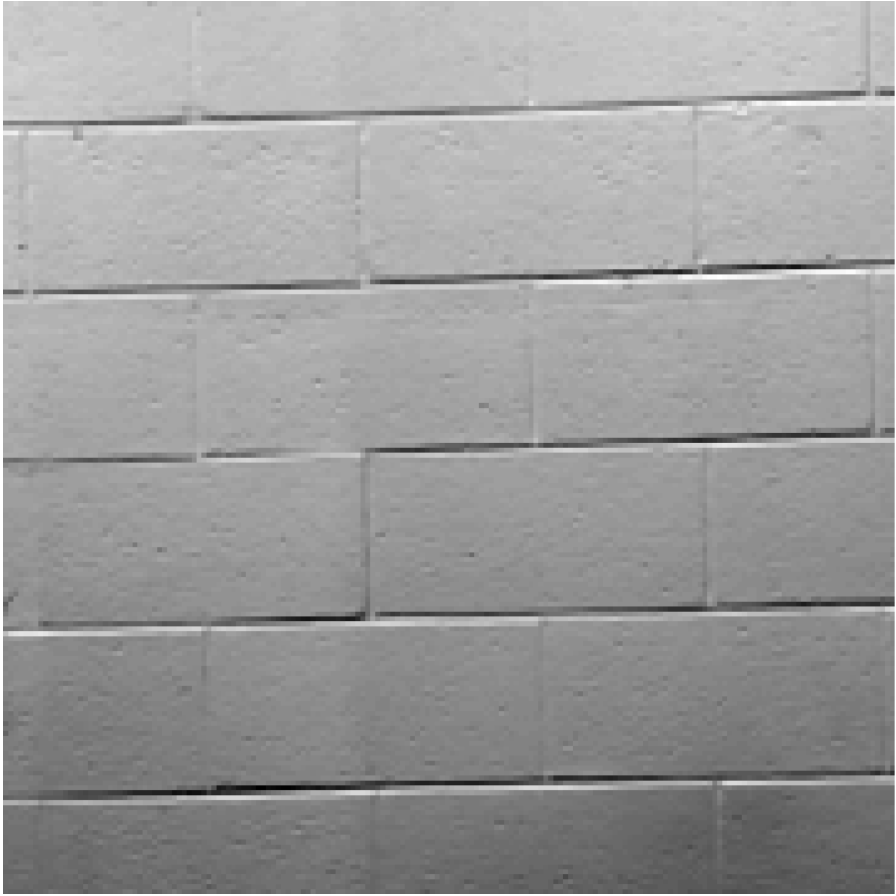}\\
 Ground truth &  LAGO &  SPCP &  SPGL1 &  Alg.1
\end{tabular}
\caption{Recovered backgrounds of the arm motion video via various methods. }\label{fig:exp2bg}
\end{figure*}

\begin{figure*}[ht]
\centering\setlength{\tabcolsep}{2pt}
\begin{tabular}{ccccc}
\includegraphics[width=0.19\textwidth]{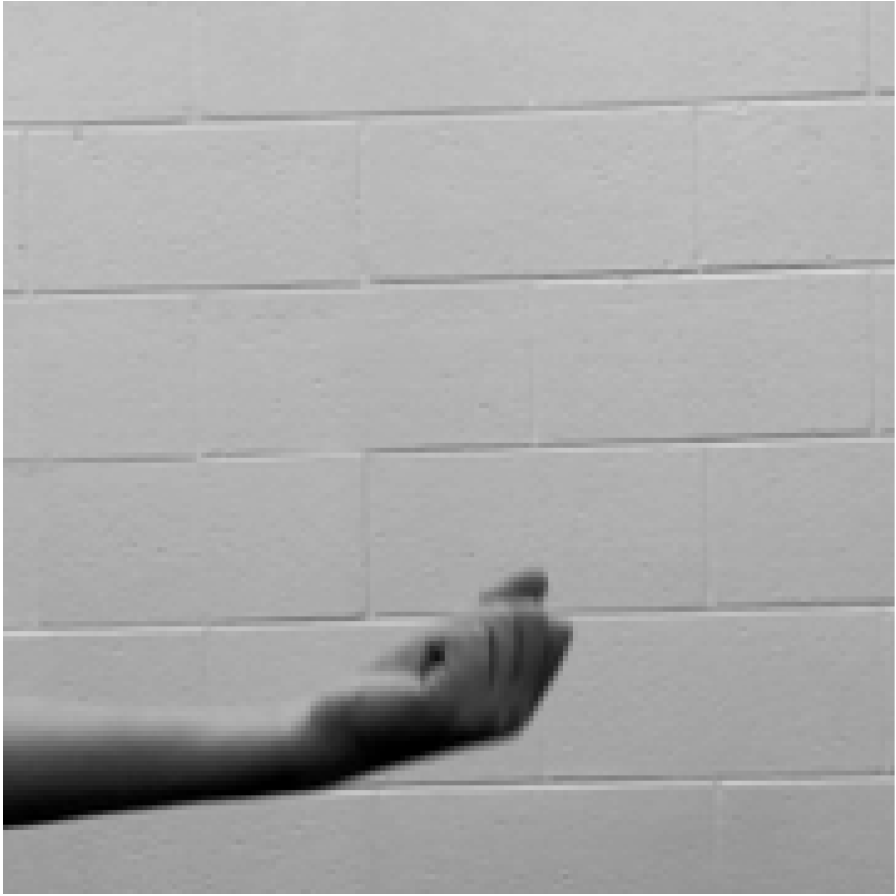}&
\includegraphics[width=0.19\textwidth]{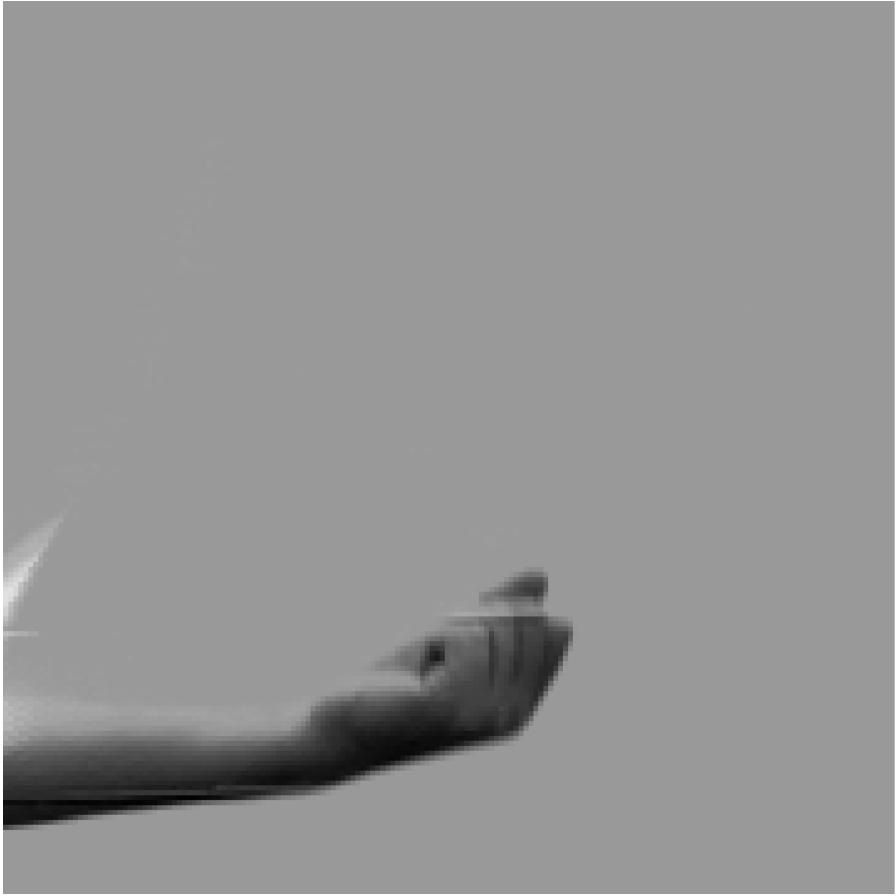}&
\includegraphics[width=0.19\textwidth]{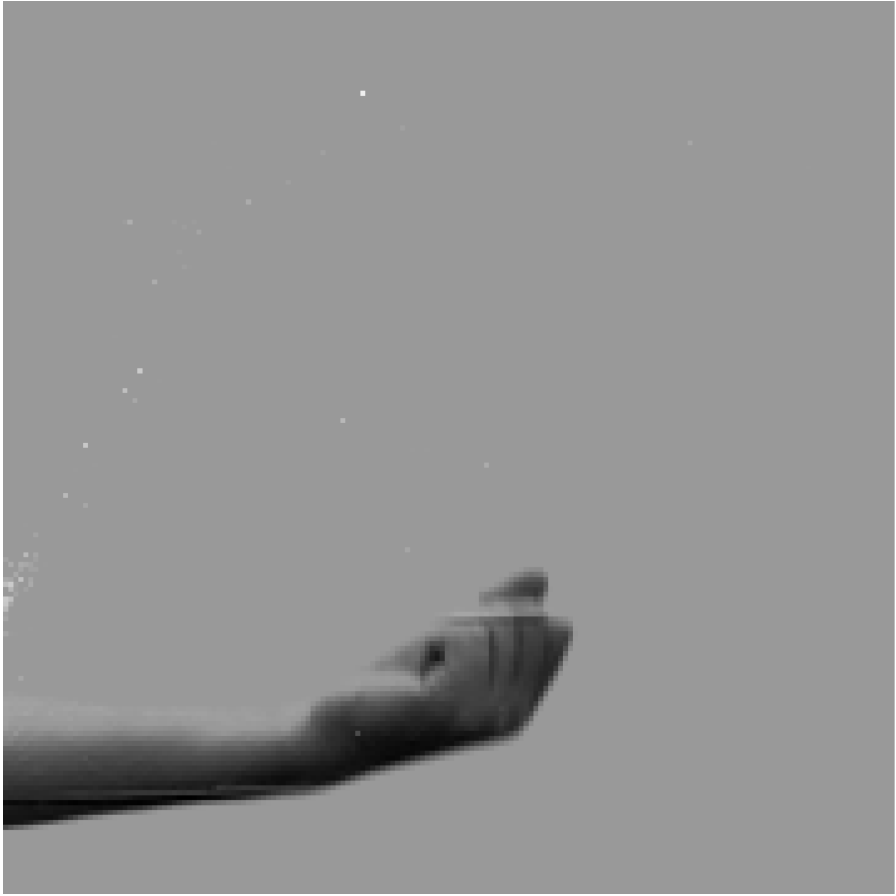}&
\includegraphics[width=0.19\textwidth]{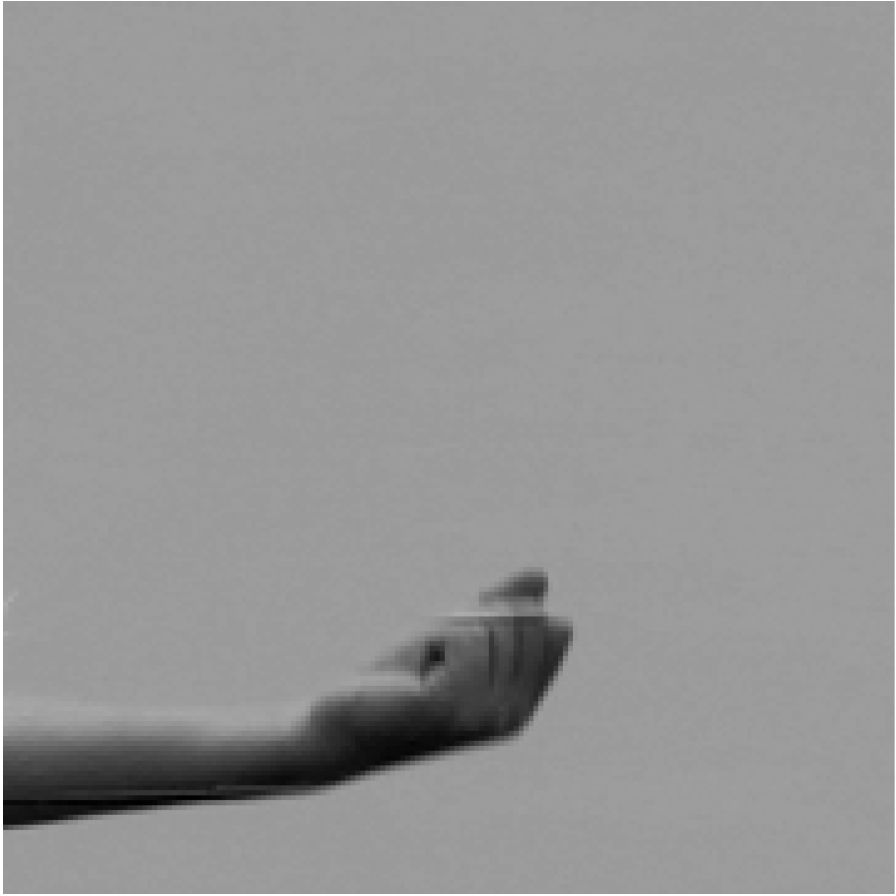}&
\includegraphics[width=0.19\textwidth]{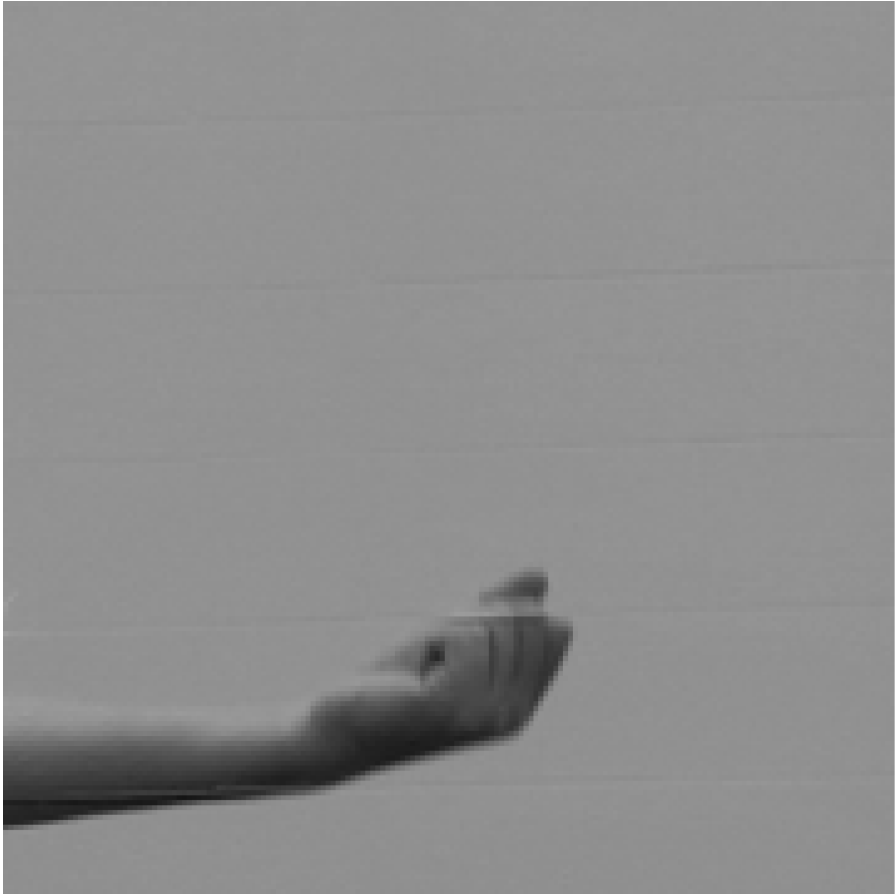}\\
\includegraphics[width=0.19\textwidth]{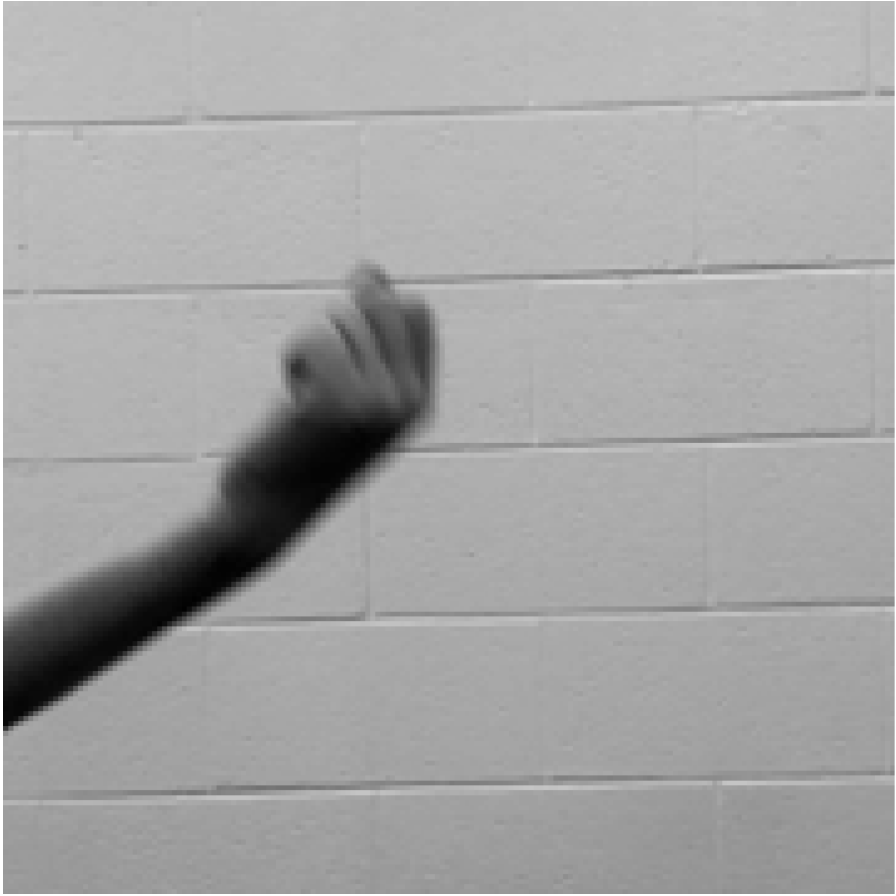}&
\includegraphics[width=0.19\textwidth]{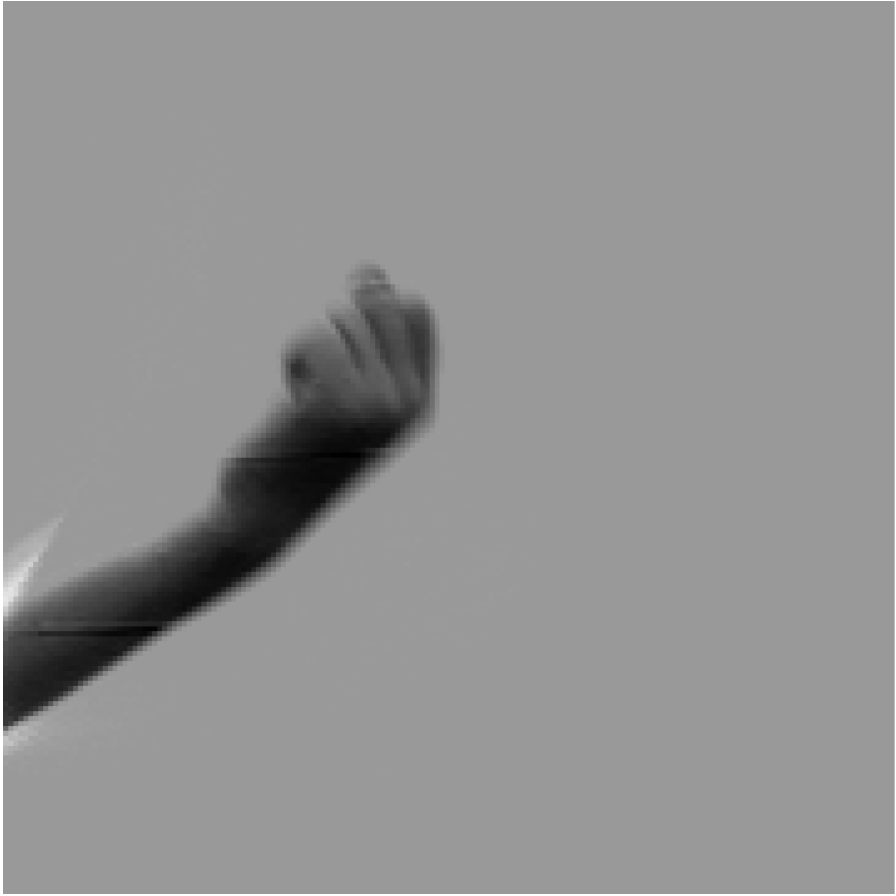}&
\includegraphics[width=0.19\textwidth]{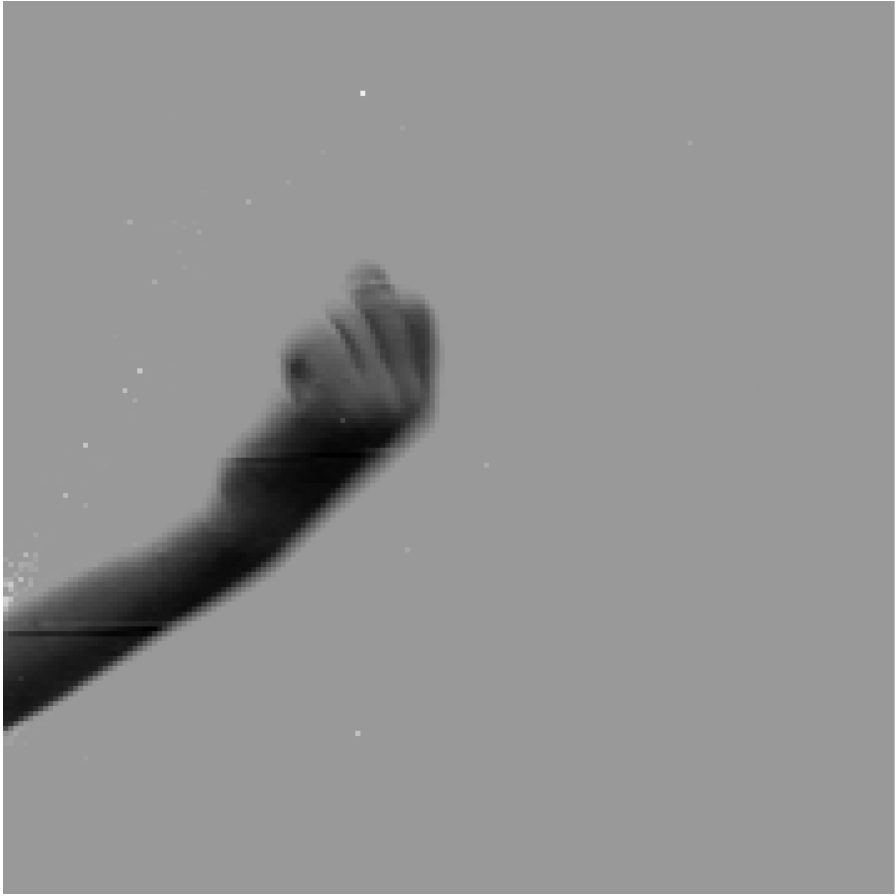}&
\includegraphics[width=0.19\textwidth]{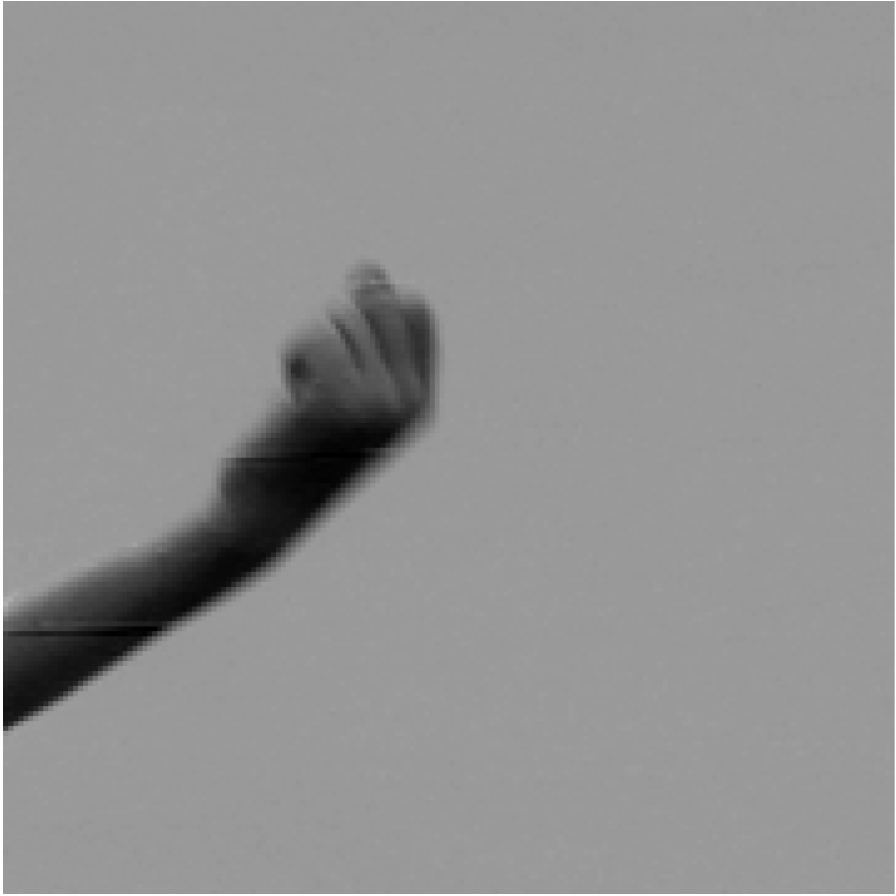}&
\includegraphics[width=0.19\textwidth]{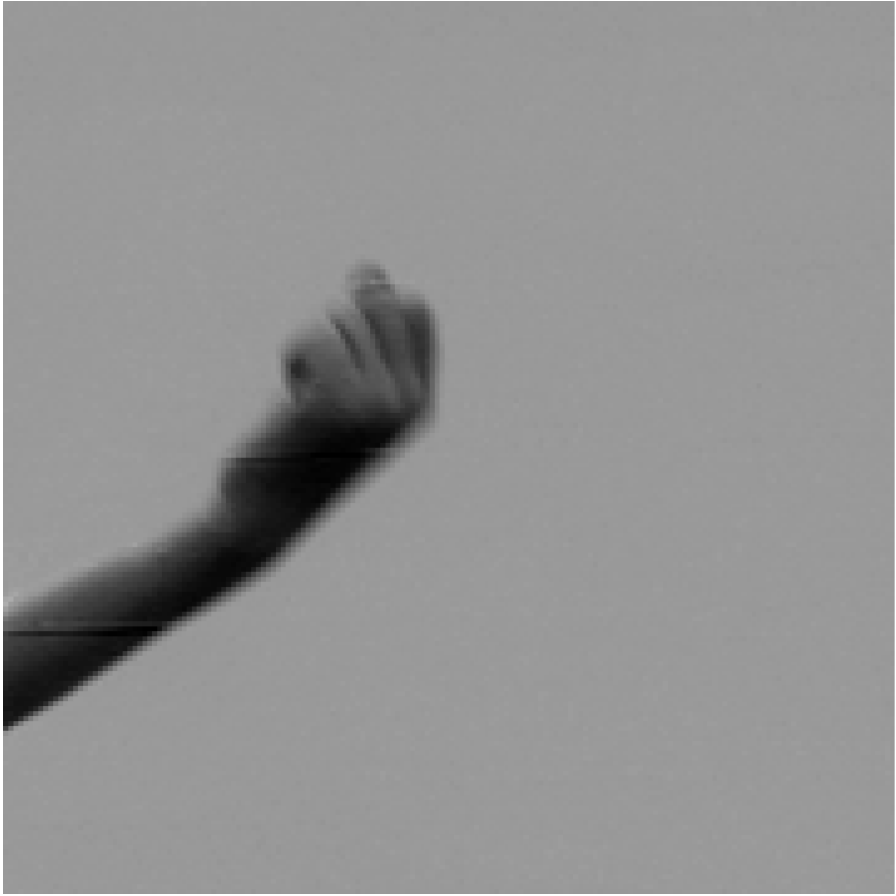}\\
\includegraphics[width=0.19\textwidth]{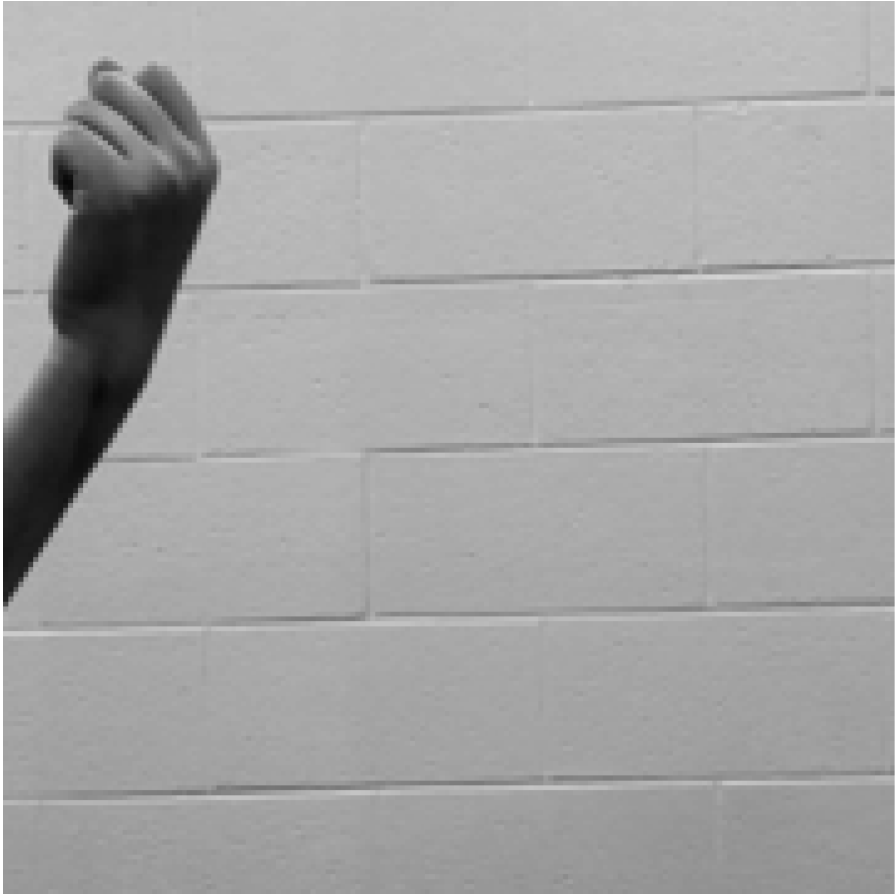}&
\includegraphics[width=0.19\textwidth]{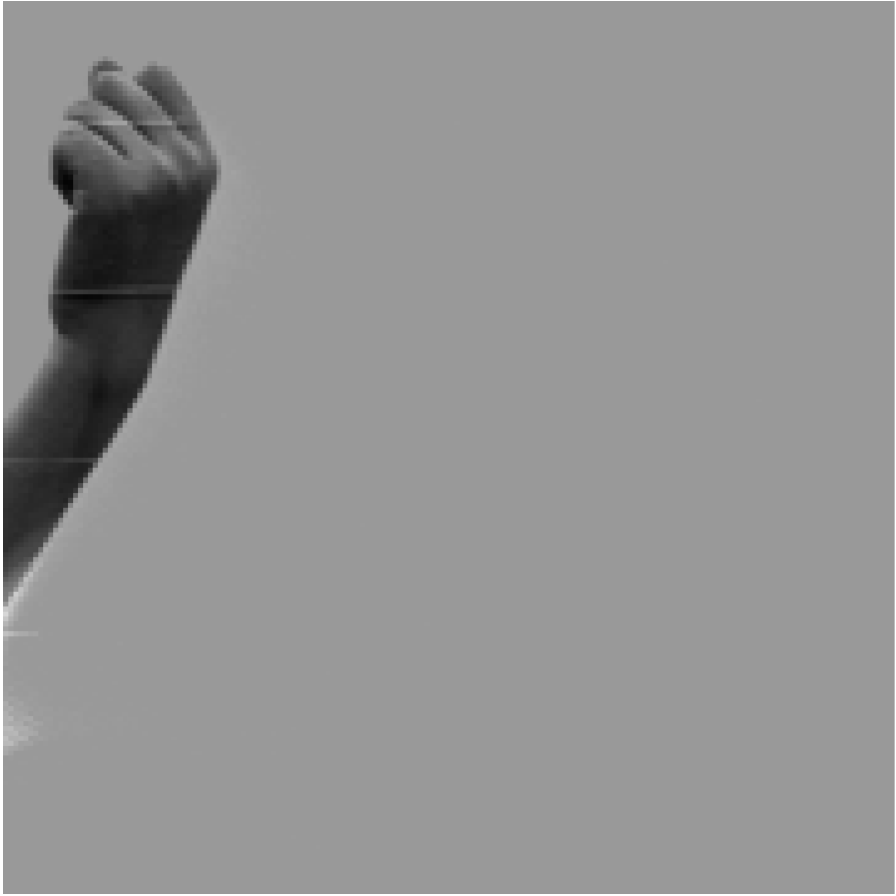}&
\includegraphics[width=0.19\textwidth]{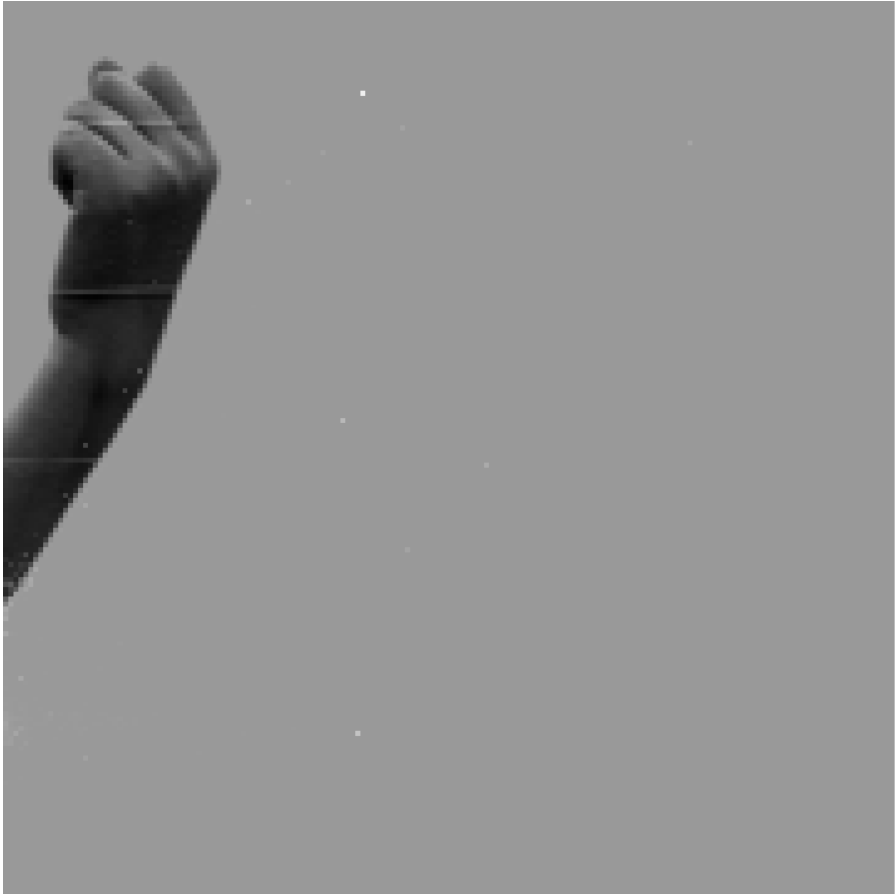}&
\includegraphics[width=0.19\textwidth]{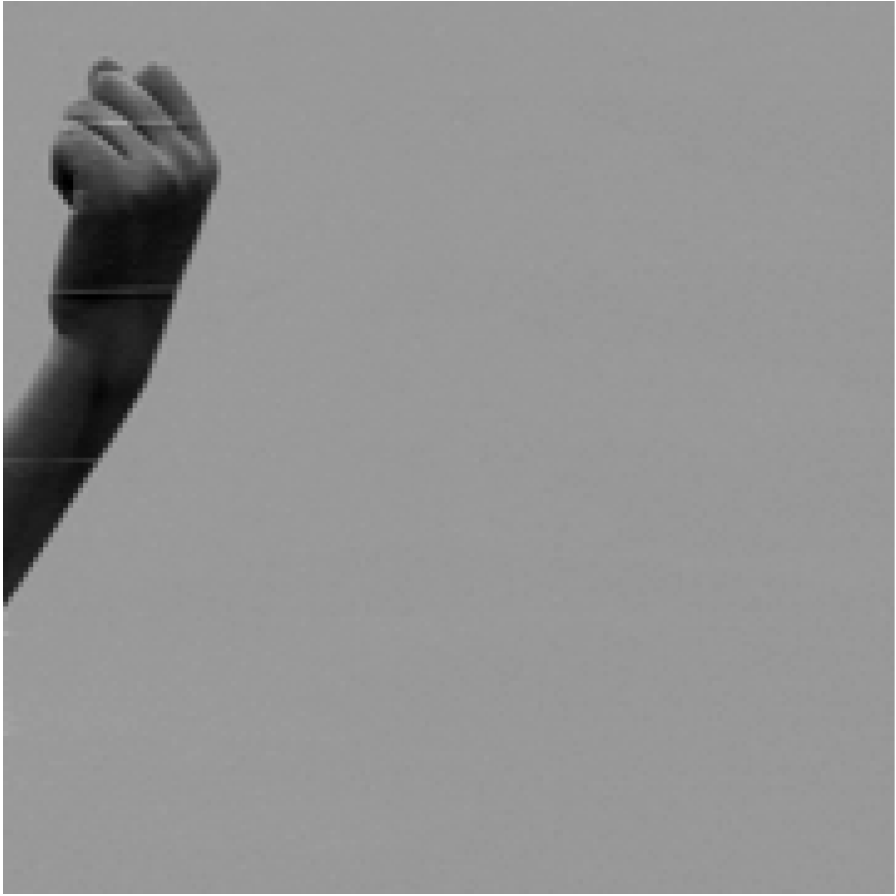}&
\includegraphics[width=0.19\textwidth]{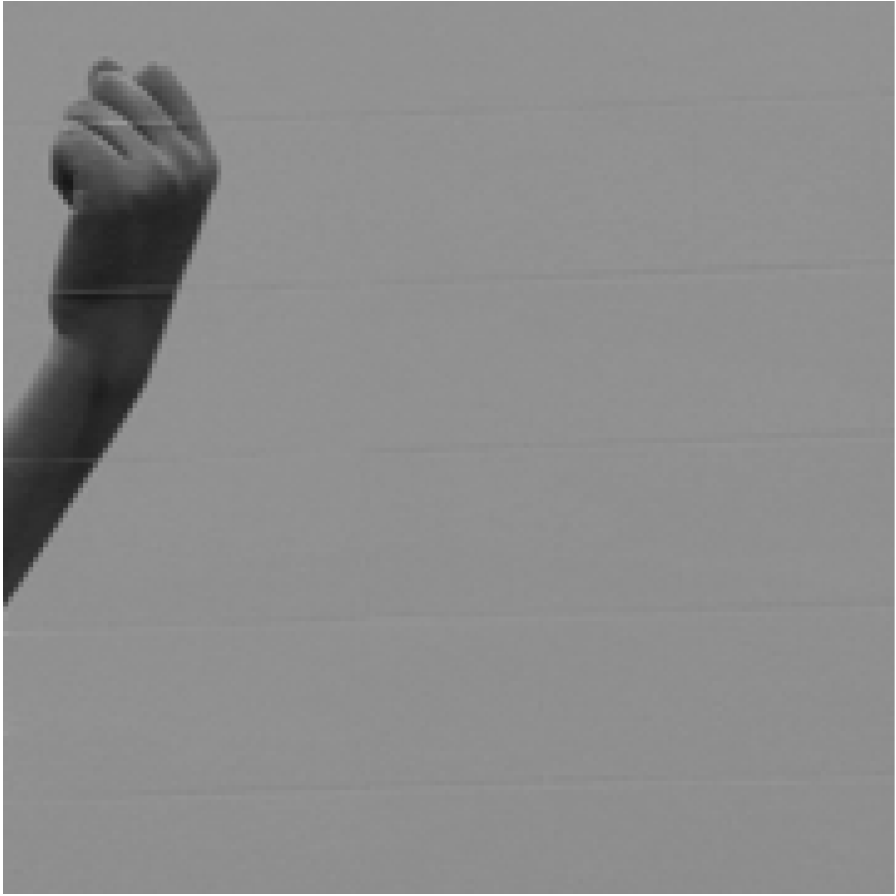}\\
Original frame &  LAGO &  SPCP & SPGL1 &  Alg.1
\end{tabular}
\caption{Detected objects for the arm motion video. The three rows correspond to the first, the middle and the last video frames, respectively.}\label{fig:exp2fg}
\end{figure*}

In the second experiment, an arm movement video was recorded when the student volunteer rotated her forearm and hand around the elbow joint slowly. It is usually difficult to detect rotation since different portions of the moving object have different motion velocities and directions, as well as one fixed origin or axis.
The tested video is generated by removing motionless frames and cropping the region of interest, which consists of 32 frames and each frame has $180\times 180$ pixels. The visual comparison of foreground and background for all results are shown in Fig.~\ref{fig:exp2bg} and Fig.~\ref{fig:exp2fg}, respectively. In Table~\ref{tab2}, we compare the qualities of the recovered background and foreground. Notice that there is movement still left on the left of the LAGO background and foreground results while some speckle noise exist in the SPCP foreground. Both SPGL1 and our approach can separate the foreground and the background clearly.

In terms of running time, SPCP takes the minimum running time ($\sim$0.1 s) while SPGL1 based on Newton's iteration takes about 50 seconds. Both LAGO and our algorithm run about 5 seconds and the graph Laplacian construction can be fast using a small number of neighbors. Overall, our method can keep a good balance in running time and detection accuracy. This phenomenon also applies to the first experiment.

\begin{table}[ht]
\centering
\caption{Quantitative comparison for the arm motion video}\label{tab2}
\begin{tabular}{c|cc|ccc}
\hline \hline
    & RE & PSNR & Pr & Re & Fm\\ \hline
LAGO &0.0151&20.36&0.9617&0.8305&0.8913\\
SPCP& 0.0132&20.44&0.9666&0.8471&0.9029 \\
SPGL1 & 0.0132 & 35.65 & 0.9665 & 0.8610&0.9107 \\
Alg.~1 & 0.0104 & 35.67 & 0.9704 & 0.8234&0.8909\\
\hline\hline
\end{tabular}
\end{table}

\subsection{Experiment 3: Multi-Object Motion Detection}
\begin{figure*}[ht]
\centering\setlength{\tabcolsep}{2pt}
\begin{tabular}{ccccc}
\includegraphics[width=0.19\textwidth]{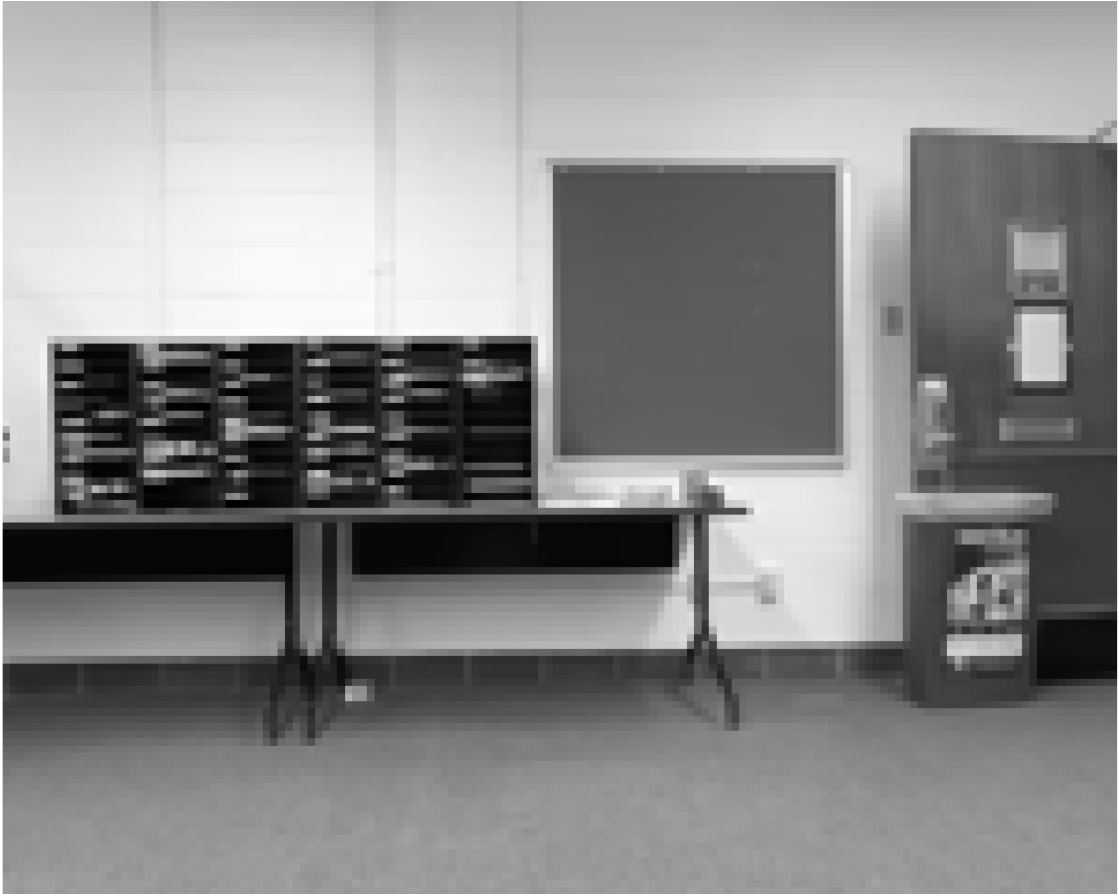}&
\includegraphics[width=0.19\textwidth]{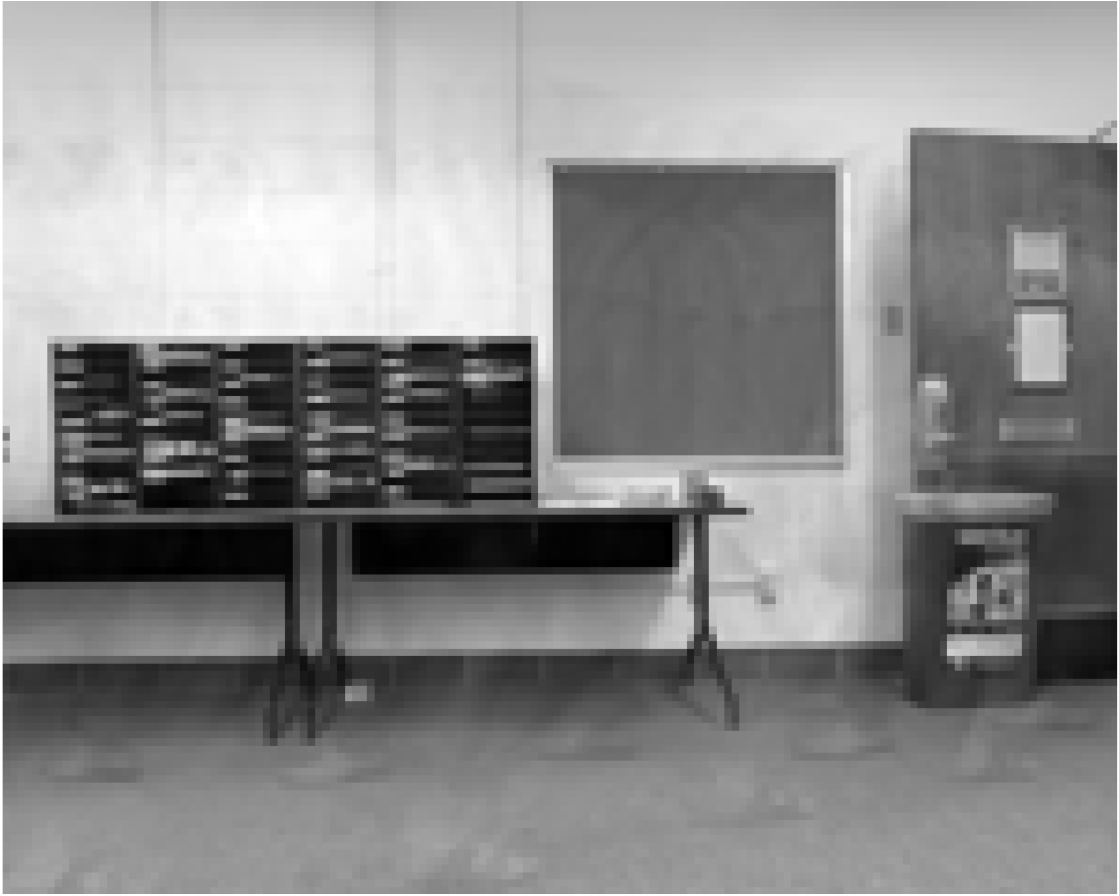}&
\includegraphics[width=0.19\textwidth]{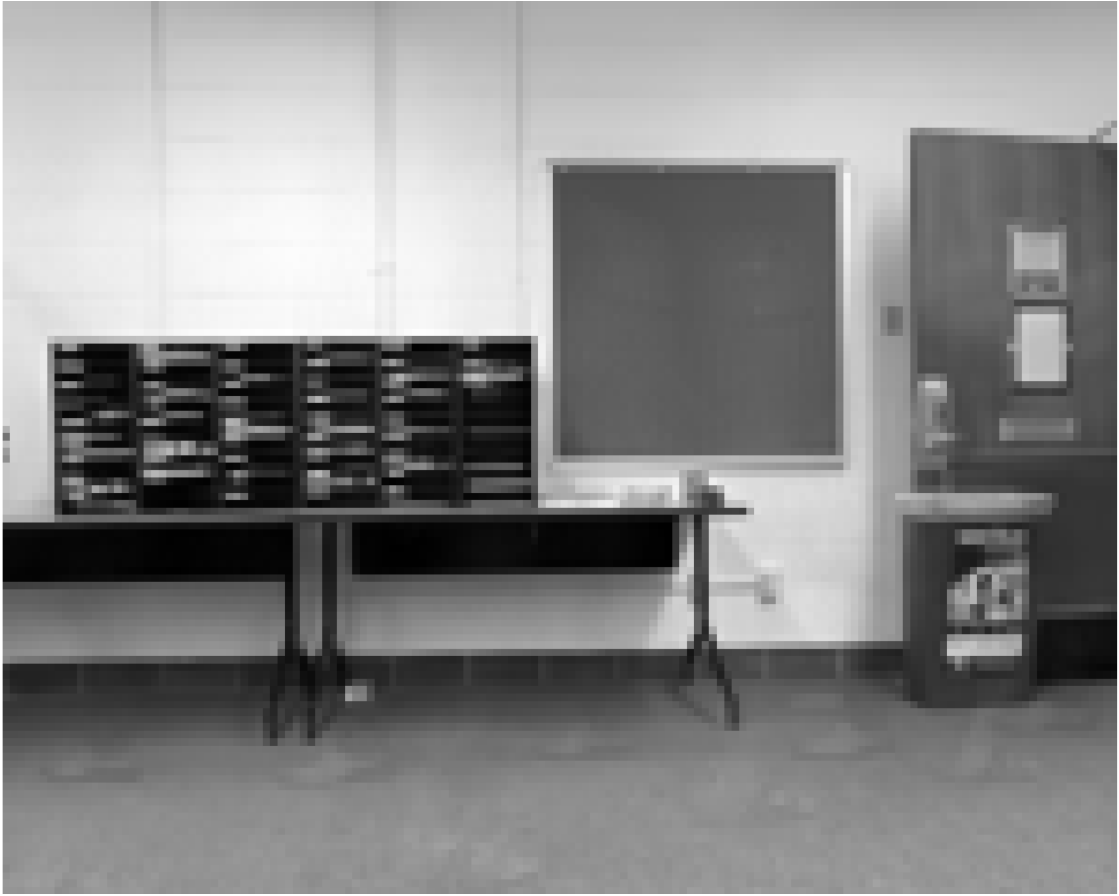}&
\includegraphics[width=0.19\textwidth]{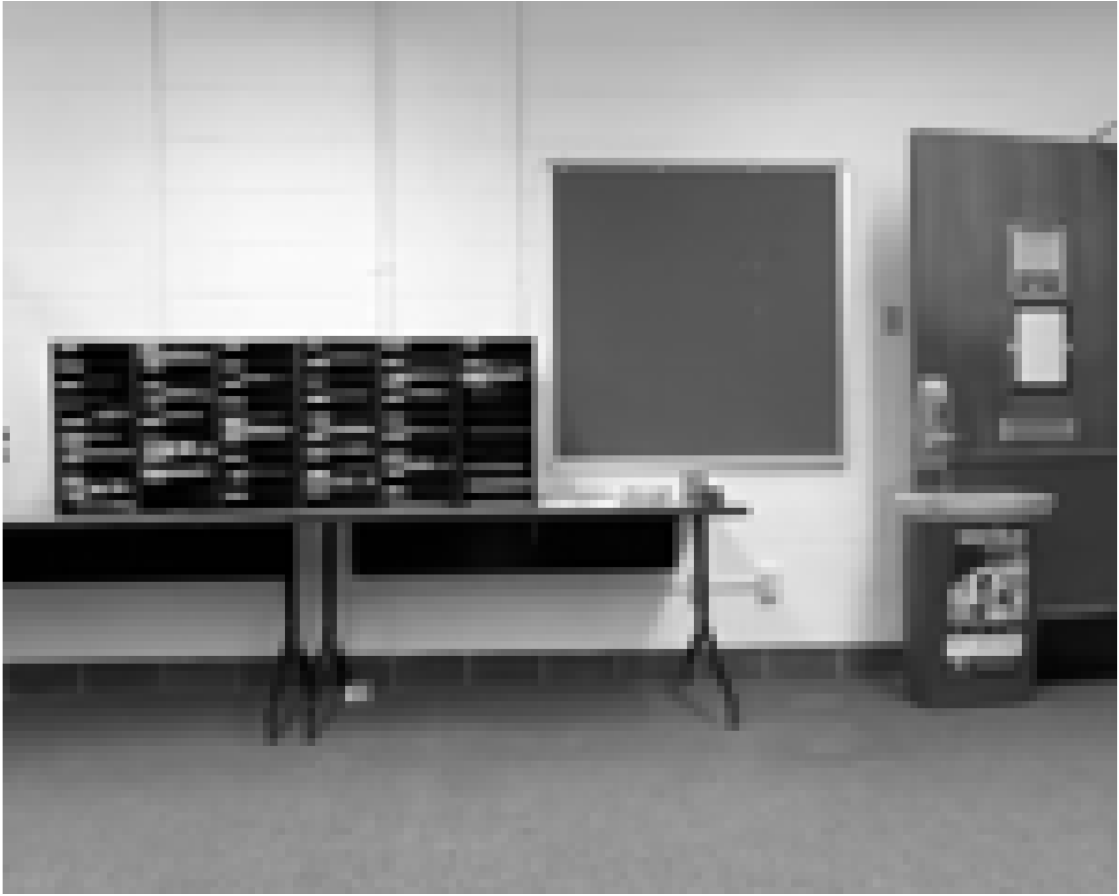}&
\includegraphics[width=0.19\textwidth]{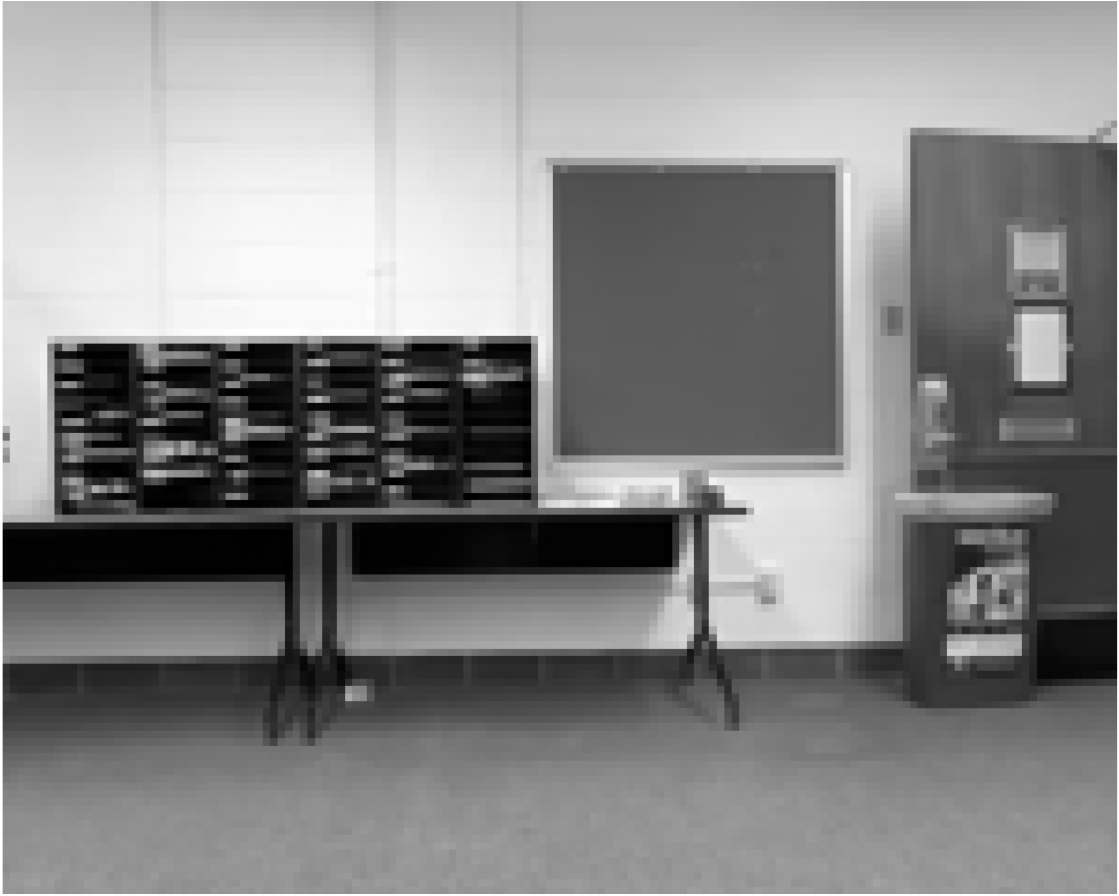}\\
 Ground truth &  LAGO &  SPCP &  SPGL1 &  Alg.1
\end{tabular}
\caption{Recovered backgrounds of the two-person walking video via various methods. }\label{fig:exp3bg}
\end{figure*}

\begin{figure*}[ht]
\centering\setlength{\tabcolsep}{2pt}
\begin{tabular}{ccccc}
\includegraphics[width=0.19\textwidth]{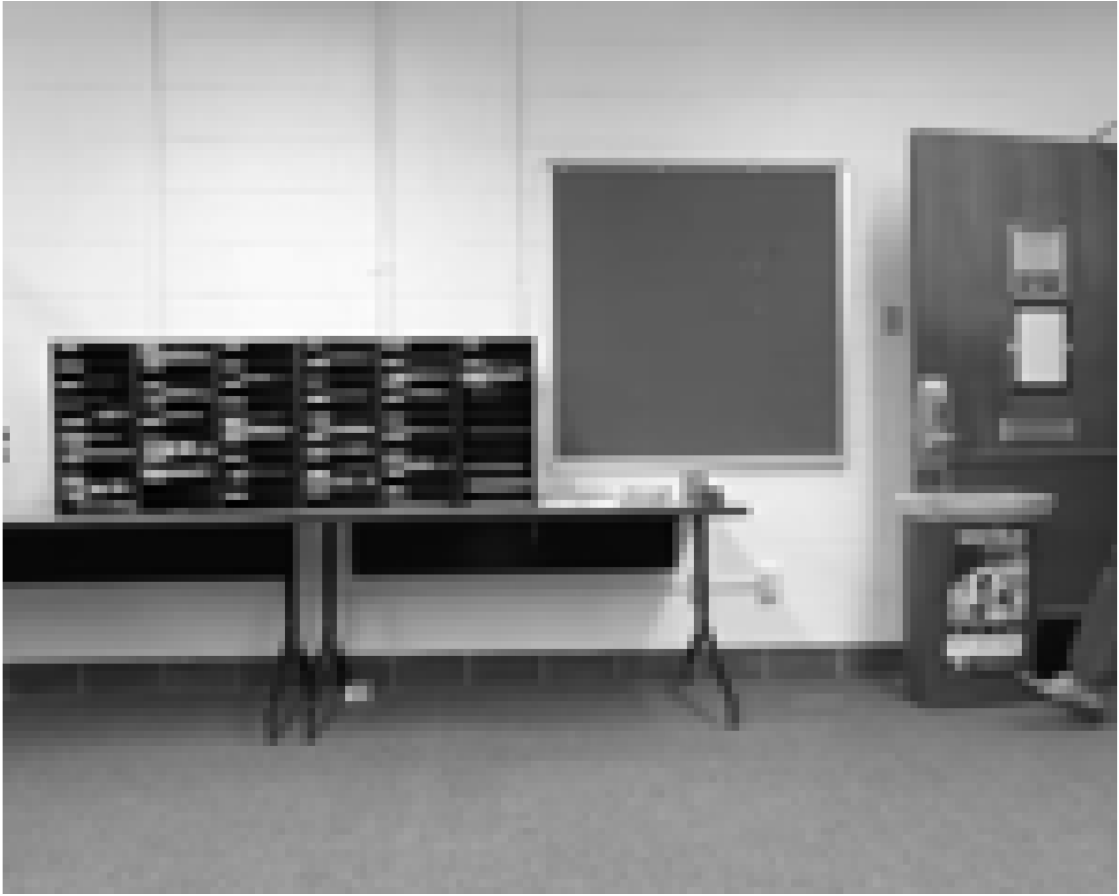}&
\includegraphics[width=0.19\textwidth]{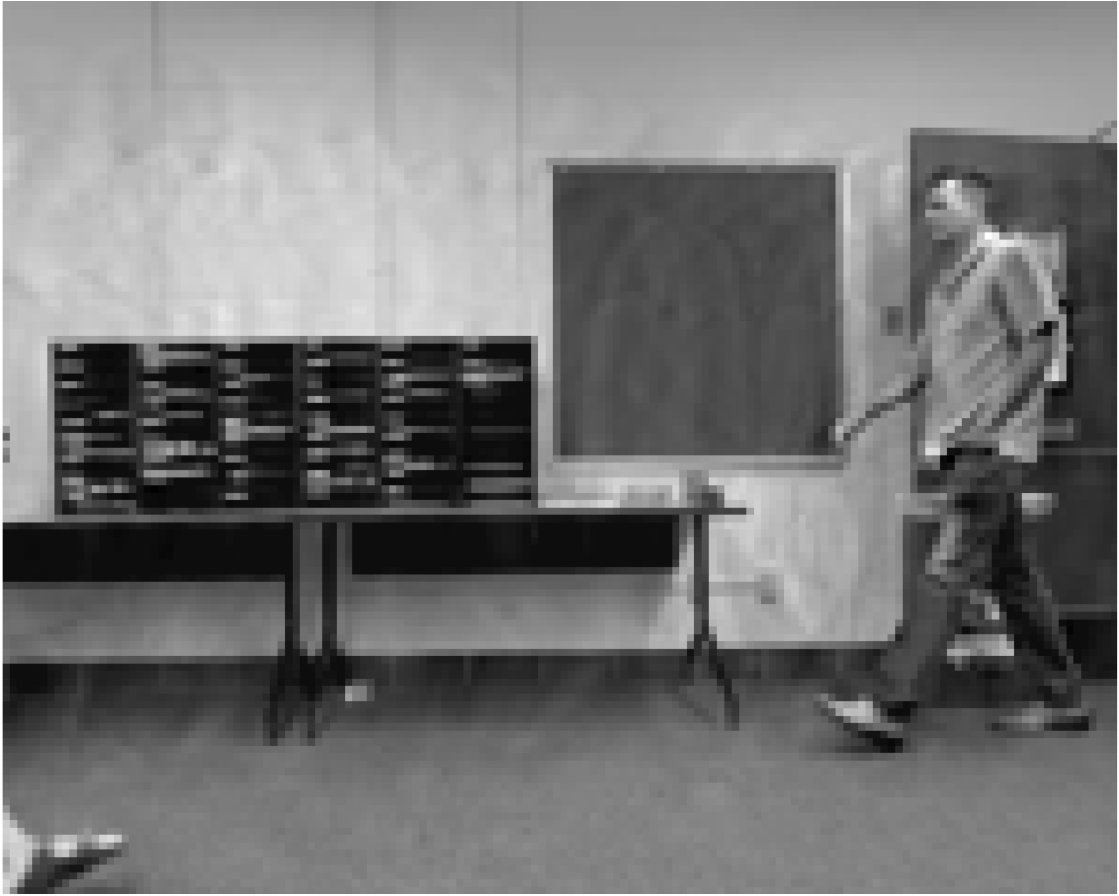}&
\includegraphics[width=0.19\textwidth]{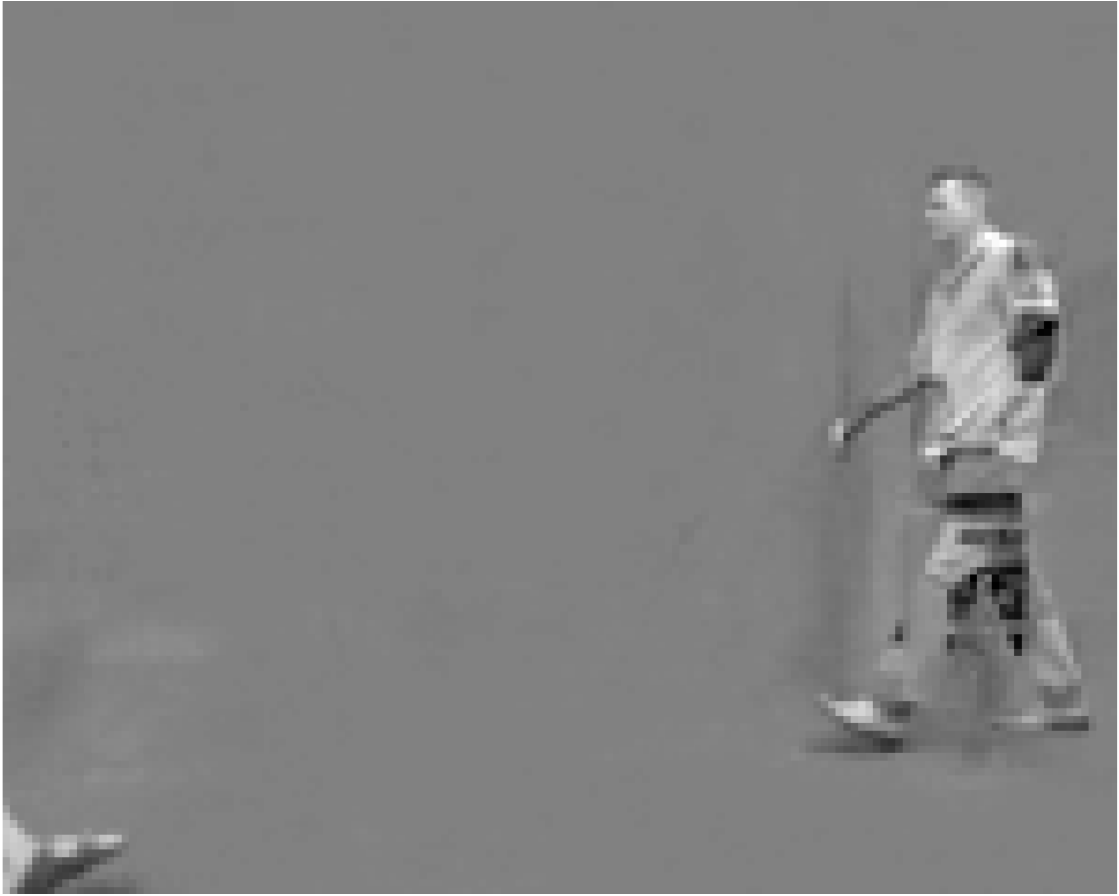}&
\includegraphics[width=0.19\textwidth]{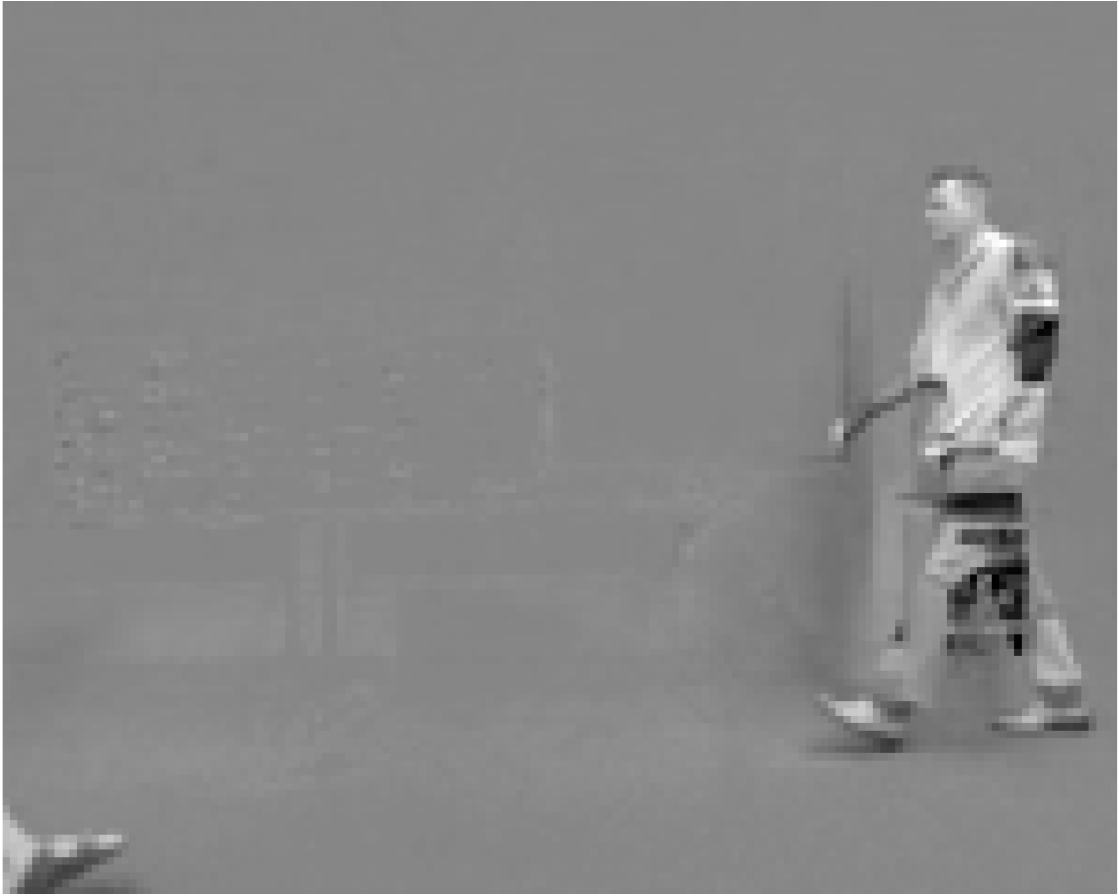}&
\includegraphics[width=0.19\textwidth]{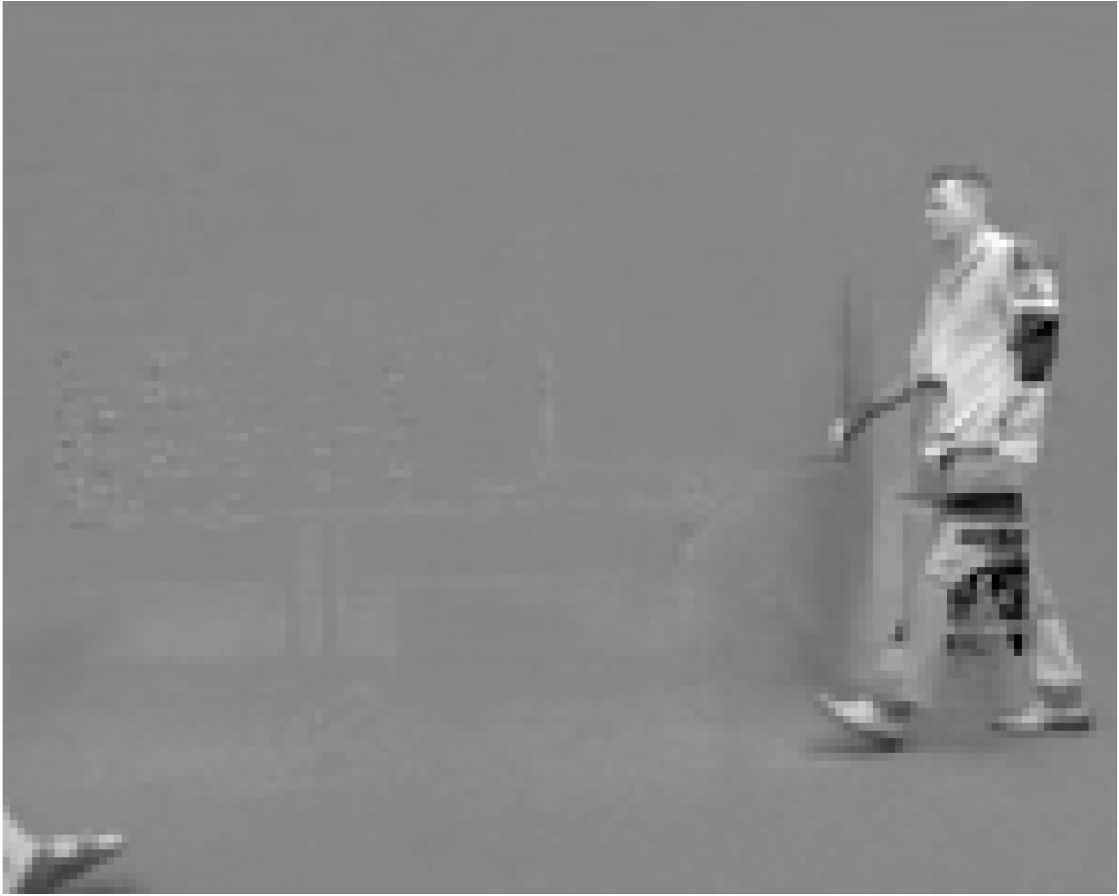}\\
\includegraphics[width=0.19\textwidth]{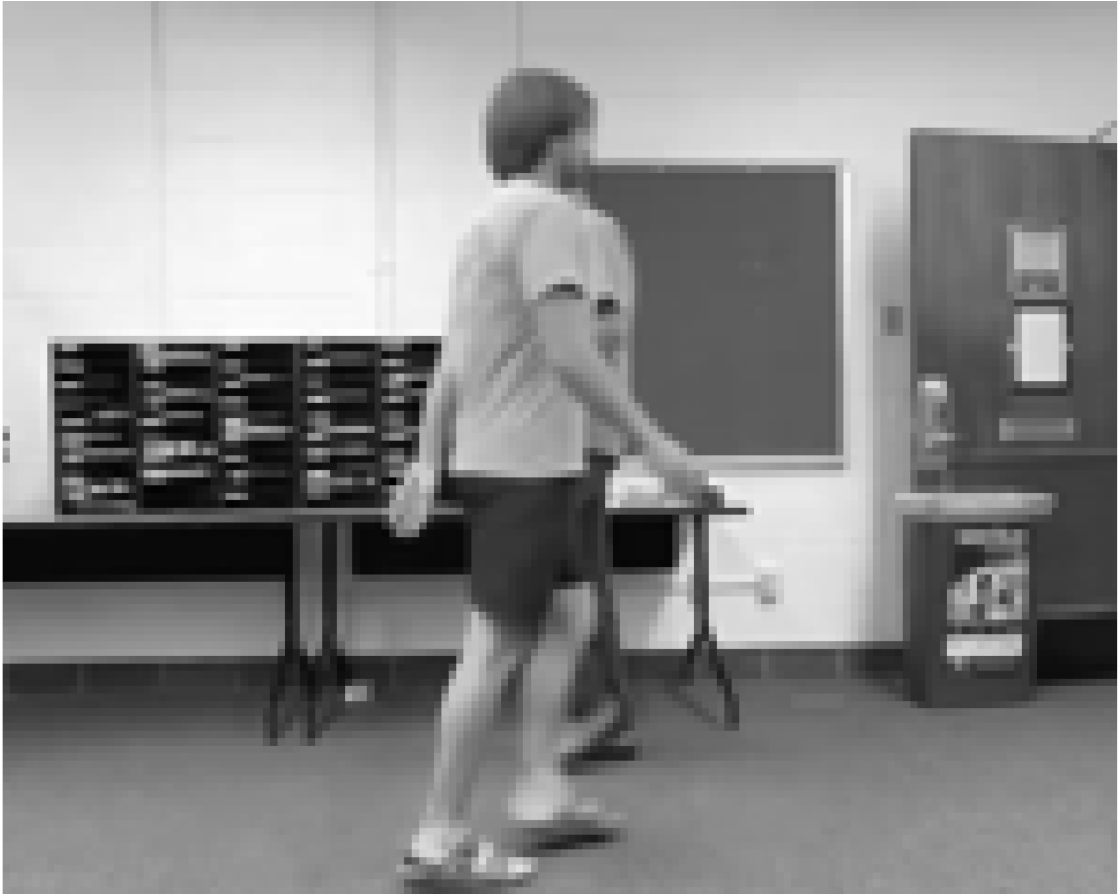}&
\includegraphics[width=0.19\textwidth]{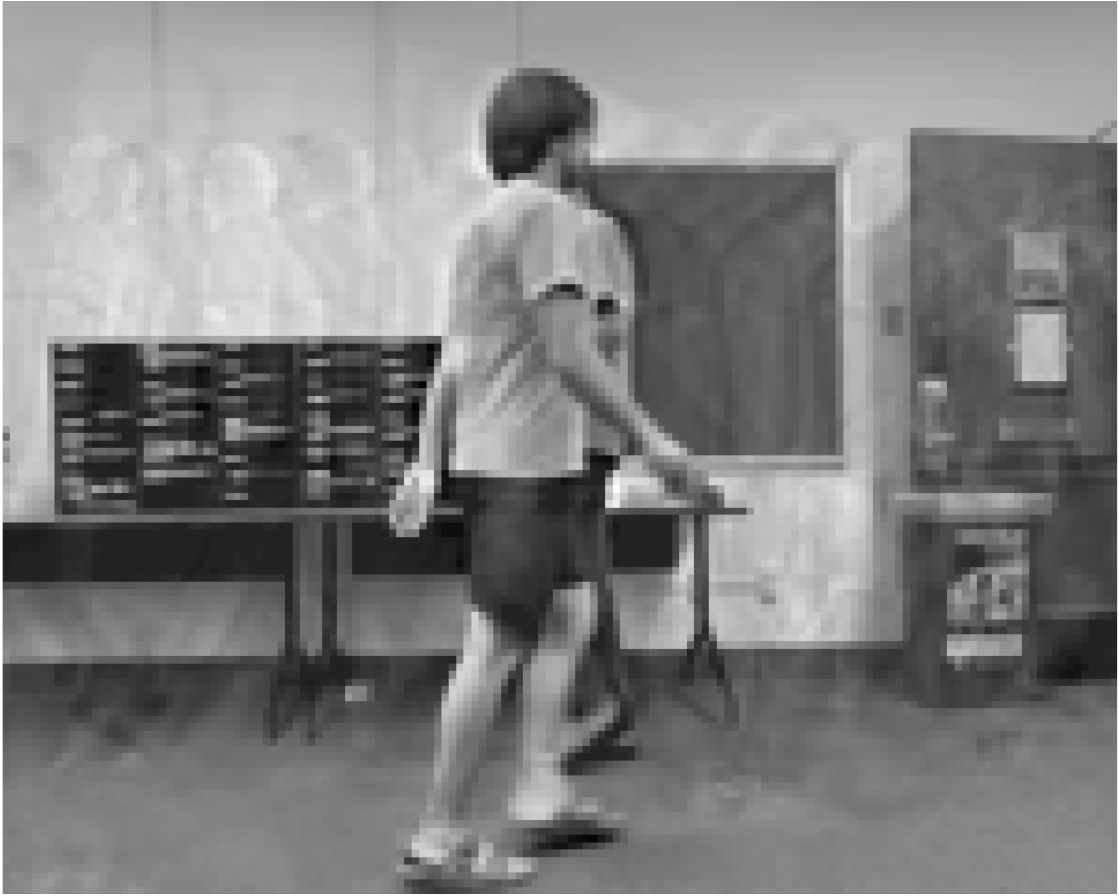}&
\includegraphics[width=0.19\textwidth]{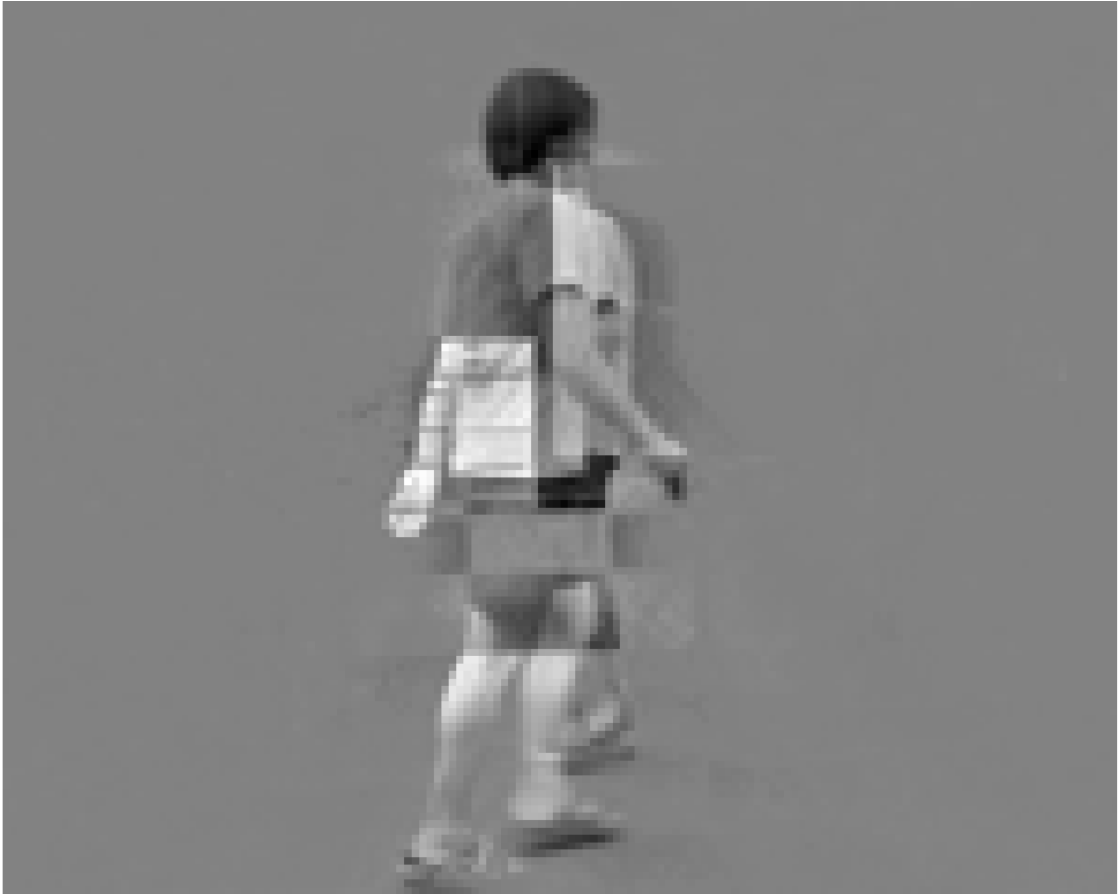}&
\includegraphics[width=0.19\textwidth]{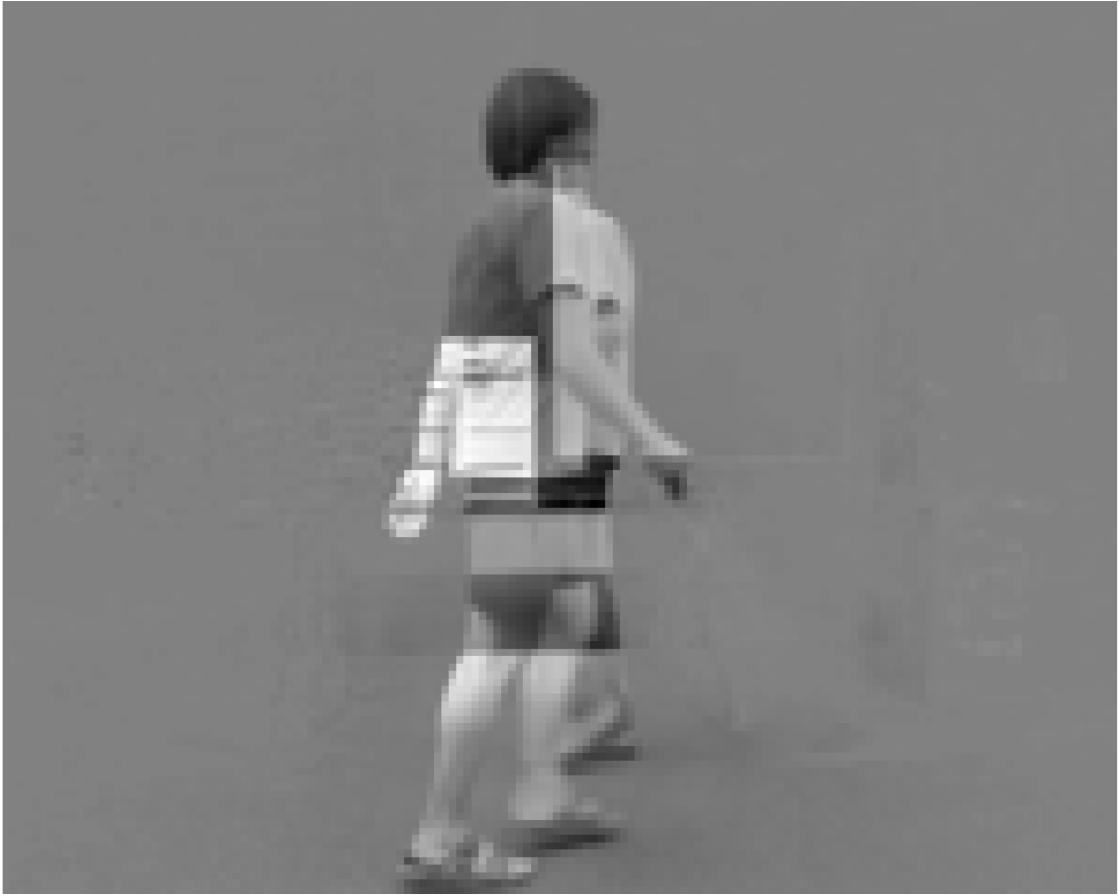}&
\includegraphics[width=0.19\textwidth]{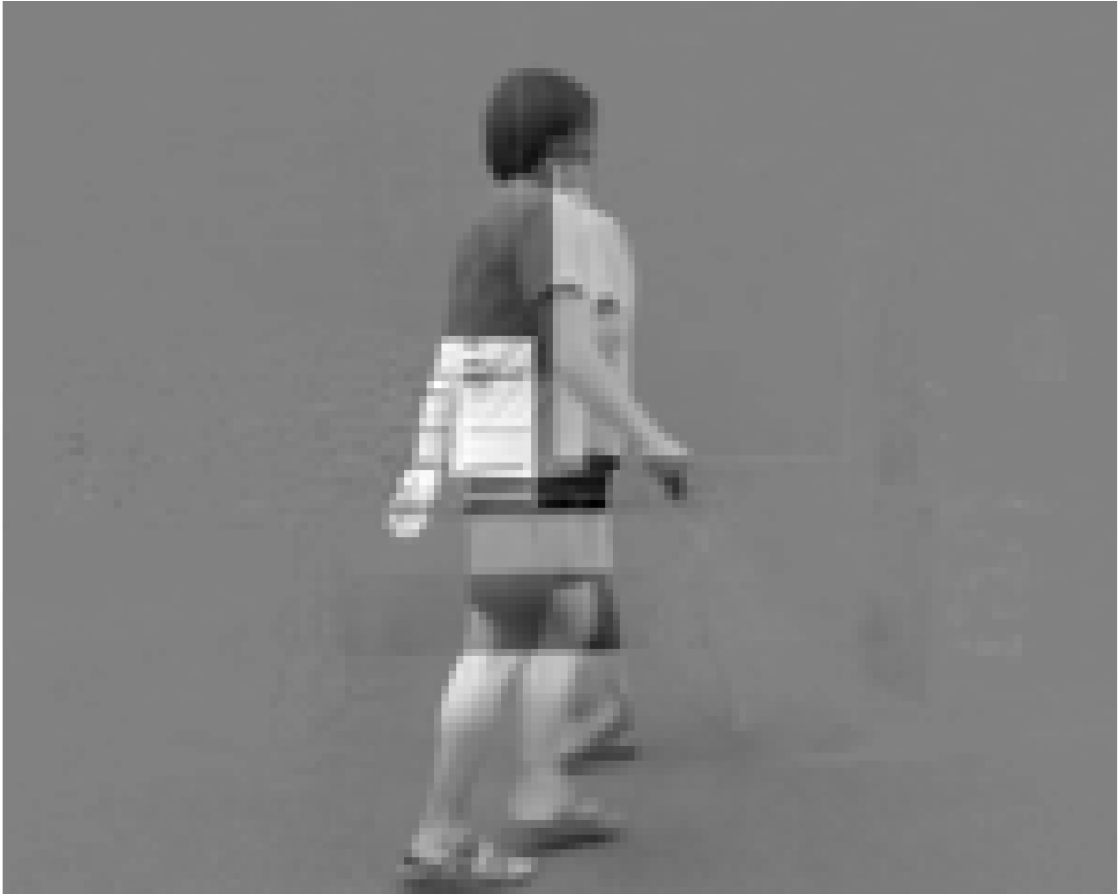}\\
\includegraphics[width=0.19\textwidth]{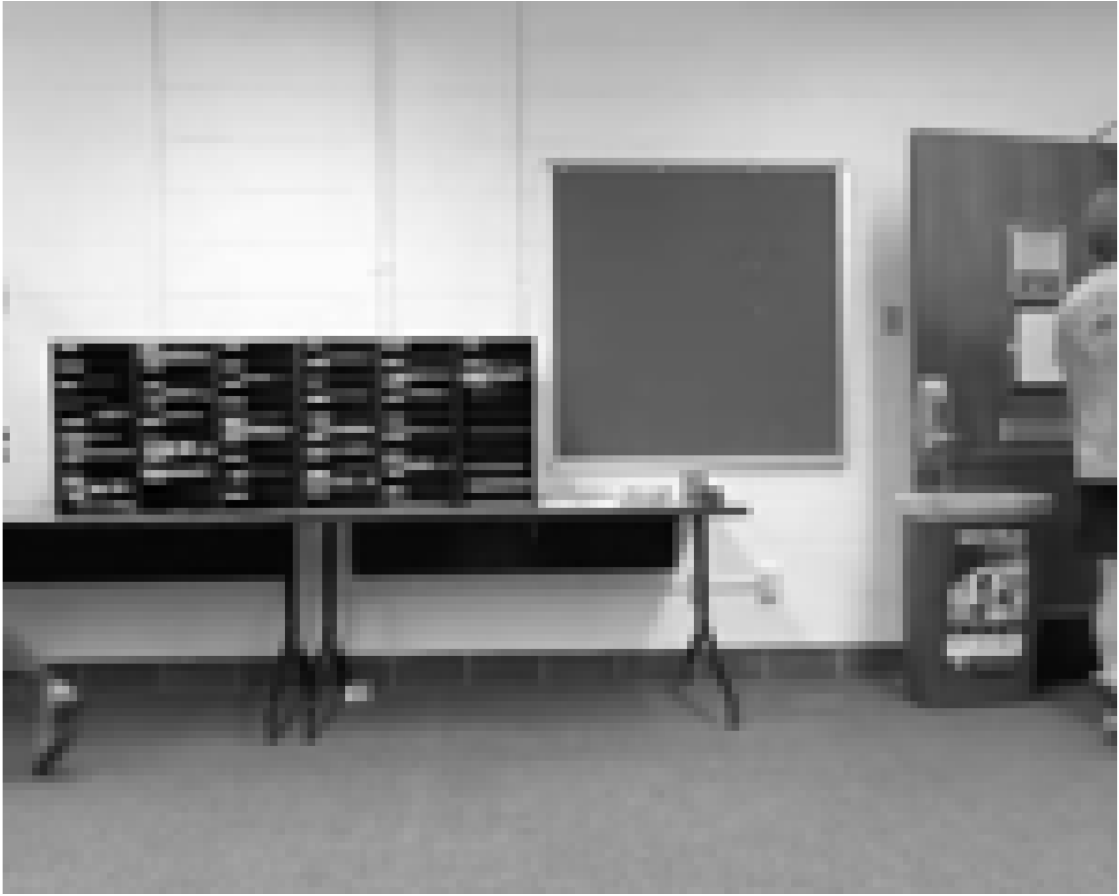}&
\includegraphics[width=0.19\textwidth]{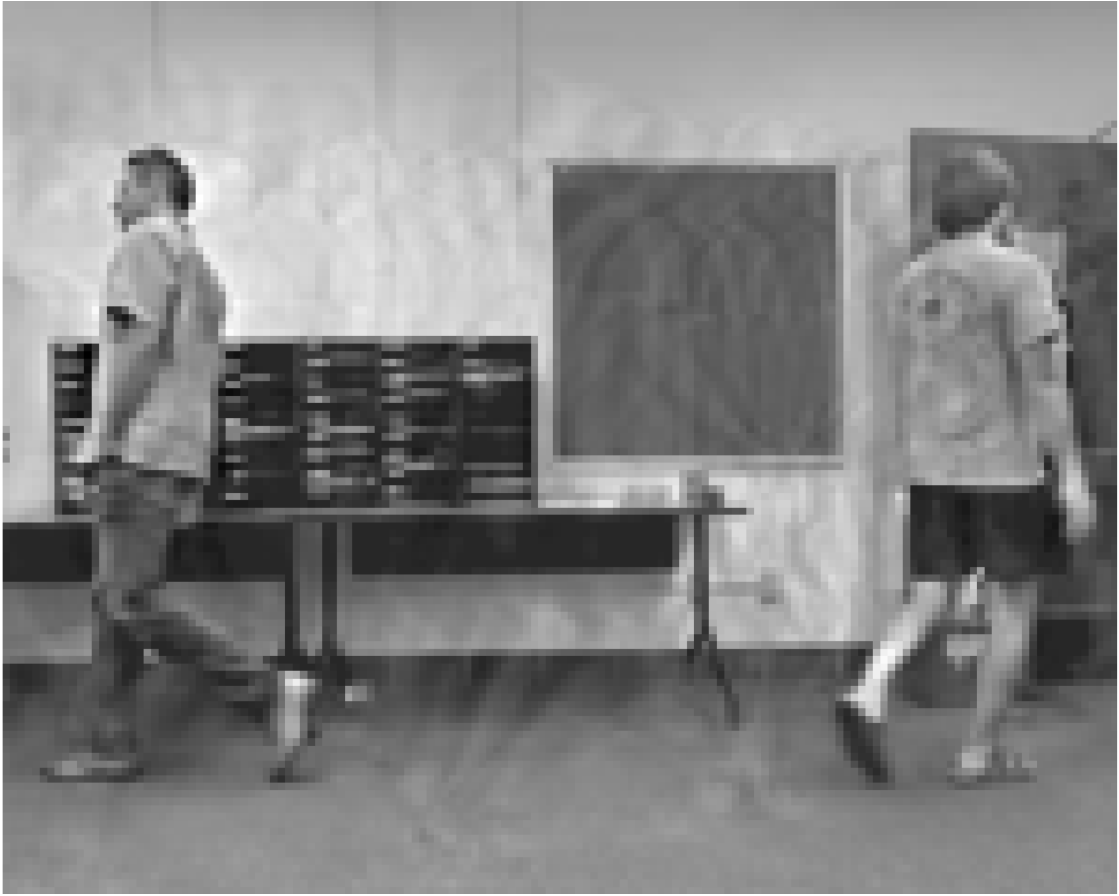}&
\includegraphics[width=0.19\textwidth]{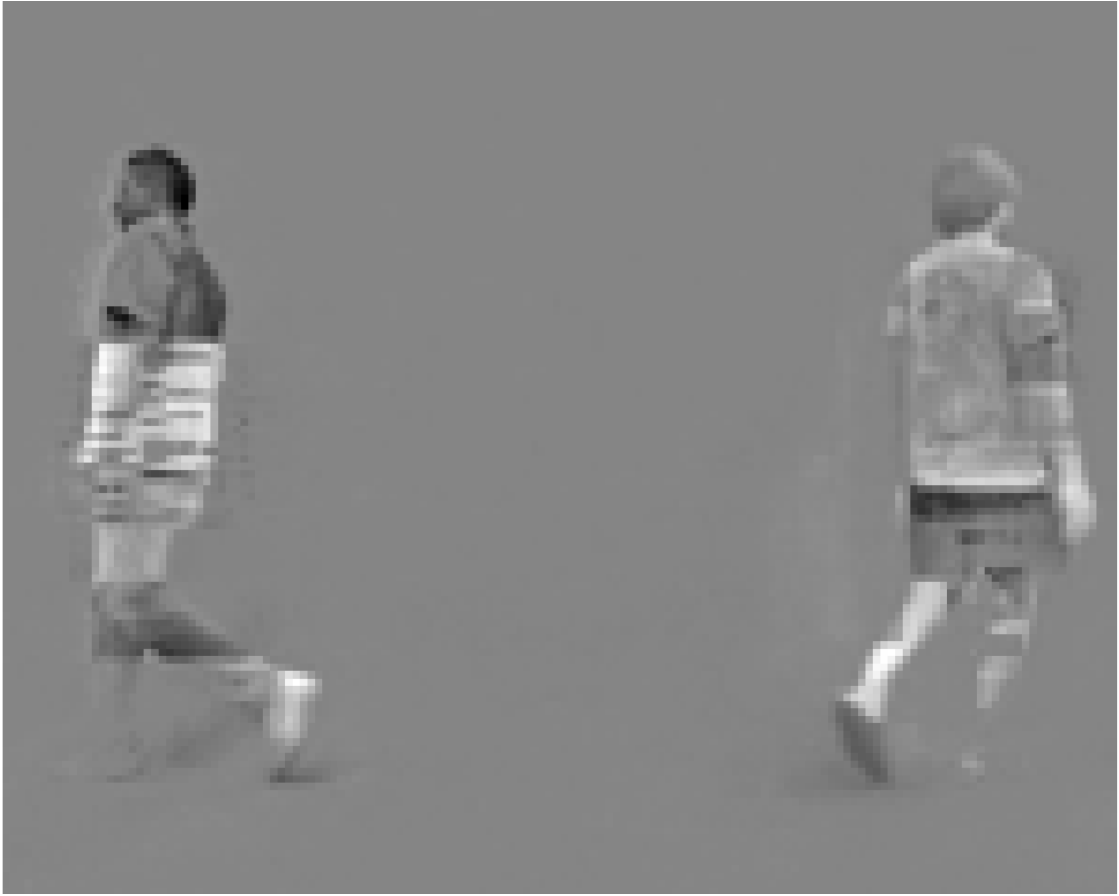}&
\includegraphics[width=0.19\textwidth]{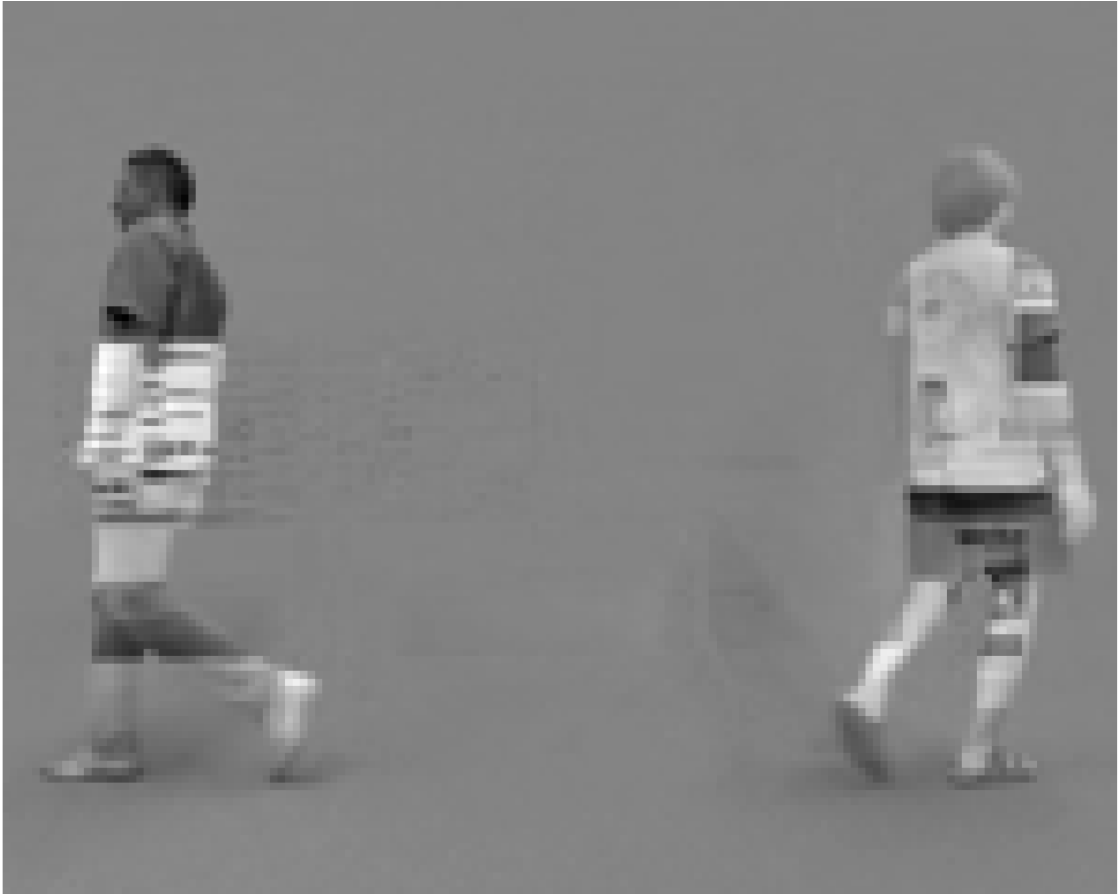}&
\includegraphics[width=0.19\textwidth]{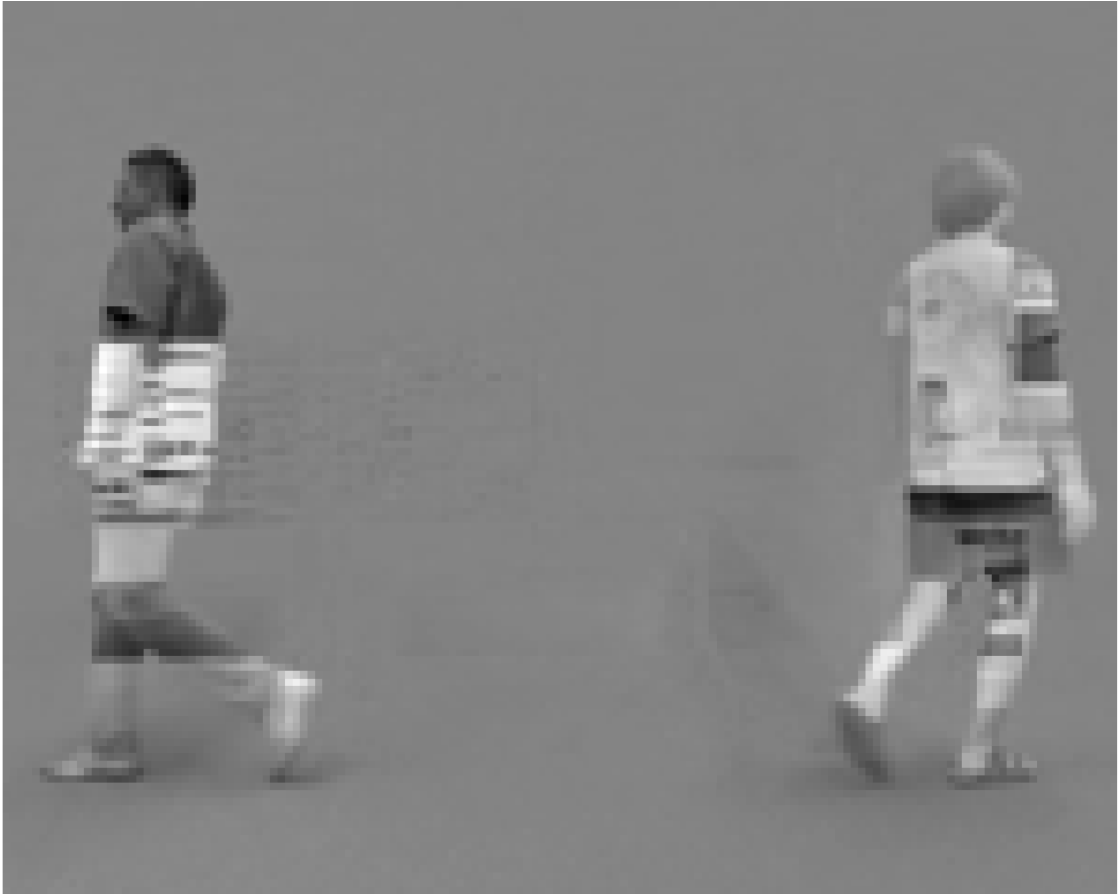}\\
 Original frame &  LAGO (fail) &  SPCP &  SPGL1 &  Alg.1
\end{tabular}
\caption{Detected objects for the two-person walking video via various methods. The three rows correspond to the first, the middle and the last video frames, respectively.}\label{fig:exp3fg}
\end{figure*}

In the third experiment, we tested a video where two persons are walking from opposite directions in a lab room. Specifically, the video consists of 100 frames, each of $120\times 150$ pixels. The ground truth for the background is estimated by taking the average of the first 135 frames which have no moving objects under stable lighting conditions. Fig.~\ref{fig:exp3bg} shows the mean images of the backgrounds recovered by each comparing method. One can see that our result has the best image quality than the others, which is also quantitatively verified in Table~\ref{tab3} comparing the relative errors between the ground truth and the recovered background, as well as image PSNR values. Different from the previous two experiments, since two persons will have body overlapped when they passed each other, we do not have the exact ground truth for motion from both persons. In Fig.~\ref{fig:exp3fg}, we show the motion detected by all the methods. Notice that under the optimal parameter setting, LAGO fails to retrieve valid motion in this test while the other results show the reasonable human motion. A proper range of intensity values is needed for better visualization.

\begin{table}[ht]
\centering
\caption{Quantitative comparison for the recovered backgrounds of two-person walking video}\label{tab3}
\begin{tabular}{c|cccc}
\hline \hline
    & LAGO & SPCP & SPGL1 & Alg.~1\\ \hline
RE &0.0664&0.0465&0.0372&0.0348\\
PSNR& 28.0854&31.1734&33.1172&33.6839 \\
%SSIM & 0.9514 & 0.9812 & 0.9877 & 0.9873 \\
\hline\hline
\end{tabular}
\end{table}

\subsection{Discussions on Graph Generation and Parameter Selections}
\paragraph{Impact of Graph Laplacians.} For a spatiotemporal video sequence, we can generate two types of graph Laplacians (in symmetric normalized form) as discussed in Section ~\ref{sec:lap}: spatial and temporal graph Laplacians. First, similarity metric is important in designing a graph Laplacian. One key component of generating graph Laplacians is to compute the pairwise or patchwise similarity. In addition to the exponential function used in defining the adjacency matrices in \eqref{eqn:sims} and \eqref{eqn:simt}, one can also use cosine similarity
\[
\cos(\vu,\vv):=\frac{\vu\cdot \vv}{\norm{\vu}_2\norm{\vv}_2}
\]
and then define the spatial and temporal adjacency matrices as follows
\begin{equation}\label{eqn:simc}
\begin{aligned}
(A_s)_{i,j}&=\cos(\vv_i^t,\vv_j^t),\\
(A_t)_{i,j}&=\cos(\cN(\vv_i^s),\cN(\vv_j^s)).
\end{aligned}
\end{equation}
Computation of cosine similarity does not need to set a proper filtering parameter but is less robust than the exponential type sometimes since cosine similarity treats the features as independent and completely different. For example, under the same parameter settings as in Experiment 1-3 with no noise, the cosine similarity based graph Laplacaian leads to the metric values shown in Table~\ref{tab6}. It can be seen that both relative errors and PSNR values using cosine similarity based graph Laplacians are slightly worse than or the same as those using exponential based graph Laplacians.
% exp1          RE: 0.0147
%         Lpsnr: 41.8699
%         Lssim: 0.9943
%    Sprecision: 0.9687
%       Srecall: 0.7191
%       Sfscore: 0.8255

\begin{table}
\centering
\begin{tabular}{c|ccc}\hline\hline
Experiment \# & 1 & 2 & 3\\ \hline
RE& 0.0147& 0.0211&0.0348\\
PSNR& 41.8700& 35.6417&  33.6845\\ \hline\hline
\end{tabular}
\caption{Comparison of recovered backgrounds in Experiments 1-3 without noise via cosine similarity based graph Laplacians.}\label{tab6}
\end{table}

Second, the size of spatial or temporal neighborhoods makes an impact on the computational time and the detection performance. The larger is the neighborhood size, more time would be expected for computing the similarities and less sparse for the generated graph Laplacians, which causes more computational time for solving the $L$-subproblem in the algorithm. However, if the neighborhood size is very small or the neighborhood is not symmetric, then it is very likely that the local smoothness is not sufficiently captured which may result in poor smoothness preservation for the recovery. Since a temporal graph is an undirected line graph, we can use a neighborhood with odd vertices centered at the target temporal point. For spatial graphs, we use a neighborhood with $4n+1$ pixels and $n$ is an integer centered at the target pixel. Nevertheless, both graphs will involve extension of boundary points, and we use mirror extension to preserve the smoothness along the boundaries. Further, optimal parameters, especially graph regularization parameters, will be different for different neighborhood sizes of graphs due to the change in the scale of both graph regularizations. Throughout our experiments, we fix the neighborhood sizes for both graphs as 5, i.e., 4 nearest spatial or temporal points being contributed for weights.

\paragraph{Parameter Selection.} At the current stage, since we do not have substantial data sets which are similar to our test videos as in our previous experiments, we resort to model based methods rather than training based ones so that we do not need to learn thousands to millions of hyper-parameters for deep learning architectures. However, we still encounter a moderate load of parameter tuning for the proposed algorithm. We use the grid search and find optimal parameters which yield the smallest background recovery error.  Surprisingly, not all parameters are necessary to be hard tuned. The parameters chosen for the Experiments 1-3 in the noise-free cases are listed in Table~\ref{tab5}, where $dt>0$ is the step size for gradient descent in the $L$-subproblem \eqref{eqn:Lupdate} and $\beta$ is a decay factor to update $\lambda_2$ every five iterations, i.e., $\lambda_2\leftarrow \lambda_2/\beta$.  If $\lambda_2$ is relatively large compared to the scale of $D-L$ in \eqref{eqn:Supdate}, then the shrink operator fails to take effect for the subsequent iterations. To avoid this, we can let $\lambda_2$ gradually decay to match the scale of $D-L-V+\widetilde{V}$. The selection of $\beta$ and $\lambda_2$ must be cautious since large $\beta\gg1$ can lead to divergence, which can also cause underflow in $\lambda_2$. Thus a minimum value for $\lambda_2$ is typically set in practice. For the noisy cases in Experiment 1, $\rho_1=\rho_2=\lambda_2=0.1$ while $\lambda_1$ needs to be tuned between 0.1 and 10.

\begin{table}
\centering
\begin{tabular}{c|ccc}
\hline\hline
Experiment \# & 1 & 2 & 3\\ \hline
$\lambda_1$ & 1e2& 1e-4& 1e5\\
$\lambda_2$& 1e-1& 1e-1 & 1\\
$\gamma_1$ &1e-6&1e-5& 1e-6\\
$\gamma_2$ &1e-8&1e5& 1e-8\\
$\rho_1$&1&1e-3& 1e1\\
$\rho_2$&1&1e1& 1e-2\\
$dt$ & 1e-1 & 1e-5& 1e-1\\
$\beta$ & 1 & 1.05 & 1\\
\hline\hline
\end{tabular}
\caption{Parameters used for all the noise-free experiments. The numeric format ``$x$e$k$'' denotes the number $x\times 10^{k}$.}\label{tab5}
\end{table}

\section{Conclusions and Future Work}\label{sec:con}
Motion detection is one of the most fundamental tasks in video processing with a wide spectrum of applications, particularly in human-robot interaction. In the case of limited lightening conditions and/or time-varying illuminations, it becomes extremely challenging to separate one or multiple moving objects with or without shadow from a static background. Naturally, one can decompose each single frame into the foreground and the background components. However, it is prone to lose spatial and temporal smoothness and increase the computational burden. In this work, we propose a novel dual-graph regularized motion detection approach. In particular, we exploit the spatiotemporal geometry of the foreground by constructing a spatial and a temporal graphs described by their respective graph Laplacians, and then employ a weighted nuclear norm regularizer based on the error function so that low-rankness of the background can be adaptively learned during the detection process. The proposed algorithm is obtained by applying the ADMM algorithm framework. A variety of numerical experiments have shown that the proposed algorithm consistently outperforms the other related ones on realistic data sets with various human motions. In the near future, we will develop fast algorithms involving the low-rank tensor decompositions and take into consideration the shadow due to the motion of objects under complex lightening environments.

\section*{Acknowledgements}
The research of Qin is supported by the NSF grant DMS-1941197, and the research of Xie is supported by the University of Kentucky College of Engineering Young Alumni Philanthropy Council Funding. The authors would like to thank Ruihan Zhu and Ruilong Shen for recording videos.

%%%%%%%%%%%%%%%%%%%%%%%%%%%%%%%%%%%%%%%%%%%%%%%%%%%%%%%%%%%%%%%%%%%%%%%%%%%%%%%%%%%%%

\bibliographystyle{unsrt}
\bibliography{ref}

\end{document}